\pdfoutput=1

\documentclass[11pt]{article}

\usepackage[final]{acl}

\usepackage{times}
\usepackage{latexsym}

\usepackage[T1]{fontenc}

\usepackage[utf8]{inputenc}

\usepackage{microtype}

\usepackage{inconsolata}

\usepackage{graphicx}

\usepackage{algorithm}
\usepackage{algpseudocode}

\usepackage{graphicx}
\usepackage{multirow,multicol}
\usepackage{needspace} 
\usepackage{booktabs}
\usepackage{amsmath}
\usepackage{amssymb}
\usepackage{adjustbox}
\usepackage{semtrans}
\usepackage{svg}
\usepackage{pdfpages}
\usepackage{graphicx}
\usepackage{subcaption}
\usepackage{placeins}
\usepackage{tipa}
\usepackage{xtab}
\usepackage{diagbox}
\usepackage{xcolor,colortbl}
\usepackage{makecell}
\usepackage{comment}
\usepackage{hyperref}
\usepackage{float}
\usepackage{xltabular}

\newcommand{\R}{\mathbb{R}}
\newcommand{\nnr}{\textsc{NN-Rank}}

\usepackage[nameinlink,capitalize,noabbrev]{cleveref}

\crefname{equation}{equation}{equations}   
\crefname{footnote}{footnote}{footnotes}   
\crefname{section}{\S}{\S\S}
\Crefname{section}{\S}{\S\S}    
\crefformat{section}{#2\S#1#3}  
\Crefformat{section}{#2\S#1#3}
\crefrangeformat{section}{\S\S#3#1#4--#5#2#6}
\Crefrangeformat{section}{\S\S#3#1#4--#5#2#6}
\crefformat{section}{#2\S#1#3}  
\crefrangeformat{section}{\S\S#3#1#4--#5#2#6}

\title{Model-Based Ranking of Source Languages \\ for Zero-Shot Cross-Lingual Transfer}

\author{Abteen Ebrahimi,${ }^{1}$ Adam Wiemerslage,${ }^{2}$ \and Katharina von der Wense${ }^{1,3}$ \\
  ${ }^{1}$University of Colorado Boulder, ${ }^{2}$Kensho Technologies \\${ }^{3}$Johannes Gutenberg University Mainz \\
  \texttt{abteen.ebrahimi@colorado.edu}}

\begin{document}

\maketitle

\begin{abstract}
We present \nnr{}, an algorithm for ranking source languages for cross-lingual transfer, which leverages hidden representations from multilingual models and unlabeled target-language data. We experiment with two pretrained multilingual models and two tasks: part-of-speech tagging (POS) and named entity recognition (NER). We consider 51 source languages and evaluate on 56 and 72 target languages for POS and NER, respectively. When using in-domain data, \nnr{} beats state-of-the-art baselines that leverage lexical and linguistic features, with average improvements of up to 35.56 NDCG for POS and 18.14 NDCG for NER. As prior approaches can fall back to language-level features if target language data is not available, we show that \nnr{} remains competitive using only the Bible, an out-of-domain corpus available for a large number of languages. Ablations on the amount of unlabeled target data show that, for subsets consisting of as few as 25 examples, \nnr{} produces high-quality rankings which achieve 92.8\% of the NDCG achieved using all available target data for ranking.
\end{abstract}

\section{Introduction} Cross-lingual transfer, where knowledge from a \textit{source} language is used to improve performance for a 
\textit{target} language, extends the coverage of languages supported by natural language processing tools. Pretrained multilingual models \cite{devlinBERTPretrainingDeep2019, conneauUnsupervisedCrosslingualRepresentation2020} are effective in this framework: the model is first finetuned on labeled data in the source language and then evaluated directly on target language data. Models show strong performance, even in a zero-shot setting where no labeled examples from the target language are used \cite{huXTREMEMassivelyMultilingual2020, ruderXTREMERMoreChallenging2021}.

\begin{table}[h]
    \begin{adjustbox}{width=\linewidth}
    \begin{tabular}{lll}
    \toprule
    Target Token & Top Five Neighbors & Source Dataset \\
    \midrule
    \multirow{5}{*}{{[}`probablement'{]}} & {[}`probabilmente'{]} &  UD\_Italian-ISDT \\
     & {[}`likely'{]} & UD\_English-GUM \\
     & {[}`verjetno'{]} &  UD\_Slovenian-SSJ \\
     & {[}`probablement'{]} &  UD\_Catalan-AnCora \\
     & {[}`provavelmente'{]} &  UD\_Portuguese-GSD \\
      \cmidrule{2-3}
     \multirow{5}{*}{{[}`\#\#isation'{]}} & {[}`\#\#ización'{]} &  UD\_Spanish-AnCora \\
     & {[}`\#\#ització'{]} & UD\_Catalan-AnCora \\
      & {[}`\#\#ación'{]} &  UD\_Spanish-AnCora \\
      & {[}`\#\#ção'{]} &  UD\_Portuguese-GSD \\
      & {[}`\#\#ció'{]} &  UD\_Catalan-AnCora \\
     \bottomrule
    \end{tabular}
    \end{adjustbox}
    \caption{Example of nearest neighbors for French target tokens calculated using mBERT representations. To create the ranking, for every target token we tally the number of occurrences of each source dataset in the right-most column, and sort them in decreasing order.}
    \label{tab:nn_main_example}
\end{table}
A critical element for successful transfer is the choice of source language, and while many prior works use only a single high-resource language such as English, this has been called into question \cite{turcRevisitingPrimacyEnglish2021}. An alternative is to rank all available source datasets for a given target language, as in LangRank \cite{linChoosingTransferLanguages2019}, which uses lexical features (such as word overlap) as well as linguistic features (such as word order, syntactic, and phylogenetic information). Other highly related prior works, which focus on analyzing how, and for which language pairs, these multilingual models are able to achieve strong cross-lingual transfer also focus on these features \cite{wuBetoBentzBecas2019,piresHowMultilingualMultilingual2019, kCrossLingualAbilityMultilingual2020, dufterIdentifyingElementsEssential2020, devriesMakeBestCrosslingual2022, riceUntanglingInfluenceTypology2025}. In general, pairs of languages with more similar features tend to yield better transfer performance.

While these \textit{static} features -- which we define as those independent of the model being evaluated -- are intuitive, we hypothesize that they cannot sufficiently describe the relationships between the representations of the languages contained within a pretrained multilingual model and, as such, do not provide an adequate signal for choosing source languages. For example, a model may be able to strongly transfer between two languages with dissimilar static features, e.g., inverted word order or no lexical overlap, given sufficient pretraining data. Conversely, two languages with high lexical overlap or similar language-level features may yield poor transfer performance due to weak representations from, e.g., low amounts of data. Consider a theoretically "perfect" source language for a specific target language; if the model is unable to encode the source language, e.g., due to script or vocabulary restrictions, strong cross-lingual performance can never be realized. 

Although additional features which describe the relationship between the model and each language (e.g., proportion of pretraining data) can be created, we propose removing hand-crafted features entirely and relying solely on hidden representations extracted from the intermediate layers of the model. We hypothesize that these representations implicitly capture (1) the relationship between the source and target language at the linguistic and lexical level and (2) the complex relationship between each language and the pretrained model, beyond the abilities of hand-crafted features, therefore serving as a stronger predictor of transfer performance. 

In this work, we present Nearest Neighbor-Rank (\nnr), a model and data-based approach to ranking source languages. Our experiments on part-of-speech tagging (POS) and named entity recognition (NER), using two popular multilingual models, multilingual-BERT \cite[mBERT;][]{devlinBERTPretrainingDeep2019} and XLM-RoBERTa \cite[XLM-R;][]{conneauUnsupervisedCrosslingualRepresentation2020}, show that this approach beats LangRank -- a state-of-the-art approach which relies on static lexical and linguistic features -- in all settings, when using the unlabeled development sets to create the ranking. We include 51 source languages for both tasks, 56 target languages for POS, and 72 target languages for NER.
As \nnr{} requires unlabeled data, we also experiment with rankings that are generated using only out-of-domain text taken from the Bible, a corpus covering over 1600 languages commonly used in data-scarce settings \cite{mccarthyJohnsHopkinsUniversity2020}, and find that \nnr{} continues to outperform LangRank in ranking quality. In addition to analysis on the impacts of domain mismatch, we conduct ablations on the amount of data required for \nnr{}, and show that for target languages with sufficient representation quality after pretraining, \nnr{} produces viable rankings with as little as 25 examples. 

\section{Related Work}
 
\paragraph{Multilingual Model Analysis} Since their release, there has been growing interest in analyzing pretrained multilingual models, in order to better understand what factors lead to strong cross-lingual performance \cite{philippyCommonUnderstandingContributing2023}. Lexical overlap -- the percentage overlap of subwords between two languages -- is often considered, though its importance is unclear: \citet{piresHowMultilingualMultilingual2019} find no correlation with downstream performance, while \citet{wuBetoBentzBecas2019} find a positive correlation.
Linguistic similarity, calculated using typological databases \cite{wals, skirgardGrambankRevealsImportance2023}, has also been considered. Transfer performance is often better for more similar languages, particularly languages with similar word order \cite{piresHowMultilingualMultilingual2019, kCrossLingualAbilityMultilingual2020, dufterIdentifyingElementsEssential2020, littellURIELLang2vecRepresenting2017}.

Architectural properties have also been shown to correlate with downstream performance. \citet{deshpandeWhenBERTMultilingual2022} show that embedding similarity is correlated with zero-shot performance. Similarly, \citet{conneauEmergingCrosslingualStructure2020} show that shared parameters in the lower layers of the model are important for multilingual representations, and \citet{mullerFirstAlignThen2021} show that these lower layers focus on aligning representations across languages. In a similar vein, \citet{douWordAlignmentFinetuning2021} find that word alignment performance \cite{ochImprovedStatisticalAlignment2000}, calculated using vector similarity metrics, is strongest when using representations from middle layers of the model. \nnr{} is motivated by these works, and calculates nearest neighbors using representations taken from intermediate layers.

\paragraph{Ranking Source Languages}
LangRank \cite{linChoosingTransferLanguages2019} is a learned model which uses lexical and linguistic features \cite{littellURIELLang2vecRepresenting2017} to rank source languages for four supported tasks, including part-of-speech tagging (POS) and entity linking (EL). Ranking models are trained on examples which consist of a pair of valid source and target datasets, along with a ranking signal taken from a trained cross-lingual model. A bi-directional LSTM CNN-CRF \cite{maEndtoendSequenceLabeling2016} is used for EL, and a character-level LSTM \cite{hochreiterLongShortTermMemory1997} is used for POS. The final ranking model is a gradient boosted decision tree \cite{NIPS2017_6449f44a}. For POS, dataset size and type--token ratio are found to be the most informative signals, while geographic and syntactic distance are most important for EL. In our experiments, we use the provided POS and EL ranking models for the respective experiments. While LangRank focused mainly on bilingual models, \citet{riceUntanglingInfluenceTypology2025} extend it to pretrained multilingual models.

\section{NN-Rank}

\subsection{Method}
The inputs to \nnr{} are an unlabeled dataset in the target language, a pool of unlabeled source datasets to be ranked, and a model which can encode the input text. While we describe the approach using task-specific datasets, \nnr{} can use any unlabeled dataset which represents the source or target languages of interest (see \cref{sec:exp_2u}). For each target subword, the general approach is to find the $k$ nearest neighbor subwords from the source pool and tally the source datasets which yielded these neighboring subwords. An example can be found in \cref{tab:nn_examples}. The ranking is calculated by sorting source datasets by their final count in descending order.
We describe the steps in detail below:

\begin{enumerate}
    \item[1a.] \textbf{Encoding the Target Dataset} Each example from the unlabeled target dataset is input into the model and, for each subword, the representation from layer $\ell$ is extracted. Assuming a model hidden dimension of $h_d$, this step yields a matrix $ T \in \R^{N_{tgt} \times h_d}$, where $N_{tgt}$ is the total number of subword representations extracted from the target dataset, omitting special tokens. We consider all target \textit{tokens}, not target \textit{types}, as the representations will depend on the context.
    
    \item[1b.] \textbf{Encoding each Source Dataset} Let $P = [s_1, \dots\ , s_n]$ define the pool of $n$ source datasets, with each yielding $N_{s_1}, \dots, N_{s_n}$ subwords, respectively, after tokenization. Repeat Step 1a for each source dataset available. This yields a pool of available source subwords, which is represented as a single concatenated matrix $S \in \R^{N_{src} \times h_d}$, where $N_{src} = \sum_{i=0}^{n} N_{s_i}$. Define a function $m: \R^{h_d} \rightarrow \{s_1, \dots, s_n\}$, which maps every source hidden representation to the dataset which yielded it.

    \item[2.] \textbf{Calculating Nearest Neighbors} Create a tally $C$ which maps each source dataset to a count initialized at 0. Iterate over the rows of $T$, i.e., every target subword, and find the $k$ nearest neighbors from the rows of $S$. For each of these top-$k$ neighboring representations, use $m$ to lookup the origin dataset of the source representation and, for each dataset, increment its tally in $C$ by 1. 

    \item[3.] \textbf{Calculating the Ranking} To calculate the final ranking for a given target dataset, sort the source datasets in $C$ by their tally in descending order.

\end{enumerate}

\subsection{Hyperparameters} For all experiments, we use either mBERT or XLM-R as the encoding model and set $h_d = 768$ and $\ell = 8$ \cite{douWordAlignmentFinetuning2021}. We describe the distributions of each $N_{tgt}$ and $N_{s_i}$ in \crefrange{tab:pos_target_token_counts}{tab:ner_source_token_counts}. 
Nearest neighbor calculations are performed using FAISS \cite{douzeFaissLibrary2025}, a vector database that performs efficient retrieval using the inner product.
We set $k$ to 5 for our main results, chosen empirically using the development set performances presented in \cref{tab:pos_nn_dev_results,tab:ner_nn_dev_results}.
\begin{table*}[t]
\begin{adjustbox}{width=\linewidth}
    \tiny

\begin{tabular}{lllrr|rr}
\toprule
 &  &  & \multicolumn{2}{c}{POS \textit{test-all}} & \multicolumn{2}{c}{NER \textit{test-all}} \\
\cmidrule(lr){4-5} \cmidrule(lr){6-7}
Source Split & Task Model & Ranking Method & Avg. Acc.@5 & NDCG@5 & Avg. F1.@5 & NDCG@5 \\
\toprule
\multirow{10}{*}{\textit{train-large}} & \multirow{4}{*}{mBERT} & \nnr-mBERT & \textbf{74.59} & \textbf{62.91} & 60.77 & \textbf{ 47.94} \\
 &  & \nnr-XLM-R & 73.49 & 58.92 & \textbf{60.85} & 46.35 \\
 &  & LangRank & 70.60 & 36.47 & 57.48 & 28.55 \\
 &  & N-LangRank-mBERT & 69.01 & 32.05 & 56.02 & 21.19 \\
 &  & N-LangRank-XLM-R & 68.70 & 33.75 & 55.07 & 20.17 \\ \cmidrule{3-7}
 & \multirow{5}{*}{XLM-R} & \nnr-mBERT & \textbf{78.38} & \textbf{60.81} & 60.74 & 47.16 \\
 &  & \nnr-XLM-R & 77.75 & 60.69 & \textbf{61.55} & \textbf{49.09} \\
 &  & LangRank & 76.00 & 37.32 & 58.14 & 33.02 \\
 &  & N-LangRank-mBERT & 73.90 & 29.96 & 56.26 & 20.41 \\
 &  & N-LangRank-XLM-R & 74.24 & 32.75 & 55.71 & 20.78 \\ \cmidrule{1-7}
\multirow{10}{*}{\textit{train-med}} & \multirow{4}{*}{mBERT} & \nnr-mBERT & \textbf{74.69} & \textbf{55.46} & 60.78 & \textbf{ 47.20} \\
 &  & \nnr-XLM-R & 72.97 & 50.76 & \textbf{60.96} & 45.35 \\
 &  & LangRank & 68.98 & 24.61 & 57.54 & 27.50 \\
 &  & N-LangRank-mBERT & 54.21 & 13.98 & 55.73 & 18.14 \\
 &  & N-LangRank-XLM-R & 57.73 & 17.97 & 54.93 & 18.76 \\ \cmidrule{3-7}
 & \multirow{5}{*}{XLM-R} & \nnr-mBERT & \textbf{78.02} & \textbf{53.65} & 60.86 & 45.07 \\
 &  & \nnr-XLM-R & 76.82 & 52.45 & \textbf{61.75} & \textbf{48.52} \\
 &  & LangRank & 73.18 & 26.22 & 58.35 & 32.07 \\
 &  & N-LangRank-mBERT & 55.87 & 14.39 & 56.24 & 18.73 \\
 &  & N-LangRank-XLM-R & 60.80 & 17.71 & 55.55 & 18.50 \\ \cmidrule{1-7}
\multirow{10}{*}{\textit{train-all}} & \multirow{4}{*}{mBERT} & \nnr-mBERT & \textbf{73.41} & \textbf{44.51} & \textbf{59.61} & \textbf{44.07} \\
 &  & \nnr-XLM-R & 71.46 & 41.07 & 57.97 & 38.88 \\
 &  & LangRank & 58.67 & 8.95 & 57.09 & 25.93 \\
 &  & N-LangRank-mBERT & 55.39 & 12.62 & 53.88 & 17.17 \\
 &  & N-LangRank-XLM-R & 57.32 & 13.43 & 54.69 & 17.71 \\ \cmidrule{3-7}
 & \multirow{5}{*}{XLM-R} & \nnr-mBERT & \textbf{75.67} & \textbf{42.04} & \textbf{58.68} & \textbf{41.36} \\
 &  & \nnr-XLM-R & 74.46 & 39.84 & 57.61 & 39.95 \\
 &  & LangRank & 60.13 & 8.89 & 57.17 & 29.68 \\
 &  & N-LangRank-mBERT & 57.40 & 13.04 & 54.23 & 17.36 \\
 &  & N-LangRank-XLM-R & 59.87 & 13.31 & 55.18 & 17.69 \\ 
 \bottomrule
\end{tabular}
\end{adjustbox}
   
    \caption{Main results. \textit{Task Model} denotes the model which was finetuned and evaluated. \textit{Ranking Method} denotes how the rankings were produced. The model used for hidden representations, in the case of \nnr{} and the model used for the training signal, in the case of N-LangRank, are denoted.}
    \label{tab:main_results_pos_and_ner}
\end{table*}
\subsection{Considerations}
\label{sec:considerations}
Importantly, because all initial counts are set to 0, source datasets whose tokens do not appear as a nearest neighbor cannot be ranked, as the final count would remain 0. In our experiments, however, we find that the majority of source languages have positive count (see \cref{fig:num_unranked_dsets}). Furthermore, we expect \nnr{} to work best for target languages which the model encodes with high-quality representations. For low-resource languages or those with unseen scripts, the ranking performance is likely to suffer. 
However, we note that this may be a minor issue in practice: if a target language is not represented well enough to produce an adequate ranking, model performance for that language on a downstream task will likely be poor regardless of source language selection. Furthermore, this is a benefit in the reverse direction: \nnr{} is unlikely to give a high rank to source datasets that the model cannot represent well -- regardless of features such as linguistic similarity. This is useful as source datasets with poor representations are not likely to lead to good downstream performance. We discuss these trade-offs in \cref{sec:results}. 

\section{Experimental Setup}

We present experiments comparing various methods for ranking source datasets. First, we finetune each task model on every source training dataset, yielding one finetuned model per training dataset. Each finetuned model is then evaluated zero-shot on every target dataset and the performance -- accuracy for POS and F1 for NER -- is recorded. This yields $\textit{num. source datasets} \times \textit{num. target datasets}$ total scores for each pretrained model. Finally, we use these performances to evaluate each ranking method using the metrics described below. A small worked example and additional details are in \cref{app:additional_metrics_info}.
\paragraph{Tasks} We focus on two tasks which allow for large scale evaluation: POS and NER. For POS, we use Universal Dependencies \cite[UD;][]{nivreUniversalDependenciesV22020}. For NER, we use the WikiANN dataset \cite{panCrosslingualNameTagging2017, rahimiMassivelyMultilingualTransfer2019}. 
\paragraph{Languages} For both datasets, the amount of data available for each language varies greatly. Therefore, we consider different language splits\footnote{Split names reflect the amount of data available \textit{per-dataset}. Datasets in \textit{large} have the most data, but the split itself contains the smallest number of languages.} -- \textit{all}, \textit{medium}, and \textit{large} -- based on a minimum threshold number of examples, which defines both our pool of possible source datasets as well as target datasets included in the evaluation. Because we have no maximum threshold, each split builds upon the prior: $large \subset medium \subset all$, which allows for comparison across splits. Splits mark increasing difficulty, with \textit{all} being the hardest; ranking becomes more difficult as the pool becomes larger, and target languages in \textit{all} are more likely to be poorly represented by the model. We only consider languages which are supported by the released LangRank models: source datasets are limited to those in the model index, and target datasets are limited to the languages supported by lang2vec \cite{littellURIELLang2vecRepresenting2017}.

The UD dataset often provides various treebanks for the same language. Any dataset which meets the minimum threshold is included in the pool of source languages to be ranked; therefore our experiments using UD data are not ranking source \textit{languages} but source \textit{datasets}. The same rule applies to the target languages: any treebank which meets the evaluation threshold is included. As NER data is extracted from Wikipedia, there is only one dataset per language. 

For the UD training data, the minimum thresholds are 500, 7500, and 15000 examples for the \textit{all}, \textit{medium}, and \textit{large} splits respectively. For UD evaluation data, the thresholds are 100, 750, and 2000. For NER, the thresholds are set to 1000, 10000, and 15000 for training, and 100, 1000, and 10000 for evaluation. In the most restrictive setting (i.e., \textit{train-large x test-large}), we have 20 unique source languages and 21 target languages for POS (corresponding to 25 source and 25 target datasets), as well as 37 source languages and 34 target languages for NER. In the least restrictive setting (i.e., \textit{train-all x test-all}), we have 51 source languages and 56 target languages for POS (corresponding to 78 source datasets and 118 test datasets), as well as 51 source languages and 72 target languages for NER.  Both tasks include languages from up to 13 different language families, however, the majority are Indo-European. Detailed information on all languages can be found in \crefrange{tab:pos_train_langs}{tab:ner_test_langs}.

\paragraph{Ranking Methods} We consider five different ranking methods in our main experiments. The pretrained LangRank model released by \citet{linChoosingTransferLanguages2019} is used as a baseline. For this model, lexical features are taken from the development set. As LangRank always produces a ranking of all available source languages, for each source language split, we skip any language in the ranking which is not valid for that case. We also skip any source language with the same ISO code as the target. We also consider two LangRank-based models trained from scratch (N-LangRank), using relevance scores calculated from the development set performance of either mBERT or XLM-R. We follow the general experimental setup of \citet{linChoosingTransferLanguages2019} and train a different ranking model for each target language. Training examples are created by considering all available pairs of train and development set datasets -- dependent on the language split -- and excluding any source or target dataset which shares the same ISO code as the target language. For $N$ target datasets, this yields $N$ different models, each of which is used at test time. Lexical features from the target development set are used for inference. 

We also consider two \nnr{} rankings, depending on if hidden representations are taken from mBERT or XLM-R. Source representations are calculated using the training split for each target task, and target representations use the development set. We set a limit of 1000 total input lines, and the source language split determines which datasets are included in the source pool $P$.

\paragraph{Task Models} We consider two task models: the \texttt{base} versions of mBERT and XLM-R, as they show strong zero-shot POS and NER performance.
We omit large language models from this work, as they are often pretrained on a smaller number of languages and may not encode all source or target languages with sufficient quality.

\paragraph{Metrics}
\label{sec:metrics}

We use Normalized Discounted Cumulative Gain \cite[NDCG;][]{jarvelinCumulatedGainbasedEvaluation2002} for ranking evaluation. We follow \citet{linChoosingTransferLanguages2019} and assign a relevance score of $\gamma_{max}$ to the top predicted transfer dataset, $\gamma_{max} - 1$ to the second predicted, and continue until the top-$\gamma_{max}$ source datasets have a relevance score greater than 0. The other source datasets are given a score of 0.

We additionally implement performance-based metrics, \textit{Average Accuracy}$@p$ and \textit{Average F1}$@p$ for POS and NER, respectively. For a given target dataset, we average the accuracy or F1 scores achieved by the task models finetuned on the predicted top-$p$ source datasets. We then average these scores across all target datasets to get the final value for the evaluation split (e.g., \textit{test-all}). $\gamma_{max}$ is set to 10, and $p$ is set to 5 \cite{linChoosingTransferLanguages2019, riceUntanglingInfluenceTypology2025} for all metrics. 

\section{Results}
\label{sec:results}
We present a summary of results in \cref{tab:main_results_pos_and_ner}, where we calculate metrics using test set performances.
Because all rankings are created with development set data, this setting ensures that we do not evaluate on any data used to generate the rankings.
\nnr{} greatly outperforms LangRank in every setting in terms of NDCG, highlighting the strength of rankings generated from model hidden states.
The increase in performance-based metrics shows that these ranking differences also have a practical impact on model performance.
While there is some variance across task and split depending on the pretrained model, we find that mBERT is often a stronger choice for ranking than XLM-R, even when XLM-R is used as the task model. This may be due to the pretraining data of mBERT, which uses Wikipedia for all languages. The similar writing style and domain may lead to better implicit alignments between languages during pretraining. This result aligns with prior work which finds that mBERT representations yield stronger word alignment performance \cite{ebrahimiMeetingNeedsLowResource2023}.

More specific results are available in the appendix: full results detailing all language splits can be found in \cref{tab:main_pos_results_table_full,tab:main_ner_results_table_full}. 

\subsection{Analysis}
In this section, we analyze the main results and focus on \nnr{}-mBERT and the \{\textit{train-all x test-all}\} language split.

\begin{figure}[]
    \centering
\includegraphics[width=1.0\linewidth]{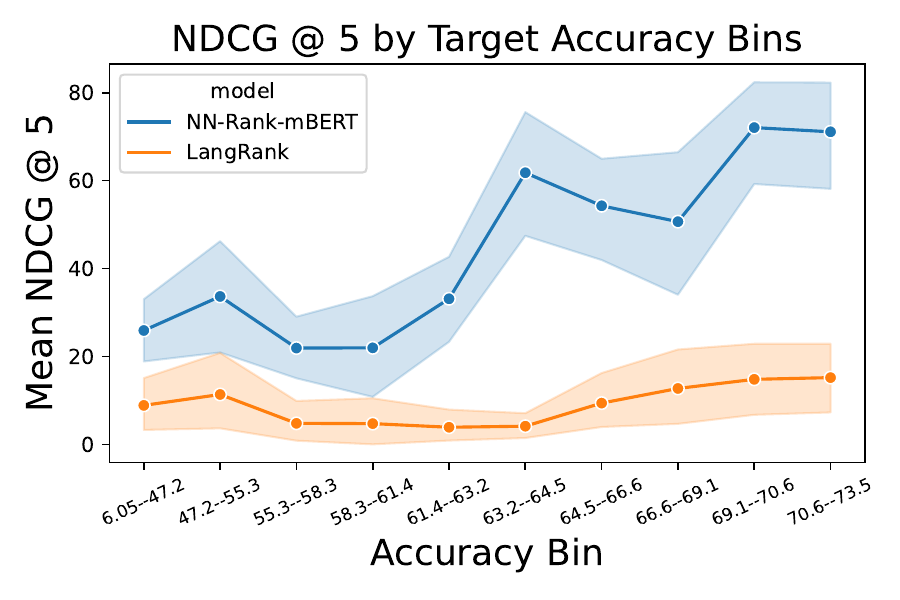}
    \caption{Mean accuracy of a target language over all source languages compared to NDCG for that target language.
    Each bin contains 10 languages, and the y-axis is the average NDCG for each bin.
    Shading represents the 95\% confidence interval for NDCG scores.
    }
    \label{fig:ndcg_by_tgt_acc}
\end{figure}

\subsubsection{Per-Language Performance}
In \cref{fig:ndcg_by_tgt_acc} we present the mean NDCG of target languages, ordered by their representation quality under mBERT. Here, we approximate quality by taking the mean POS accuracy achieved by mBERT on the target language, over all source languages.
\nnr{} is sensitive to quality, where it predicts stronger rankings for languages that tend to be higher accuracy.
LangRank, on the other hand, is invariant to this quality measure.
This illustrates one fault of \nnr{}: it may struggle to find the best source datasets for poorly represented target datasets.
However, for every bin, including the lowest accuracy bin, \nnr{} outperforms LangRank on average.

Indeed \nnr{} outperforms LangRank for almost every target dataset with the exception of 8 datasets covering Korean, Latin, Armenian and Turkish. We might expect mBERT to encode these target datasets with high quality representations -- leading to strong ranking performance -- as they are well represented in the pretraining data.
We find that the reduced ranking performance is due to a Finnish source dataset which achieves top five performance for 7 of the 8 datasets, but is not ranked highly.
This indicates that \nnr{} is not only sensitive to the representation quality of the target dataset, but the source as well.
In cases such as these, where the model cannot properly encode a source dataset, \nnr{} will fail to produce a strong ranking for target languages that the source language transfers well to.

\subsubsection{Analysis of Poor Source Datasets}
\begin{figure}[]
    \centering
    \includegraphics[width=\linewidth]{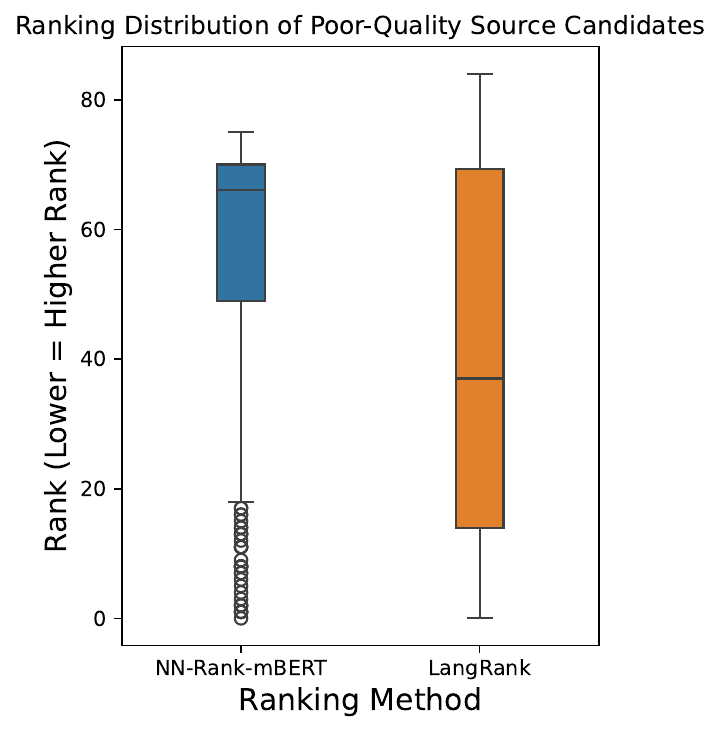}
    \caption{Distribution of ranking position for poor source datasets.
    }
    \label{fig:main_poor_source_cand_distr}
\end{figure}

To further analyze quality, we plot the distribution of ranking position given to \textit{poor} source datasets, which we define as those found in the bottom 15\% of the gold ranking for a given target language. This analysis complements NDCG, which assigns the same relevance score to all source datasets outside of the top ten, by focusing on the worst performing source datasets. We present results in \cref{fig:main_poor_source_cand_distr}, which shows that \nnr{} consistently gives a low rank to poor source datasets while the distribution is flat for LangRank. We also plot the ranking distribution for source datasets with greater than 5\% unknown tokens in \cref{fig:unk_ranking_distrib}.

\section{Experiment 2: General Rankings without Task Datasets}
\label{sec:exp_2u}

While ranking approaches such as LangRank can fall back to language-level features, \nnr{} requires unlabeled data in the target language. Relevant domain-specific data in the target language may be difficult to collect, particularly for data-scarce languages. For this experiment, we ask: \textit{Can \nnr{} produce a high-quality general purpose ranking without task- or domain-specific data in both the source and target languages?} This may be useful in cases where, for example, we wish to create an all-purpose ranking of source languages prior to having access to data in the domain of interest, such as in an online setting. Here, we assume that the only data available is the Bible, a corpus which is currently available for around 1600 languages \cite{mccarthyJohnsHopkinsUniversity2020}. While the Bible is often used for its coverage of languages, it has multiple drawbacks which include biased language, infrequently used vocabulary, and a limited domain -- all of which may impact the performance of \nnr{}.
\begin{table}[]
\begin{adjustbox}{width=\linewidth}
\large


\begin{tabular}{llrr}
\toprule
 &  & \multicolumn{2}{c}{NER \textit{Test-all}} \\
 \cmidrule(lr){3-4} 
Task Model & Ranking Method & F1.@5 & NDCG@5 \\
\midrule
\multirow{3}{*}{mBERT} & \nnr-mBERT & 61.23 & 36.07 \\
 & \nnr XLM-R & 60.08 & 35.76 \\
 & LangRank & 60.38 & 27.29 \\ \cmidrule{2-4} 
\multirow{3}{*}{XLM-R} & \nnr-mBERT & 60.43 & 39.87 \\
 & \nnr XLM-R & 58.70 & 37.67 \\
 & LangRank & 59.69 & 30.79 \\ 
 \bottomrule
\end{tabular}
\end{adjustbox}
   
    \caption{Experiment 2 results. \nnr{} results take both source and target representations from the Bible. For reference, LangRank outputs using lexical features from the task-specific datasets are included.}
    \label{tab:exp2u_main_summary}
\end{table}
\paragraph{Experimental Setup} These experiments largely follow the experimental setup of the main experiment. However, instead of using the unlabeled development set to extract model hidden representations for the target language, we use the associated Bible from the JHU Bible Corpus \cite{mccarthyJohnsHopkinsUniversity2020}. To maximize the similarity between the source and target domains, we also use the Bible to represent the source languages. We omit POS in this experiment, as this general ranking is at the language level, while the UD dataset has multiple train datasets for a single source language. \cref{tab:ner_bible_data_map} describes the Bible used for each language. For reference, we include LangRank results when using task-specific data in the target language (as Bible data does not have the task-specific labels needed to calculate relevance scores). There are 46 source datasets in the \textit{train-all} split, and again 62 target datasets. 

\paragraph{Results} Summary of results can be found in \cref{tab:exp2u_main_summary}, and full results can be found in \cref{tab:exp2u_general_ranking_ner_results_full}. For both task models, the \textit{Avg. F1} is very close between the best \nnr{} ranking and LangRank, with a difference of 0.85 when evaluating with mBERT, and 0.74 for XLM-R. For both cases, using LangRank outperforms rankings generated using XLM-R, but rankings created using mBERT achieve the best performance across all approaches. When considering NDCG scores however, \nnr{} -- using either mBERT or XLM-R to generate representations -- outperforms LangRank. Across both task models, the worst-performing \nnr{} ranking beats LangRank by 8.47 and 6.85 NDCG, respectively. These results show that \nnr{} does not require in-domain data to create a strong ranking for the tasks in our experiments; general purpose \nnr{} rankings are competitive with task-specific LangRank rankings. 
\begin{table}[]
\begin{adjustbox}{width=\linewidth}
\large



\begin{tabular}{llrr}
\toprule
\multicolumn{2}{l}{\textit{POS Tagging Results \{train-all x test-all\}}} \\
\midrule
Task Model & Ranking Method & Acc.@5 & NDCG@5 \\
\toprule
\multirow{3}{*}{mBERT} & \nnr-mBERT & 74.85 & 38.66 \\
 & \nnr-XLM-R & 68.95 & 28.26 \\
 & LangRank & 53.20 & 5.95 \\
 \cmidrule{2-4}
\multirow{3}{*}{XLM-R} & \nnr-mBERT & 78.57 & 38.88 \\
 & \nnr-XLM-R & 73.21 & 26.17 \\
 & LangRank & 57.00 & 4.70 \\
\toprule
\multicolumn{2}{l}{\textit{NER Results \{train-all x test-all\}}} \\
\midrule
Task Model & Ranking Method & F1@5 & NDCG@5 \\
\toprule
\multirow{3}{*}{mBERT} & \nnr-mBERT & 59.95 & 28.74 \\
 & \nnr-XLM-R & 58.77 & 28.72 \\
 & LangRank & 61.27 & 29.18 \\
 \cmidrule{2-4}
\multirow{3}{*}{XLM-R} & \nnr-mBERT & 58.23 & 32.12 \\
 & \nnr-XLM-R & 56.74 & 32.80 \\
 & LangRank & 60.30 & 31.43 \\
\toprule
\end{tabular}
\end{adjustbox}
   
    \caption{Domain mismatch results. Target language representations are taken from the Bible, while source representations are taken from the task-specific datasets. }
    \label{tab:exp3u_main_summary}
\end{table}
\section{Analysis and Ablations}
We conduct three analysis experiments, focused on (1) the impact of domain mismatch, (2) the layer at which model representations are taken, and (3) the number of target subwords used for ranking. We use task-specific data for the layer and target data ablations.

\subsection{Impact of Domain Mismatch}

In the prior experiments, the data used to represent both source and target languages is taken from the same domain. For this analysis, we ask: \textit{How strongly does domain mismatch affect the performance of \nnr{}?}

\paragraph{Experimental Setup}  To simulate realistic domain mismatch, representations for the source datasets are taken from the available training sets for each task while representations for the target datasets are taken from the Bible. For fair comparison, we present results when using LangRank with lexical features from the Bible. Performances, however, are not directly comparable to the main results, as not all train and evaluation languages have a corresponding Bible. For POS, there are 103 target datasets, and 62 target datasets for NER.

\paragraph{Results} We present a summary of results in \cref{tab:exp3u_main_summary}, with full results in \cref{tab:exp3u_bible_pos_results_table_full,tab:exp3u_bible_ner_results_table_full}. For POS, \nnr{} continues to outperform LangRank across both metrics, and using mBERT hidden representations offers the best ranking. For NER, ranking performances are much more mixed. When considering average F1, LangRank is consistently stronger. However, for NDCG, LangRank is only stronger when mBERT is used as the task model. This indicates that, while \nnr{} remains competitive, it is sensitive to domain mismatch. In practice, this mismatch should be avoided by changing the domain of the source languages to match that of the target.

\subsection{Layer Ablation} 

For this ablation, we focus on the difference in ranking performance if we use hidden representations taken from Layer 8 and Layer 0 (the embedding layer). A summary of the results can be found in \cref{tab:main_layer_ablation_results} with full results in \cref{tab:layer_ablation_pos_full,tab:layer_ablation_ner_full}. In practically all cases, performance improves when using the intermediate layer, with large gains in NDCG. Using embedding representations only leads to better performance when measuring average accuracy or average F1, with the maximum difference across both tasks being less than -0.5. This result further highlights the weakness of static features; ranking quality improves as we move away from the embedding layer -- the closest model-based feature to lexical overlap -- and allow the model the encode and align the input sequences.

\begin{table}[]

\begin{adjustbox}{width=\linewidth}
\large


\begin{tabular}{llrr}
\toprule
\multicolumn{2}{l}{\textit{POS Tagging Results \{train-all x test-all\}}} \\ 
 \midrule
Task Model & Ranking Model & $\Delta$ Acc.@5 & $\Delta$ NDCG@5 \\ 
\toprule
\multirow{2}{*}{mBERT} & mBERT & 3.27 & 12.01 \\
 & XLM-R & 1.17 & 7.57 \\
 \cmidrule{2-4}
\multirow{2}{*}{XLM-R} & mBERT & 3.36 & 12.27 \\
 & XLM-R & 1.55 & 7.68 \\ 
 \toprule
\multicolumn{2}{l}{\textit{NER Results \{train-all x test-all\}}} \\
\midrule
Eval Model & Ranking Model & $\Delta$ F1@5 & $\Delta$ NDCG@5 \\ 
\toprule
\multirow{2}{*}{mBERT} & mBERT & 2.07 & 10.53 \\
 & XLM-R & 0.55 & 7.78 \\
 \cmidrule{2-4}
\multirow{2}{*}{XLM-R} & mBERT & 1.24 & 5.61 \\
 & XLM-R & -0.26 & 4.02 \\
 \bottomrule
\end{tabular}
\end{adjustbox}
   
    \caption{Layer ablation results. Positive scores indicate higher performance when using Layer 8.}
    \label{tab:main_layer_ablation_results}
\end{table}

\subsection{Target Data Ablations} \label{sec:target_data_ablations} Here, we are interested in understanding how the amount of target language data available influences ranking quality. We focus solely on \nnr{} performance using mBERT -- as both the ranking and task model -- for POS only. We discuss the limitations of these experiments in \cref{sec:limitations}.
\begin{figure}[!ht]
    \centering
    \includegraphics[width=0.75\linewidth]{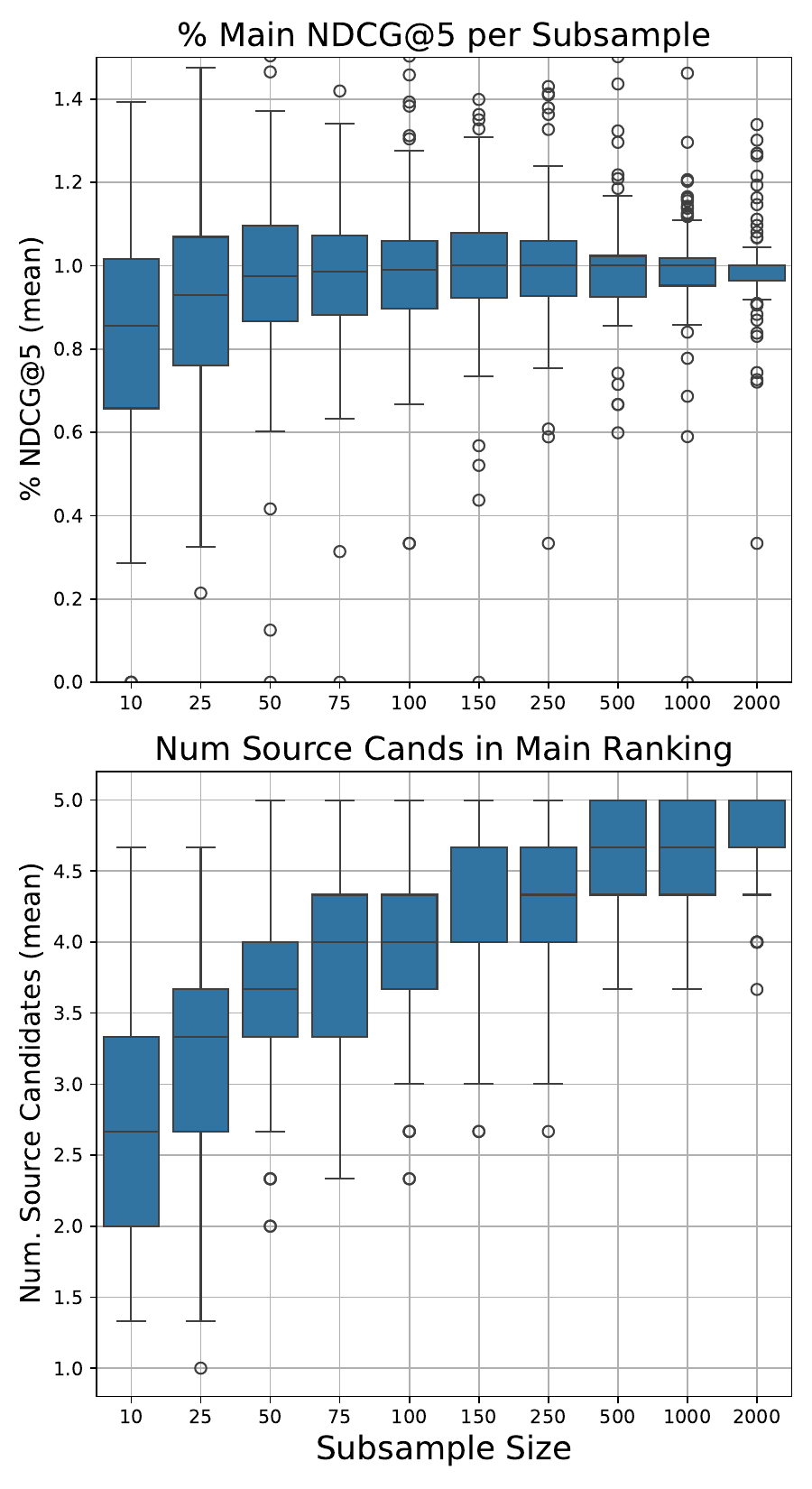}
    \caption{Data ablation results. Compares the performance using each subsample size to the \textit{Main} results. The lower subplot shows the number of overlapping source datasets between the top five predicted datasets from the subsample ranking and main ranking.}
    \label{fig:data_ablation_sub_to_main}
\end{figure}
\begin{figure*}[ht]
    \centering
    \includegraphics[width=0.9\linewidth]{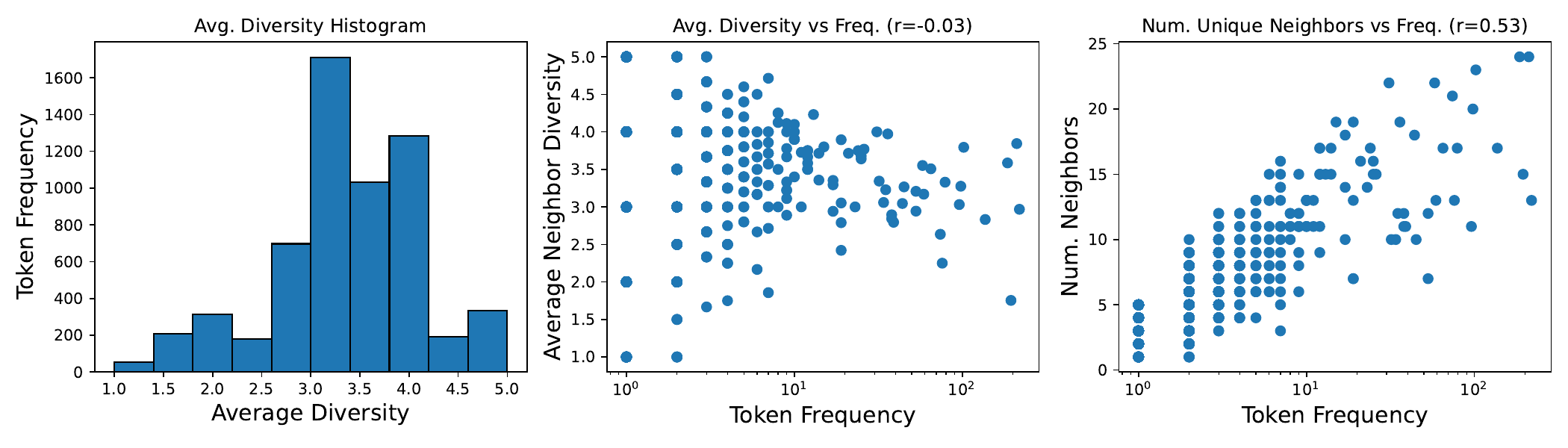}
    \caption{Case study results for French. Token frequency plotted using log scale. Pearson's \textit{r} is used.}
    \label{fig:french_cs}
\end{figure*}
\paragraph{Experimental Setup} For this experiment, we subsample the number of \textit{target hidden representations} used, i.e., the rows of $T$, and consider sample sizes from between 10 and 2000. 
We take three different samples for each size, calculate a ranking from each, and consider the mean NDCG and average accuracy across each sample. 

\paragraph{Results} 
\label{sec:target_data_ablation_results}
Full data ablation results can be found in \cref{tab:data_ablation_pos_full}. The average accuracy is surprisingly stable across all sample sizes, while the NDCG, on average, increases consistently with larger samples. This indicates that strong source datasets are found quickly, and the ranking quality improves with more target tokens. We present the distribution of performances in \cref{fig:data_ablation_sub_to_main}, which compares the ranking performance using each subsample to the \textit{main} ranking, which uses all available data (limited to 1000 input sequences); we also analyze the difference between each consecutive subsample in \cref{fig:data_ablation_sub_to_sub}. The former shows that, even with the smallest subsample, we recover over half of the source datasets predicted in the \textit{main} top five, when considering the median over all target languages. Similarly, when considering how the NDCG achieved using the subsamples compares to when we use all target data, the median using a sample size of 10 hidden representations is 85.6\%. For subsamples of 25 and 50, the median is 92.8\% and 97.4\%, respectively.

To understand how a ranking can be created with so few target representations (see \cref{sec:considerations}), we conduct a case study on French. The goal is to examine how each target subword contributes to the tally. We quantify the contribution using two measures: the first is \textit{diversity}, defined as the number of unique source datasets found in the top five nearest neighbors of a target token, averaged across each instance. The second is the \textit{total count of unique source datasets} in the nearest neighbors, summed across each instance. Both measures are necessary: a token may have high diversity (e.g., all five neighbors are from different datasets), but yield the same set of five datasets in each instance (low total count). Further details can be found in \cref{sec:french_and_wolof_app}, with an example in \cref{tab:nn_examples}. 

We present results in \cref{fig:french_cs}. We see that the majority of tokens are diverse, yielding at least three different source datasets in their neighbors on average. The total count of unique source datasets is large as well: across all frequencies, we find tokens which achieve the maximum possible total count (a token which appears only once can increase the tally of at most five source datasets). Interestingly, we also see that diversity is not correlated with frequency: while diverse tokens are more frequently found, frequency is not necessary for a token to be diverse. These findings explain how rankings are created with very few target subwords: the majority of target tokens increase the tally of at least three different source datasets, and these source datasets change depending on the context of the target token.


\section{Conclusion}

In this work, we present \nnr{}, a data-driven approach to ranking source languages for cross-lingual transfer, which leverages model hidden representations. The approach outperforms LangRank and remains competitive when using out-of-domain data. Our results highlight a critical weakness of prior approaches to ranking and general multilingual analysis: the use of \textit{static} features, such as language-level or lexical features. These features fail to account for the \textit{model itself} and, as such, cannot be used to sufficiently explain cross-lingual performance. We hope that our findings help motivate future work on better understanding the cross-lingual properties of multilingual models. 

\clearpage

\section*{Limitations}
\label{sec:limitations}

\paragraph{Model Selection} Decoder-only multilingual large language models (LLMs) such as BLOOM \cite{workshopBLOOM176BParameterOpenAccess2023} have also been proposed and have shown strong cross-lingual performance. In this work, we choose to focus solely on encoder-only models. This choice was guided by the multilingual LLMs available at the time of experimentation, which are less multilingual than their encoder-only counterparts. For example, BLOOM only covers 46 languages and does not include particular high-resource languages that we expect may be helpful source languages, like German. Furthermore, a method to rank source languages should ideally be lightweight and quick to run -- for example, LangRank does not require any GPU resources, which makes it more applicable to a real-world setting.

We believe that using \nnr{} to evaluate and analyze multilingual LLMs is a promising direction for better understanding the dynamics of multilingual pretraining. However, due to the size of these models, the differences in pretraining procedures and objectives, as well as the various ways in which they can be used to achieve cross-lingual transfer, we believe that these experiments are better suited to dedicated future work. 

\paragraph{Experimental Setup} In this work, we solely experiment with single-source cross-lingual transfer, as this allows for more focused analysis on the effectiveness of specific languages as the source language. We chose this setting to allow comparison to prior analysis works which also focus on a single-source setting \cite{wuBetoBentzBecas2019,piresHowMultilingualMultilingual2019, kCrossLingualAbilityMultilingual2020, dufterIdentifyingElementsEssential2020, devriesMakeBestCrosslingual2022, riceUntanglingInfluenceTypology2025}. Multi-source transfer has been shown to yield better performance \cite{wuSingleMultiSourceCrossLingual2020, adelaniMasakhaNER20Africacentric2022, garcia-ferreroMedMT5OpenSourceMultilingual2024} and, while the usage of NDCG should reflect the quality of the top five datasets, it remains to be shown which ranking method is best for a multi-source training setup.

\paragraph{Target Data Ablations} For this experiment, we subsample the number of target hidden representations available independently across all available target tokens, i.e., sampling directly from $T$ with no constraints. This is an unrealistic setting if we want to simulate the case where we only have, e.g., 10 sentences available in a dataset.
Because representations are taken from Layer 8, the sampled vectors will change depending on the context. As such, 10 tokens sampled from the same input sequence will likely yield different results than 10 tokens each sampled from different contexts. 
For these reasons, results from the target data ablation cannot be extrapolated to cases where we have fewer input sequences than the sample size -- especially for the smaller sample sizes. 

Furthermore, we stress that the performance of this method relies on the quality of model representation for the target languages. While we show that the approach works with as little as 10 input sequences -- considering the mean NDCG --, this is qualified with the assumption that the representations are strong. A high-quality ranking using 10 input sequences should not be expected for, e.g., a low-resource language not contained in the pretraining data of the model.

\section*{Acknowledgments} We would like to thank Enora Rice, Alexis Palmer, and Minh Duc Bui for their helpful discussions and feedback. This work utilized the Blanca condo computing resource at the University of Colorado Boulder. Blanca is jointly funded by computing users and the University of Colorado Boulder.

\bibliography{custom, bibs/all_zotero_dedup}

\clearpage

\appendix
\section*{Appendix}
\section{Additional Metric Details}
\label{app:additional_metrics_info}

We provide a small worked example for our experimental setup and how to calculate \textit{Average F1}$@2$ (here $p$ = 2 is used for simplicity) for XLM-R. Assume we have 3 source NER datasets, associated with English (en), Spanish (es), and French (fr). Assume we have 5 target NER datasets: Czech (cs), Igbo (ig), Irish (ga), Finnish (fi), and German (de). We first finetune XLM-R on each source dataset, yielding 3 finetuned models. Each model is then evaluated zero-shot on every target dataset, yielding 15 total pairs of \textit{(source\_dataset, target\_dataset}) F1 scores.

For each target dataset, the selected ranking method generates an ordering of the 3 source datasets. Assume for German, the predicted ranking is \texttt{[English, French, Spanish]}. We then average the F1 scores for the English-finetuned model and French-finetuned model to get the average F1 for German. This process is repeated for the 4 other target datasets. To get the final \textit{Average F1}$@2$, we average the resultant five averages. 

\paragraph{NDCG} As discussed in \cref{sec:considerations}, \nnr{} may not always provide a complete ordering of all source datasets. As shown in the case study and target data ablations, in practice this is highly unlikely. However, in our evaluation, any unordered source dataset is assigned a rank of infinity, yielding a relevance score of 0 to maintain fair evaluation. This problem could be alleviated by initializing the tally with values that induce a default ordering; this represents one way in which linguistic features could be incorporated into \nnr{}.

\section{Model Training} 
We use established hyperparameters: a batch size of 32, learning rate of 2e-5, and train for 10 epochs \cite{ebrahimiHowAdaptYour2021} and assign labels to the last subword.

\section{Languages}
\label{app:languages}

We use Glottolog \cite{hammarstromGlottologDatabase2022} to obtain the language name and family information for each ISO code. To convert between two-letter and three-letter ISO codes, we use the map provided by LangRank.

\section{Case Study: French and Wolof}
\label{sec:french_and_wolof_app}

For this analysis we use \textit{token diversity}, defined as the number of unique source datasets found among the top five nearest neighbors of a given target token, averaged across every instance of the token. Therefore, the lower bound of target diversity is 0, and the upper bound is 5. We also measure the total number of unique source datasets discovered in the top five neighbors of a specific target token -- summed across all instances. The lower bound of this value is 0, and the upper bound is 5 times the number of occurrences of the token (a token which appears once can maximally have 5 total unique neighbors, while a token that appears twice can have 10). Both metrics are required for a complete picture. A specific token may have low diversity (e.g., the nearest neighbors all come from the same source dataset), but a large number of unique source datasets (e.g., the source datasets of the neighbors change depending on the token context). Conversely, a token may have high diversity (e.g., all 5 neighbors come from a different source dataset), but a low number of unique source datasets (e.g., every instance of the token yields the same 5 source datasets). Counts are calculated across the three samples of size 2000.

\begin{figure}[H]
    \centering
    \includegraphics[width=0.75\linewidth]{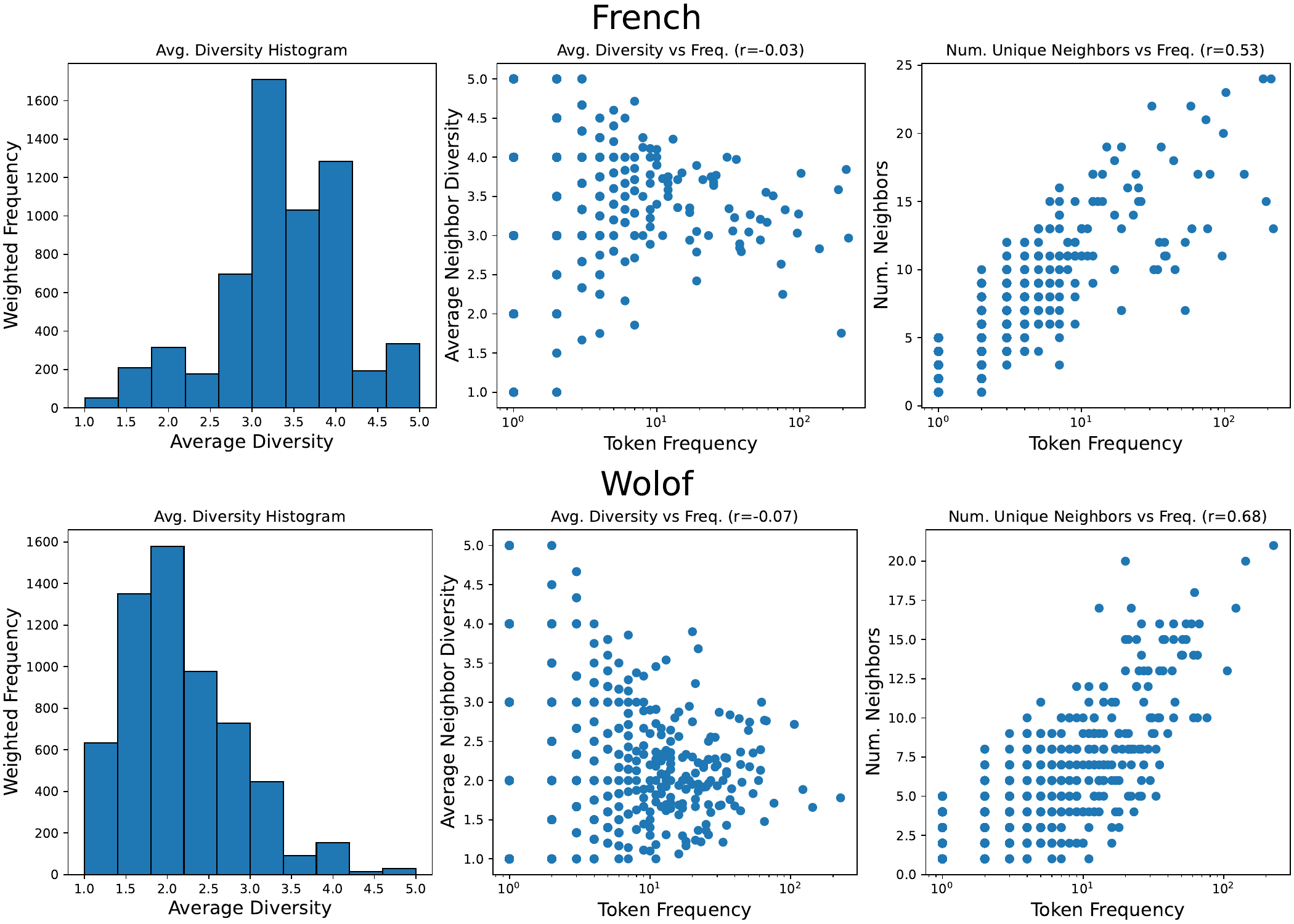}
    \caption{Case study results for French. Token frequency plotted using log scale. Pearson's \textit{r} is used. We also include Wolof, as it is a low-resource and unseen language. Details provided in \cref{sec:french_and_wolof_app}.}
    \label{fig:app_french_wolof}
\end{figure}

French is a relatively high-resource language which is very closely related to English. In addition to the main case study, here we also include Wolof, a low-resource language not contained in the pretraining data of mBERT. Results can be found in \cref{fig:app_french_wolof}. Wolof tokens are less diverse, however the majority still yield on average two different source datasets. There is still no correlation between token frequency and diversity, and similar to French, Wolof target tokens still yield a large number of unique neighbor datasets.

\clearpage

\onecolumn

\section{Figures}

\begin{figure}[H]
    \centering
    \includegraphics[width=0.75\linewidth]{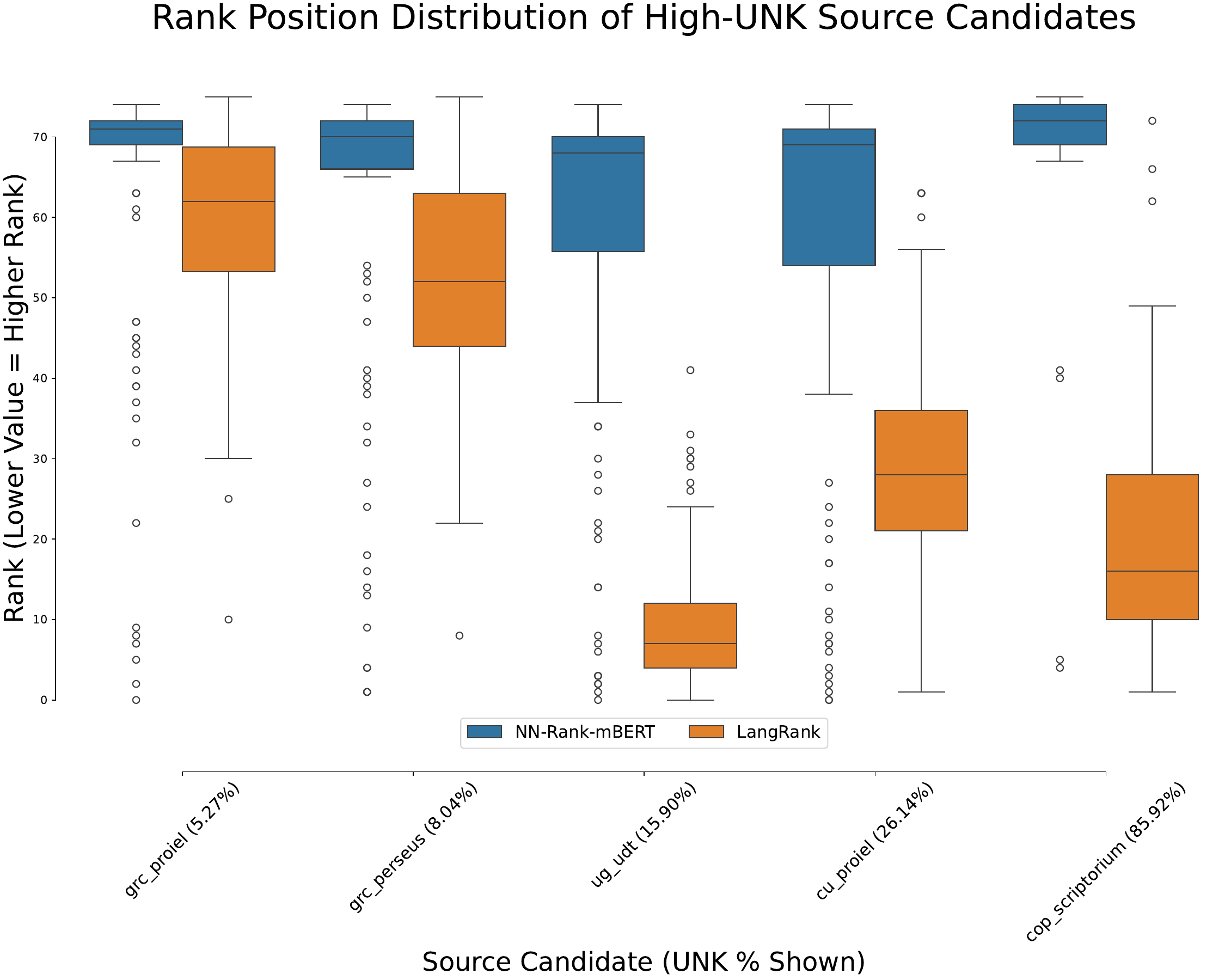}
    \caption{Distribution of ranking position for source datasets with high unknown token percentage. A rank position of 0 is used to mark the top-ranked candidate; in the figure, a lower value signifies that the ranking method gave the source candidate a higher rank. We consider the source datasets with greater than 5\% UNK tokens, using the mBERT tokenizer.}
    \label{fig:unk_ranking_distrib}
\end{figure}

\clearpage

\begin{figure}[!ht]
    \centering
    \includegraphics[width=0.7\linewidth]{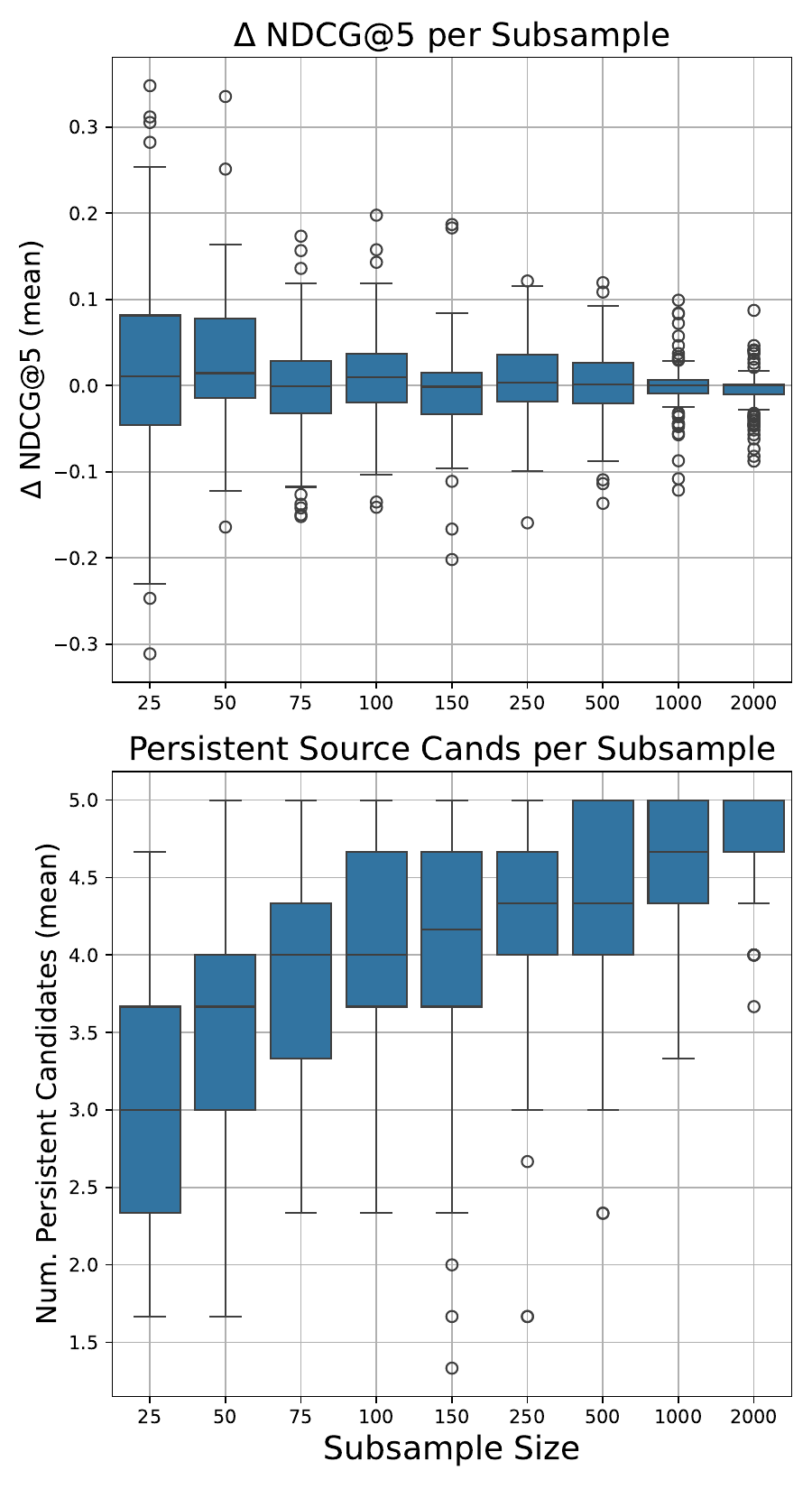}
    \caption{Data ablation results - Performance from one subsample to the next, omitting the first subsample of size 10. $\Delta$NDCG refers to the change in NDCG per target language, from one subsample to the next. \textit{Persistent Source Candidates} refers to the number of source datasets found in the top five predicted datasets of the subsample which were also in the top five predicted datasets of the previous subsample.}
    \label{fig:data_ablation_sub_to_sub}
\end{figure}

\begin{figure}[H]
    \centering
    \includegraphics[width=0.9\linewidth]{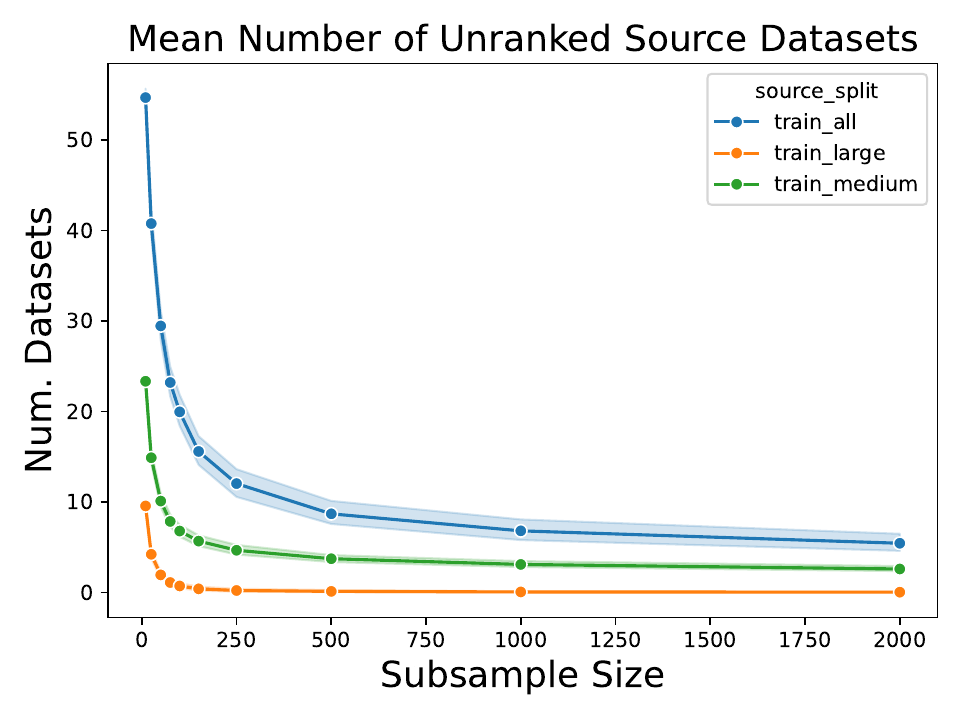}
    \caption{Number of unranked (i.e. zero count) source datasets per target language subsample. Plot shows that with only a small number of target tokens, the majority of source datasets get included in the ranking tally. Subsamples are calculated at the token level, and are averaged across all target datasets in the \textit{test-all} split.}
    \label{fig:num_unranked_dsets}
\end{figure}

\clearpage

\section{Nearest Neighbor Examples}

\begin{table}[H]
    \begin{adjustbox}{width=\linewidth}
        \begin{tabular}{lllll}
\toprule
Target Dataset & Target Token & Top-5 Neighbors & Source File & Source Treebank  \\
\midrule
 \multirow{42}{*}{fr\_gsd-ud-dev} & \multirow{5}{*}{{[}`le'{]}} & {[}`el'{]} & ca\_ancora-ud-train & UD\_Catalan-AnCora \\
 &  & {[}`el'{]} & es\_ancora-ud-train & UD\_Spanish-AnCora \\
 &  & {[}`il'{]} & it\_isdt-ud-train & UD\_Italian-ISDT \\
 &  & {[}`il'{]} & it\_isdt-ud-train & UD\_Italian-ISDT \\
 &  & {[}`il'{]} & it\_isdt-ud-train & UD\_Italian-ISDT \\
 \cmidrule{2-5}
 & \multirow{5}{*}{{[}`.'{]}} & {[}`.'{]} & es\_ancora-ud-train & UD\_Spanish-AnCora \\
 &  & {[}`.'{]} & es\_gsd-ud-train & UD\_Spanish-GSD \\
 &  & {[}`.'{]} & it\_isdt-ud-train & UD\_Italian-ISDT \\
 &  & {[}`.'{]} & ca\_ancora-ud-train & UD\_Catalan-AnCora \\
 &  & {[}`.'{]} & it\_isdt-ud-train & UD\_Italian-ISDT \\
  \cmidrule{2-5}
 & \multirow{5}{*}{{[}`française'{]}} & {[}`civil'{]} & es\_ancora-ud-train & UD\_Spanish-AnCora \\
 &  & {[}\textit{Al Arabiya*}{]} & ar\_padt-ud-train & UD\_Arabic-PADT \\
 &  & {[}`Wereldoorlog'{]} & nl\_lassysmall-ud-train & UD\_Dutch-LassySmall \\
 &  & {[}`1918'{]} & ar\_padt-ud-train & UD\_Arabic-PADT \\
 &  & {[}`Wereldoorlog'{]} & nl\_lassysmall-ud-train & UD\_Dutch-LassySmall \\
  \cmidrule{2-5}
 & \multirow{5}{*}{{[}`reliant'{]}} & {[}`liga'{]} & pt\_gsd-ud-train & UD\_Portuguese-GSD \\
 &  & {[}`from'{]} & en\_ewt-ud-train & UD\_English-EWT \\
 &  & {[}`entre'{]} & es\_gsd-ud-train & UD\_Spanish-GSD \\
 &  & {[}`des'{]} & ca\_ancora-ud-train & UD\_Catalan-AnCora \\
 &  & {[}`\#\#ndo'{]} & pt\_gsd-ud-train & UD\_Portuguese-GSD \\
  \cmidrule{2-5}
 & \multirow{5}{*}{Maria} & {[}`Maria'{]} & it\_isdt-ud-train & UD\_Italian-ISDT \\
 &  & {[}`Maria'{]} & ro\_nonstandard-ud-train & UD\_Romanian-Nonstandard \\
 &  & {[}`Maria'{]} & ro\_nonstandard-ud-train & UD\_Romanian-Nonstandard \\
 &  & {[}`Maria'{]} & pt\_gsd-ud-train & UD\_Portuguese-GSD \\
 &  & {[}`Maria'{]} & it\_isdt-ud-train & UD\_Italian-ISDT \\
  \cmidrule{2-5}
 & \multirow{5}{*}{{[}`.'{]}} & {[}`.'{]} & it\_isdt-ud-train & UD\_Italian-ISDT \\
 &  & {[}`.'{]} & sl\_ssj-ud-train & UD\_Slovenian-SSJ \\
 &  & {[}`.'{]} & it\_isdt-ud-train & UD\_Italian-ISDT \\
 &  & {[}`.'{]} & ru\_syntagrus-ud-train & UD\_Russian-SynTagRus \\
 &  & {[}`.'{]} & cs\_cac-ud-train & UD\_Czech-CAC \\
  \cmidrule{2-5}
 & \multirow{5}{*}{{[}`\#\#ttes'{]}} & {[}`\#\#eras'{]} & pt\_gsd-ud-train & UD\_Portuguese-GSD \\
 &  & {[}`\#\#der'{]} & no\_nynorsk-ud-train & UD\_Norwegian-Nynorsk \\
 &  & {[}`\#\#s'{]} & pt\_gsd-ud-train & UD\_Portuguese-GSD \\
 &  & {[}`\#\#os'{]} & pt\_gsd-ud-train & UD\_Portuguese-GSD \\
 &  & {[}`\#\#ers'{]} & nl\_lassysmall-ud-train & UD\_Dutch-LassySmall \\
 \bottomrule
\end{tabular}
    \end{adjustbox}
    \caption{Example of nearest neighbors for French tokens. Arabic tokens are transliterated for the table and marked with an (*). Tokens and neighbors are calculated using mBERT representations. To calculate the ranking, we tally the number of occurrences of each treebank or source dataset found in the right-most column, and sort them in decreasing order. If we assume that these are the only two instances of the period token, we would calculate the diversity to be \textit{avg(}4,4) = 4 and the total number of unique source datasets to be 7 (see \cref{sec:target_data_ablation_results}).}
    \label{tab:nn_examples}
\end{table}

\clearpage

\section{Full Results Tables: Main Results}

\begin{table}[H]
\begin{adjustbox}{width=\linewidth}
\begin{tabular}{lllll|ll|ll}
\toprule
 &  &  & \multicolumn{2}{c}{\textit{test-large}} & \multicolumn{2}{c}{\textit{test-medium}} & \multicolumn{2}{c}{\textit{test-all}}  \\
Source Split& Eval Model & Ranking Method  & Acc.@5 & NDCG@5 & Acc.@5 & NDCG@5 & Acc.@5 & NDCG@5 \\
 \midrule
\multirow{10}{*}{\textit{train-large}} & \multirow{4}{*}{mBERT} & NN-Rank-mBERT &  \textbf{82.00} & \textbf{69.11} & \textbf{77.18} & \textbf{64.28} & \textbf{74.59 }& \textbf{62.91} \\
 &  & NN-Rank-XLM-R &  80.43 & 65.99 & 76.17 & 60.92 & 73.49 & 58.92 \\
 &  & LangRank &  77.54 & 40.90 & 73.80 & 40.52 & 70.60 & 36.47  \\
 &  & N-LangRank-mBERT &  77.23 & 38.00 & 71.66 & 31.93 & 69.01 & 32.05 \\
  &  & N-LangRank-XLM-R & 75.92 & 36.01 & 72.05 & 35.20 & 68.70 & 33.75 \\
\cmidrule{3-9}
 & \multirow{5}{*}{XLM-R} & NN-Rank-mBERT & \textbf{84.09} & 65.47 & \textbf{81.10} & 62.70 & \textbf{78.38} & \textbf{60.81} \\
 &  & NN-Rank-XLM-R &  83.11 & \textbf{67.89} & 80.51 & \textbf{62.80} & 77.75 & 60.69 \\
 &  & LangRank & 82.07 & 42.55 & 79.22 & 41.65 & 76.00 & 37.32  \\
 &  & N-LangRank-mBERT &  80.73 & 36.97 & 76.75 & 31.84 & 73.90 & 29.96  \\
  &  & N-LangRank-XLM-R &  80.03 & 30.97 & 77.48 & 33.78 & 74.24 & 32.75 \\
 \midrule
 \multirow{10}{*}{\textit{train-medium}} & \multirow{4}{*}{mBERT} & NN-Rank-mBERT &  \textbf{82.56} & \textbf{63.45} & 77.44 & \textbf{57.76} & \textbf{74.69} & \textbf{55.46}  \\
 &  & NN-Rank-XLM-R &  80.68 & 61.55 & \textbf{75.72} & 53.98 & 72.97 & 50.76  \\
 &  & LangRank & 75.69 & 29.20 & 72.32 & 28.16 & 68.98 & 24.61  \\
 &  & N-LangRank-mBERT & 62.98 & 20.49 & 56.02 & 14.41 & 54.21 & 13.98 \\
  &  & N-LangRank-XLM-R & 65.95 & 21.72 & 61.38 & 19.33 & 57.73 & 17.97  \\
\cmidrule{3-9}
 & \multirow{5}{*}{XLM-R} & NN-Rank-mBERT &  \textbf{84.42} & 61.43 & \textbf{80.86} & 55.86 & \textbf{78.02} & \textbf{53.65}  \\
 &  & NN-Rank-XLM-R &  82.94 & \textbf{64.00} & 79.59 & \textbf{56.81} & 76.82 & 52.45  \\
 &  & LangRank & 78.25 & 30.26 & 76.42 & 29.55 & 73.18 & 26.22\\
 &  & N-LangRank-mBERT &  63.92 & 19.27 & 58.02 & 15.29 & 55.87 & 14.39 \\
  &  & N-LangRank-XLM-R & 68.46 & 20.94 & 64.46 & 19.13 & 60.80 & 17.71  \\
 \midrule
 \multirow{10}{*}{\textit{train-all}} & \multirow{4}{*}{mBERT} & NN-Rank-mBERT & \textbf{81.46} &\textbf{ 54.51} & \textbf{75.92} & \textbf{47.12} & \textbf{73.41} & \textbf{44.51}  \\
 &  & NN-Rank-XLM-R &  79.94 & 51.79 & 74.19 & 43.68 & 71.46 & 41.07  \\
 &  & LangRank &  63.12 & 10.84 & 61.06 & 10.19 & 58.67 & 8.95  \\
 &  & N-LangRank-mBERT &  65.29 & 19.92 & 57.64 & 13.38 & 55.39 & 12.62 \\
  &  & N-LangRank-XLM-R & 66.81 & 15.87 & 60.90 & 14.17 & 57.32 & 13.43  \\
\cmidrule{3-9}
 & \multirow{5}{*}{XLM-R} & NN-Rank-mBERT & \textbf{82.63} & \textbf{51.44} & \textbf{78.07} & \textbf{45.20} & \textbf{75.67} & \textbf{42.04}  \\
 &  & NN-Rank-XLM-R &  81.22 & 49.86 & 77.20 & 43.53 & 74.46 & 39.84 \\
 &  & LangRank & 63.49 & 10.47 & 62.79 & 10.76 & 60.13 & 8.89  \\
 &  & N-LangRank-mBERT &  66.08 & 16.40 & 59.88 & 14.42 & 57.40 & 13.04 \\
  &  & N-LangRank-XLM-R &  68.83 & 15.45 & 63.37 & 13.99 & 59.87 & 13.31 \\
 \midrule
\end{tabular}
\end{adjustbox}
   
    \caption{Main POS results -- Full table highlighting ranking performances for all language split combinations.}
    \label{tab:main_pos_results_table_full}
\end{table}

\clearpage

\begin{table}[H]
\begin{adjustbox}{width=\linewidth}
\begin{tabular}{lllll|ll|ll}
\toprule
 &  &  & \multicolumn{2}{c}{\textit{test-large}} & \multicolumn{2}{c}{\textit{test-medium}} & \multicolumn{2}{c}{\textit{test-all}} \\
Source Split& Evaluation Model & Ranking Method  & F1.@5 & NDCG@5 & F1.@5 & NDCG@5 & F1.@5 & NDCG@5   \\
 \midrule
\multirow{10}{*}{\textit{train-large}} & \multirow{4}{*}{mBERT} & NN-Rank-mBERT & 69.41 & 53.98 & 67.17 & 50.95 & 60.77 & 47.94  \\
 &  & NN-Rank-XLM-R &  69.29 & 51.88 & 67.10 & 49.15 & 60.85 & 46.35 \\
 &  & LangRank &  65.82 & 29.76 & 64.01 & 30.74 & 57.48 & 28.55  \\
 &  & N-LangRank-mBERT & 64.51 & 22.87 & 62.21 & 22.34 & 56.02 & 21.19  \\
  &  & N-LangRank-XLM-R & 63.92 & 22.94 & 61.81 & 22.41 & 55.07 & 20.17 \\
\cmidrule{3-9}
 & \multirow{5}{*}{XLM-R} & NN-Rank-mBERT & 68.29 & 61.21 & 66.35 & 53.69 & 60.74 & 47.16  \\
 &  & NN-Rank-XLM-R &  68.31 & 60.09 & 66.43 & 53.80 & 61.55 & 49.09 \\
 &  & LangRank & 64.03 & 39.33 & 62.72 & 36.95 & 58.14 & 33.02 \\
 &  & N-LangRank-mBERT &  62.53 & 24.23 & 60.53 & 21.34 & 56.26 & 20.41  \\
  &  & N-LangRank-XLM-R &  62.42 & 25.83 & 60.70 & 23.18 & 55.71 & 20.78\\
 \midrule
 \multirow{10}{*}{\textit{train-medium}} & \multirow{4}{*}{mBERT} & NN-Rank-mBERT &  69.56 & 53.20 & 67.12 & 49.65 & 60.78 & 47.20  \\
 &  & NN-Rank-XLM-R &  69.37 & 52.12 & 67.21 & 48.95 & 60.96 & 45.35 \\
 &  & LangRank & 66.01 & 27.56 & 64.14 & 28.59 & 57.54 & 27.50 \\
 &  & N-LangRank-mBERT &  63.78 & 20.37 & 61.80 & 20.19 & 55.73 & 18.14 \\
  &  & N-LangRank-XLM-R &  63.98 & 22.18 & 61.43 & 21.52 & 54.93 & 18.76 \\
\cmidrule{3-9}
 & \multirow{5}{*}{XLM-R} & NN-Rank-mBERT & 68.20 & 57.56 & 66.28 & 50.41 & 60.86 & 45.07 \\
 &  & NN-Rank-XLM-R &  68.19 & 58.37 & 66.50 & 52.68 & 61.75 & 48.52  \\
 &  & LangRank & 64.06 & 36.99 & 62.84 & 34.39 & 58.35 & 32.07 \\
 &  & N-LangRank-mBERT & 61.98 & 21.47 & 60.39 & 19.52 & 56.24 & 18.73 \\
  &  & N-LangRank-XLM-R & 62.43 & 25.41 & 60.41 & 21.91 & 55.55 & 18.50 \\
 \midrule
 \multirow{10}{*}{\textit{train-all}} & \multirow{4}{*}{mBERT} & NN-Rank-mBERT & 69.05 & 51.97 & 66.36 & 47.59 & 59.61 & 44.07  \\
 &  & NN-Rank-XLM-R & 68.03 & 47.98 & 64.93 & 44.21 & 57.97 & 38.88  \\
 &  & LangRank &  65.46 & 25.84 & 63.31 & 26.94 & 57.09 & 25.93 \\
 &  & N-LangRank-mBERT & 61.95 & 19.38 & 59.90 & 19.32 & 53.88 & 17.17  \\
  &  & N-LangRank-XLM-R & 64.08 & 21.74 & 61.27 & 20.38 & 54.69 & 17.71 \\
\cmidrule{3-9}
 & \multirow{5}{*}{XLM-R} & NN-Rank-mBERT & 67.33 & 56.04 & 64.87 & 48.08 & 58.68 & 41.36  \\
 &  & NN-Rank-XLM-R & 66.11 & 53.36 & 63.19 & 46.79 & 57.61 & 39.95 \\
 &  & LangRank & 62.95 & 35.44 & 61.32 & 32.32 & 57.17 & 29.68  \\
 &  & N-LangRank-mBERT &  59.95 & 20.10 & 58.27 & 18.25 & 54.23 & 17.36 \\
  &  & N-LangRank-XLM-R & 62.55 & 25.19 & 60.08 & 21.08 & 55.18 & 17.69  \\
 \midrule
\end{tabular}
\end{adjustbox}
   
    \caption{Main NER results -- Full table highlighting ranking performances for all language split combinations.}
    \label{tab:main_ner_results_table_full}
\end{table}
\clearpage

\section{Full Results Tables: Standard Deviations}

\begin{table}[H]
\begin{adjustbox}{width=\linewidth}
\begin{tabular}{lllll|ll|ll}
\toprule
 &  &  & \multicolumn{2}{c}{\textit{test-large}} & \multicolumn{2}{c}{\textit{test-medium}} & \multicolumn{2}{c}{\textit{test-all}}  \\
Source Split& Eval Model & Ranking Method  & Acc.@5 & NDCG@5 & Acc.@5 & NDCG@5 & Acc.@5 & NDCG@5 \\
 \midrule
\multirow{10}{*}{\textit{train-large}} & \multirow{4}{*}{mBERT} & NN-Rank-mBERT &  9.56 & 24.05 & 15.81 & 23.16 & 17.71 & 22.87 \\
 &  & NN-Rank-XLM-R &  11.05 & 29.39 & 16.18 & 26.86 & 18.09 & 26.48\\
 &  & LangRank & 11.40 & 26.21 & 16.22 & 24.49 & 17.41 & 23.75    \\
 &  & N-LangRank-mBERT & 8.76 & 20.18 & 14.99 & 19.78 & 16.45 & 18.40   \\
  &  & N-LangRank-XLM-R & 8.64 & 18.14 & 14.81 & 19.56 & 16.21 & 19.26  \\
\cmidrule{3-9}
 & \multirow{5}{*}{XLM-R} & NN-Rank-mBERT & 9.03 & 27.39 & 12.54 & 24.00 & 15.61 & 23.93 \\
 &  & NN-Rank-XLM-R & 9.57 & 27.35 & 12.55 & 26.19 & 15.40 & 25.74   \\
 &  & LangRank & 8.70 & 27.16 & 13.23 & 23.23 & 15.70 & 23.23   \\
 &  & N-LangRank-mBERT & 8.38 & 21.11 & 12.54 & 19.54 & 14.99 & 19.07   \\
  &  & N-LangRank-XLM-R & 6.82 & 17.65 & 11.73 & 19.79 & 14.25 & 19.20  \\
 \midrule
 \multirow{10}{*}{\textit{train-medium}} & \multirow{4}{*}{mBERT} & NN-Rank-mBERT & 10.23 & 30.07 & 15.48 & 27.63 & 17.56 & 27.60  \\
 &  & NN-Rank-XLM-R & 11.40 & 34.22 & 16.04 & 32.07 & 17.90 & 31.91  \\
 &  & LangRank & 10.64 & 26.63 & 15.52 & 24.39 & 16.78 & 22.64\\
 &  & N-LangRank-mBERT & 14.38 & 18.03 & 14.73 & 17.85 & 14.95 & 17.75  \\
  &  & N-LangRank-XLM-R & 10.46 & 17.35 & 13.98 & 18.05 & 15.22 & 17.65  \\
\cmidrule{3-9}
 & \multirow{5}{*}{XLM-R} & NN-Rank-mBERT &  9.90 & 28.18 & 12.48 & 27.83 & 15.73 & 28.29   \\
 &  & NN-Rank-XLM-R & 10.20 & 30.64 & 12.65 & 30.23 & 15.53 & 30.95  \\
 &  & LangRank & 9.74 & 26.05 & 12.83 & 22.62 & 15.42 & 21.65 \\
 &  & N-LangRank-mBERT & 15.34 & 18.31 & 14.50 & 17.93 & 14.89 & 17.68  \\
  &  & N-LangRank-XLM-R & 11.38 & 16.08 & 13.13 & 17.92 & 14.79 & 17.73   \\
 \midrule
 \multirow{10}{*}{\textit{train-all}} & \multirow{4}{*}{mBERT} & NN-Rank-mBERT & 10.83 & 24.27 & 16.15 & 28.00 & 17.82 & 27.76  \\
 &  & NN-Rank-XLM-R & 11.79 & 22.83 & 16.80 & 27.46 & 18.45 & 28.55   \\
 &  & LangRank & 10.45 & 12.65 & 12.98 & 12.21 & 13.36 & 11.79   \\
 &  & N-LangRank-mBERT & 14.16 & 19.25 & 15.70 & 17.65 & 15.85 & 17.79  \\
  &  & N-LangRank-XLM-R &  10.61 & 16.81 & 14.10 & 17.65 & 15.17 & 16.56  \\
\cmidrule{3-9}
 & \multirow{5}{*}{XLM-R} & NN-Rank-mBERT &  9.71 & 21.15 & 13.85 & 27.84 & 16.05 & 27.62 \\
 &  & NN-Rank-XLM-R & 10.55 & 21.90 & 13.70 & 26.61 & 16.23 & 27.97 \\
 &  & LangRank & 9.54 & 12.20 & 10.98 & 12.48 & 12.16 & 11.79  \\
 &  & N-LangRank-mBERT & 14.92 & 17.69 & 14.84 & 17.32 & 15.35 & 17.03 \\
  &  & N-LangRank-XLM-R & 11.69 & 15.57 & 13.27 & 17.26 & 14.66 & 16.61  \\
 \midrule
\end{tabular}
\end{adjustbox}
   
    \caption{Main POS results -- Full table showing standard deviations across \textit{all target languages} for all language split combinations.}
    \label{tab:main_pos_results_table_std_full}
\end{table}

\clearpage
\begin{table}[H]
\begin{adjustbox}{width=\linewidth}
\begin{tabular}{lllll|ll|ll}
\toprule
 &  &  & \multicolumn{2}{c}{\textit{test-large}} & \multicolumn{2}{c}{\textit{test-medium}} & \multicolumn{2}{c}{\textit{test-all}}  \\
Source Split& Eval Model & Ranking Method  & Acc.@5 & NDCG@5 & Acc.@5 & NDCG@5 & Acc.@5 & NDCG@5 \\
 \midrule
\multirow{10}{*}{\textit{train-large}} & \multirow{4}{*}{mBERT} & NN-Rank-mBERT & 20.19 & 20.50 & 19.30 & 20.62 & 21.81 & 19.60  \\
 &  & NN-Rank-XLM-R & 20.07 & 20.53 & 19.47 & 19.90 & 21.76 & 20.42 \\
 &  & LangRank & 21.81 & 24.59 & 20.42 & 24.04 & 22.53 & 23.01 \\
 &  & N-LangRank-mBERT &  20.78 & 17.34 & 19.55 & 17.39 & 21.58 & 16.47 \\
  &  & N-LangRank-XLM-R & 21.39 & 20.00 & 20.27 & 18.87 & 22.25 & 17.16  \\
\cmidrule{3-9}
 & \multirow{5}{*}{XLM-R} & NN-Rank-mBERT & 17.94 & 20.28 & 16.35 & 23.81 & 17.68 & 25.69 \\
 &  & NN-Rank-XLM-R & 18.13 & 18.57 & 16.82 & 22.24 & 17.02 & 22.29   \\
 &  & LangRank & 19.10 & 24.92 & 17.37 & 24.07 & 17.42 & 23.69  \\
 &  & N-LangRank-mBERT & 18.57 & 17.55 & 16.97 & 17.68 & 16.86 & 17.38  \\
  &  & N-LangRank-XLM-R & 18.58 & 19.99 & 16.73 & 19.09 & 17.04 & 18.06   \\
 \midrule
 \multirow{10}{*}{\textit{train-medium}} & \multirow{4}{*}{mBERT} & NN-Rank-mBERT & 20.19 & 19.64 & 19.31 & 20.59 & 21.70 & 19.71  \\
 &  & NN-Rank-XLM-R & 20.07 & 19.84 & 19.56 & 19.93 & 21.77 & 21.56  \\
 &  & LangRank & 22.03 & 26.26 & 20.68 & 24.48 & 22.77 & 23.60 \\
 &  & N-LangRank-mBERT & 20.60 & 17.34 & 19.53 & 17.31 & 21.48 & 16.14  \\
  &  & N-LangRank-XLM-R & 21.46 & 19.83 & 20.28 & 18.91 & 22.09 & 17.34   \\
\cmidrule{3-9}
 & \multirow{5}{*}{XLM-R} & NN-Rank-mBERT & 17.89 & 20.82 & 16.13 & 23.67 & 17.28 & 24.52  \\
 &  & NN-Rank-XLM-R & 18.15 & 18.92 & 16.59 & 21.68 & 16.75 & 21.60   \\
 &  & LangRank & 19.55 & 26.39 & 17.76 & 25.93 & 17.58 & 24.91 \\
 &  & N-LangRank-mBERT & 18.39 & 17.32 & 16.96 & 17.56 & 16.73 & 16.80  \\
  &  & N-LangRank-XLM-R & 18.61 & 20.19 & 16.69 & 19.91 & 16.84 & 18.80   \\
 \midrule
 \multirow{10}{*}{\textit{train-all}} & \multirow{4}{*}{mBERT} & NN-Rank-mBERT & 20.29 & 20.64 & 19.58 & 21.57 & 21.93 & 20.72   \\
 &  & NN-Rank-XLM-R & 19.99 & 21.84 & 19.89 & 22.07 & 22.23 & 22.72   \\
 &  & LangRank & 21.92 & 25.77 & 20.82 & 24.38 & 22.67 & 23.03  \\
 &  & N-LangRank-mBERT & 20.93 & 17.59 & 19.90 & 17.56 & 21.46 & 16.33  \\
  &  & N-LangRank-XLM-R & 21.52 & 19.77 & 20.34 & 18.46 & 22.09 & 17.06  \\
\cmidrule{3-9}
 & \multirow{5}{*}{XLM-R} & NN-Rank-mBERT & 18.19 & 21.28 & 16.70 & 23.98 & 18.30 & 25.43 \\
 &  & NN-Rank-XLM-R & 18.31 & 22.79 & 17.28 & 23.68 & 17.86 & 24.30  \\
 &  & LangRank & 19.16 & 25.88 & 17.71 & 25.39 & 17.37 & 24.51  \\
 &  & N-LangRank-mBERT & 18.89 & 17.41 & 17.65 & 17.61 & 17.25 & 16.66  \\
  &  & N-LangRank-XLM-R & 18.68 & 20.05 & 16.88 & 19.81 & 16.94 & 18.61  \\
 \midrule
\end{tabular}
\end{adjustbox}
   
    \caption{Main NER results -- Full table showing standard deviations across \textit{all target languages} for all language split combinations.}
    \label{tab:main_ner_results_table_std_full}
\end{table}
\clearpage

\section{Full Results using Bible Data}

\begin{table}[h]
\begin{adjustbox}{width=\linewidth}

\begin{tabular}{lllll|ll|lllll}
\toprule
 &  &  & \multicolumn{2}{c}{\textit{test-large}} & \multicolumn{2}{c}{\textit{test-medium}} & \multicolumn{2}{c}{\textit{test-all}} &  &  &  \\
Source Split& Task Model & Ranking Method  & F1.@5 & NDCG@5 & F1.@5 & NDCG@5 & F1.@5 & NDCG@5 &  &  &  \\
 \midrule
\multirow{6}{*}{\textit{train-large}} & \multirow{3}{*}{mBERT} & NN-Rank-mBERT &  71.84 & 43.55 & 69.47 & 41.95 & 62.34 & 39.15 \\
 &  & NN-Rank XLM-R &  71.33 & 45.26 & 69.17 & 43.27 & 61.98 & 40.27  \\
  &  & LangRank (Rahimi) &  69.87 & 32.03 & 67.54 & 30.76 & 60.55 & 30.11  \\
\cmidrule{3-10}
 & \multirow{3}{*}{XLM-R} & NN-Rank-mBERT &  69.68 & 51.44 & 67.54 & 46.57 & 61.68 & 43.98 \\
 &  & NN-Rank XLM-R &  69.18 & 50.47 & 67.36 & 46.62 & 61.18 & 42.92\\
  &  & LangRank (Rahimi) & 67.54 & 40.92 & 65.70 & 36.85 & 60.39 & 34.37\\
 \midrule
 \multirow{6}{*}{\textit{train-medium}} & \multirow{3}{*}{mBERT} & NN-Rank-mBERT & 71.56 & 42.78 & 69.11 & 40.37 & 61.98 & 36.86 \\
 &  & NN-Rank XLM-R & 71.18 & 44.73 & 68.93 & 42.46 & 61.79 & 39.32\\
  &  & LangRank (Rahimi) & 70.06 & 29.94 & 67.73 & 28.97 & 60.69 & 28.99\\
\cmidrule{3-10}
 & \multirow{3}{*}{XLM-R} & NN-Rank-mBERT &  69.36 & 49.20 & 67.36 & 42.94 & 61.63 & 41.37 \\
 &  & NN-Rank XLM-R &  69.10 & 49.63 & 67.30 & 44.84 & 61.08 & 41.64 \\
  &  & LangRank (Rahimi) &  67.59 & 37.93 & 65.78 & 33.74 & 60.63 & 32.44 \\
 \midrule
 \multirow{6}{*}{\textit{train-all}} & \multirow{3}{*}{mBERT} & NN-Rank-mBERT & 71.62 & 43.06 & 68.85 & 40.57 & 61.23 & 36.07 \\
 &  & NN-Rank XLM-R & 70.14 & 42.24 & 67.80 & 40.61 & 60.08 & 35.76 \\
  &  & LangRank (Rahimi)&  69.37 & 27.86 & 67.09 & 27.53 & 60.38 & 27.29 \\
\cmidrule{3-10}
 & \multirow{3}{*}{XLM-R} & NN-Rank-mBERT & 68.96 & 48.80 & 66.55 & 42.67 & 60.43 & 39.87  \\
 &  & NN-Rank XLM-R & 67.54 & 46.69 & 65.48 & 42.56 & 58.70 & 37.67 \\
  &  & LangRank (Rahimi) &  66.27 & 36.24 & 64.55 & 32.35 & 59.69 & 30.79\\
 \midrule
\end{tabular}
\end{adjustbox}
   
    \caption{Experiment 2 - General rankings with no task dataset - Full NER Results. \nnr{} uses the Bible for the source dataset and target dataset. LangRank takes lexical features from the Rahimi split (same as main results but slightly different source/target pools due to bible availability).}
    \label{tab:exp2u_general_ranking_ner_results_full}
\end{table}

\clearpage

\begin{table}[h]
\begin{adjustbox}{width=\linewidth}

\begin{tabular}{lllll|ll|lllll}
\toprule
 &  &  & \multicolumn{2}{c}{\textit{test-large}} & \multicolumn{2}{c}{\textit{test-medium}} & \multicolumn{2}{c}{\textit{test-all}} &  &  &  \\
Source Split& Task Model & Ranking Method  & Acc.@5 & NDCG@5 & Acc.@5 & NDCG@5 & Acc.@5 & NDCG@5 &  &  &  \\
 \midrule
\multirow{6}{*}{\textit{train-large}} & \multirow{3}{*}{mBERT} & NN-Rank-mBERT & 81.23 & 54.82 & 78.15 & 56.98 & 75.77 & 58.62\\
 &  & NN-Rank-XLM-R & 76.40 & 25.30 & 73.84 & 29.05 & 70.97 & 31.53 \\
 &  & LangRank & 78.67 & 46.35 & 76.22 & 47.35 & 72.86 & 44.42 \\
\cmidrule{3-10}
 & \multirow{3}{*}{XLM-R} & NN-Rank-mBERT & 84.05 & 55.17 & 82.61 & 57.05 & 79.87 & 58.45  \\
 &  & NN-Rank-XLM-R & 80.22 & 25.12 & 78.93 & 26.11 & 76.07 & 29.44 \\
 &  & LangRank &  82.95 & 46.95 & 81.77 & 49.31 & 78.45 & 47.26  \\
 \midrule
 \multirow{6}{*}{\textit{train-medium}} & \multirow{3}{*}{mBERT} & NN-Rank-mBERT & 80.80 & 45.79 & 77.37 & 46.73 & 74.70 & 43.73 \\
 &  & NN-Rank-XLM-R & 74.75 & 18.90 & 72.48 & 21.09 & 70.13 & 23.00 \\
 &  & LangRank & 77.29 & 36.27 & 74.48 & 35.97 & 71.59 & 33.61  \\
\cmidrule{3-10}
 & \multirow{3}{*}{XLM-R} & NN-Rank-mBERT & 83.31 & 47.00 & 81.32 & 47.11 & 78.49 & 44.33  \\
 &  & NN-Rank-XLM-R &  78.18 & 19.66 & 76.97 & 19.69 & 74.52 & 21.57 \\
 &  & LangRank & 79.96 & 35.74 & 78.68 & 36.73 & 75.53 & 35.34  \\
 \midrule
 \multirow{6}{*}{\textit{train-all}} & \multirow{3}{*}{mBERT} & NN-Rank-mBERT &  81.90 & 48.03 & 77.12 & 40.94 & 74.85 & 38.66\\
 &  & NN-Rank-XLM-R & 75.18 & 35.97 & 71.45 & 30.43 & 68.95 & 28.26  \\
 &  & LangRank & 55.72 & 4.92 & 55.68 & 5.82 & 53.20 & 5.95\\
\cmidrule{3-10}
 & \multirow{3}{*}{XLM-R} & NN-Rank-mBERT & 84.34 & 46.67 & 80.83 & 40.42 & 78.57 & 38.88  \\
 &  & NN-Rank-XLM-R & 78.61 & 33.61 & 75.89 & 28.76 & 73.21 & 26.17\\
 &  & LangRank &  59.49 & 4.13 & 59.74 & 4.89 & 57.00 & 4.70  \\
 \midrule
\end{tabular}
\end{adjustbox}
   
    \caption{Domain Mismatch: POS results when no target task data available -- full results. Here, the source datasets are from UD, and the target data is taken from the Bible. }
    \label{tab:exp3u_bible_pos_results_table_full}
\end{table}

\begin{table}[h]
\begin{adjustbox}{width=\linewidth}


\begin{tabular}{lllll|ll|lllll}
\toprule
 &  &  & \multicolumn{2}{c}{\textit{test-large}} & \multicolumn{2}{c}{\textit{test-medium}} & \multicolumn{2}{c}{\textit{test-all}} &  &  &  \\
Source Split& Task Model & Ranking Method  & F1.@5 & NDCG@5 & F1.@5 & NDCG@5 & F1.@5 & NDCG@5 &  &  &  \\
 \midrule
\multirow{6}{*}{\textit{train-large}} & \multirow{3}{*}{mBERT} & NN-Rank-mBERT &71.41 & 41.26 & 69.24 & 40.76 & 61.92 & 36.88  \\
 &  & NN-Rank-XLM-R &  71.06 & 40.01 & 68.77 & 39.07 & 61.49 & 37.25  \\
 &  & LangRank & 71.63 & 38.12 & 69.01 & 35.68 & 62.05 & 35.23 \\
\cmidrule{3-10}
 & \multirow{3}{*}{XLM-R} & NN-Rank-mBERT & 69.46 & 48.19 & 67.16 & 42.61 & 61.06 & 39.93  \\
 &  & NN-Rank-XLM-R &  69.20 & 50.67 & 67.03 & 44.32 & 60.93 & 42.31 \\
 &  & LangRank & 69.78 & 45.87 & 67.44 & 41.35 & 61.87 & 38.26 \\
 \midrule
 \multirow{6}{*}{\textit{train-medium}} & \multirow{3}{*}{mBERT} & NN-Rank-mBERT &  70.92 & 36.06 & 68.64 & 35.97 & 61.75 & 33.40 \\
 &  & NN-Rank-XLM-R &  71.04 & 37.57 & 68.55 & 37.76 & 61.65 & 35.95 \\
 &  & LangRank & 70.99 & 31.94 & 68.51 & 30.21 & 61.89 & 31.31\\
\cmidrule{3-10}
 & \multirow{3}{*}{XLM-R} & NN-Rank-mBERT &  68.78 & 41.70 & 66.71 & 37.81 & 60.88 & 36.67  \\
 &  & NN-Rank-XLM-R & 69.12 & 45.50 & 66.83 & 40.90 & 61.00 & 41.13  \\
 &  & LangRank & 68.89 & 39.18 & 66.81 & 35.01 & 61.62 & 33.19  \\
 \midrule
 \multirow{6}{*}{\textit{train-all}} & \multirow{3}{*}{mBERT} & NN-Rank-mBERT &  69.61 & 31.94 & 67.06 & 31.87 & 59.95 & 28.74  \\
 &  & NN-Rank-XLM-R & 68.99 & 33.38 & 66.43 & 32.64 & 58.77 & 28.72  \\
 &  & LangRank &  70.02 & 29.21 & 67.86 & 28.77 & 61.27 & 29.18  \\
\cmidrule{3-10}
 & \multirow{3}{*}{XLM-R} & NN-Rank-mBERT & 66.18 & 36.64 & 63.92 & 33.70 & 58.23 & 32.12  \\
 &  & NN-Rank-XLM-R & 65.75 & 40.35 & 63.16 & 35.74 & 56.74 & 32.80\\
 &  & LangRank & 67.00 & 36.26 & 65.41 & 33.59 & 60.30 & 31.43 \\
 \midrule
\end{tabular}
\end{adjustbox}
   
    \caption{Domain Mismatch: NER Results when no target task data available -- full results. Here, the source datasets are from the Rahimi splits, and the target data is taken from the Bible. }
    \label{tab:exp3u_bible_ner_results_table_full}
\end{table}

\clearpage

\section{Full Results Tables: Layer Ablations}

\begin{table}[H]
\begin{adjustbox}{width=\linewidth}
\begin{tabular}{lllll|ll|ll}
\toprule
 &  &  & \multicolumn{2}{c}{\textit{test-large}} & \multicolumn{2}{c}{\textit{test-medium}} & \multicolumn{2}{c}{\textit{test-all}}  \\
Source Split& Task Model & Ranking Method  & Acc.@5 & NDCG@5 & Acc.@5 & NDCG@5 & Acc.@5 & NDCG@5 \\
 \midrule
\multirow{4}{*}{\textit{train-large}} & \multirow{2}{*}{mBERT} & NN-Rank-mBERT &  1.38 & 9.29 & 1.36 & 7.92 & 1.58 & 6.87 \\
 &  & NN-Rank XLM-R & 0.06 & 9.47 & 0.13 & 4.50 & 0.40 & 3.63 \\
\cmidrule{3-9}
 & \multirow{2}{*}{XLM-R} & NN-Rank-mBERT & 0.70 & 6.02 & 0.67 & 4.74 & 0.92 & 3.20 \\
 &  & NN-Rank XLM-R &  -0.47 & 9.71 & -0.26 & 4.74 & -0.01 & 3.96 \\
 \midrule
 \multirow{4}{*}{\textit{train-medium}} & \multirow{2}{*}{mBERT} & NN-Rank-mBERT &  1.98 & 6.95 & 2.06 & 9.01 & 2.08 & 9.24  \\
 &  & NN-Rank XLM-R & 0.94 & 6.43 & 0.74 & 4.27 & 0.99 & 3.62 \\
\cmidrule{3-9}
 & \multirow{2}{*}{XLM-R} & NN-Rank-mBERT & 1.64 & 8.18 & 1.47 & 6.26 & 1.59 & 7.90  \\
 &  & NN-Rank XLM-R & 0.47 & 9.69 & 0.38 & 4.39 & 0.75 & 3.90  \\
 \midrule
 \multirow{4}{*}{\textit{train-all}} & \multirow{2}{*}{mBERT} & NN-Rank-mBERT & 1.90 & 10.83 & 3.16 & 11.85 & 3.27 & 12.01  \\
 &  & NN-Rank XLM-R & -0.01 & 6.12 & 1.19 & 7.56 & 1.17 & 7.57  \\
\cmidrule{3-9}
 & \multirow{2}{*}{XLM-R} & NN-Rank-mBERT & 2.42 & 12.16 & 2.95 & 11.28 & 3.36 & 12.27  \\
 &  & NN-Rank XLM-R & 0.51 & 9.08 & 1.56 & 8.24 & 1.55 & 7.68 \\
 \midrule
\end{tabular}
\end{adjustbox}
   
    \caption{Layer ablation: POS results. Difference when using Layer 8 - Layer 0 (positive is better)}
    \label{tab:layer_ablation_pos_full}
\end{table}

\begin{table}[H]
\begin{adjustbox}{width=\linewidth}
\begin{tabular}{lllll|ll|ll}
\toprule
 &  &  & \multicolumn{2}{c}{\textit{test-large}} & \multicolumn{2}{c}{\textit{test-medium}} & \multicolumn{2}{c}{\textit{test-all}}  \\
Source Split& Task Model & Ranking Method  & F1.@5 & NDCG@5 & F1.@5 & NDCG@5 & F1.@5 & NDCG@5 \\
 \midrule
\multirow{4}{*}{\textit{train-large}} & \multirow{2}{*}{mBERT} & NN-Rank-mBERT &  0.24 & 9.59 & 0.70 & 9.32 & 0.65 & 7.65 \\
 &  & NN-Rank XLM-R & 0.91 & 11.98 & 1.06 & 11.86 & 1.41 & 11.60 \\
\cmidrule{3-9}
 & \multirow{2}{*}{XLM-R} & NN-Rank-mBERT & 0.38 & 8.32 & 0.84 & 7.67 & 0.21 & 5.21\\
 &  & NN-Rank XLM-R & 0.96 & 9.94 & 0.94 & 9.18 & 1.21 & 9.24 \\
 \midrule
 \multirow{4}{*}{\textit{train-medium}} & \multirow{2}{*}{mBERT} & NN-Rank-mBERT &  2.70 & 15.86 & 3.19 & 13.84 & 0.51 & 10.21  \\
 &  & NN-Rank XLM-R & 1.47 & 13.71 & 1.87 & 13.18 & 2.08 & 12.87 \\
\cmidrule{3-9}
 & \multirow{2}{*}{XLM-R} & NN-Rank-mBERT & 2.29 & 9.68 & 2.92 & 9.42 & 0.58 & 4.16  \\
 &  & NN-Rank XLM-R &1.58 & 13.39 & 1.54 & 11.44 & 1.41 & 10.74  \\
 \midrule
 \multirow{4}{*}{\textit{train-all}} & \multirow{2}{*}{mBERT} & NN-Rank-mBERT & 3.73 & 13.33 & 3.03 & 10.93 & 2.07 & 10.53  \\
 &  & NN-Rank XLM-R & 0.60 & 8.36 & -0.18 & 6.22 & 0.55 & 7.78 \\
\cmidrule{3-9}
 & \multirow{2}{*}{XLM-R} & NN-Rank-mBERT &3.59 & 11.45 & 3.14 & 7.63 & 1.24 & 5.61  \\
 &  & NN-Rank XLM-R & 0.93 & 5.89 & -0.22 & 4.03 & -0.26 & 4.02 \\
 \midrule
\end{tabular}
\end{adjustbox}
   
    \caption{Layer ablation: NER tagging results. Difference when using Layer 8 - Layer 0 (positive is better).}
    \label{tab:layer_ablation_ner_full}
\end{table}

\clearpage

\section{Data Ablation Results}
\small{
\begin{xltabular}{\linewidth}{llllrrrr}
\toprule
Source Split & Task Model & Ranking Model & Subsample Size & \multicolumn{2}{c}{Avg. Acc@5} & \multicolumn{2}{c}{Avg. NDCG@5} \\
\cmidrule(lr){5-6} \cmidrule(lr){7-8}
&&&& Mean & STD & Mean & STD \\
\toprule
\endfirsthead
\toprule
Source Split & Task Model & Ranking Model & Subsample Size & \multicolumn{2}{c}{Avg. Acc@5} & \multicolumn{2}{c}{Avg. NDCG@5} \\
&&&& Mean & STD & Mean & STD \\
\toprule
\endhead
\toprule
&&&&\multicolumn{2}{c}{\textit{Continued on next page}}  \\
\toprule
\endfoot
\bottomrule
\\ 
\caption{All POS Data Ablation Results} \label{tab:data_ablation_pos_full}
\endlastfoot
\multirow{40}{*}{\textit{large}} & \multirow{20}{*}{mBERT} & \multirow{10}{*}{mBERT} & 10 & 73.56 & 0.0999 & 57.67 & 1.0486 \\
 &  &  & 25 & 74.02 & 0.2591 & 59.48 & 0.5208 \\
 &  &  & 50 & 74.48 & 0.2190 & 62.06 & 0.0999 \\
 &  &  & 75 & 74.35 & 0.2800 & 61.95 & 0.7042 \\
 &  &  & 100 & 74.44 & 0.0677 & 62.90 & 0.5349 \\
 &  &  & 150 & 74.43 & 0.0522 & 62.22 & 0.5954 \\
 &  &  & 250 & 74.46 & 0.0612 & 62.99 & 0.6453 \\
 &  &  & 500 & 74.48 & 0.0679 & 63.26 & 0.3153 \\
 &  &  & 1000 & 74.55 & 0.0324 & 63.16 & 0.3059 \\
 &  &  & 2000 & 75.13 & 0.0314 & 63.05 & 0.1436 \\
  \cmidrule{4-8}
 &  & \multirow{10}{*}{XLM-R} & 10 & 72.64 & 0.2656 & 52.46 & 0.8792 \\
 &  &  & 25 & 72.93 & 0.1687 & 55.80 & 0.7510 \\
 &  &  & 50 & 72.89 & 0.1090 & 55.47 & 0.3483 \\
 &  &  & 75 & 73.09 & 0.1661 & 57.04 & 0.8348 \\
 &  &  & 100 & 73.33 & 0.1113 & 57.75 & 0.1754 \\
 &  &  & 150 & 73.18 & 0.2106 & 57.39 & 0.9737 \\
 &  &  & 250 & 73.32 & 0.1253 & 58.57 & 0.5841 \\
 &  &  & 500 & 73.37 & 0.0942 & 58.30 & 0.2289 \\
 &  &  & 1000 & 73.35 & 0.1191 & 58.46 & 0.1992 \\
 &  &  & 2000 & 73.83 & 0.0841 & 58.77 & 0.3297 \\
  \cmidrule{3-8}
 & \multirow{20}{*}{XLM-R} & \multirow{10}{*}{mBERT} & 10 & 77.81 & 0.0988 & 55.75 & 0.4481 \\
 &  &  & 25 & 78.01 & 0.1861 & 57.08 & 0.7400 \\
 &  &  & 50 & 78.30 & 0.2015 & 60.14 & 0.5176 \\
 &  &  & 75 & 78.23 & 0.2273 & 60.17 & 0.5350 \\
 &  &  & 100 & 78.27 & 0.0507 & 60.74 & 0.8992 \\
 &  &  & 150 & 78.36 & 0.0364 & 60.39 & 0.4339 \\
 &  &  & 250 & 78.28 & 0.0775 & 60.53 & 0.7546 \\
 &  &  & 500 & 78.31 & 0.0616 & 60.94 & 0.4437 \\
 &  &  & 1000 & 78.38 & 0.0234 & 61.19 & 0.1991 \\
 &  &  & 2000 & 79.01 & 0.0151 & 61.05 & 0.2984 \\
  \cmidrule{4-8}
 &  & \multirow{10}{*}{XLM-R} & 10 & 77.20 & 0.1125 & 53.67 & 1.4657 \\
 &  &  & 25 & 77.42 & 0.1231 & 56.56 & 0.6499 \\
 &  &  & 50 & 77.30 & 0.1159 & 56.25 & 0.7295 \\
 &  &  & 75 & 77.56 & 0.1931 & 58.83 & 0.7268 \\
 &  &  & 100 & 77.62 & 0.1107 & 59.39 & 0.4887 \\
 &  &  & 150 & 77.55 & 0.1795 & 59.35 & 1.1807 \\
 &  &  & 250 & 77.62 & 0.1000 & 60.19 & 0.3460 \\
 &  &  & 500 & 77.70 & 0.1177 & 60.59 & 0.4153 \\
 &  &  & 1000 & 77.66 & 0.1100 & 60.44 & 0.3318 \\
 &  &  & 2000 & 78.17 & 0.0506 & 60.65 & 0.2407 \\
  \cmidrule{1-8}
\multirow{20}{*}{\textit{medium}} & \multirow{20}{*}{mBERT} & \multirow{10}{*}{mBERT} & 10 & 73.68 & 0.3690 & 50.08 & 1.9981 \\
 &  &  & 25 & 74.22 & 0.3696 & 53.05 & 1.5703 \\
 &  &  & 50 & 74.53 & 0.2225 & 55.32 & 0.4839 \\
 &  &  & 75 & 74.53 & 0.2037 & 54.80 & 0.4711 \\
 &  &  & 100 & 74.60 & 0.0304 & 55.91 & 0.5138 \\
 &  &  & 150 & 74.61 & 0.1233 & 55.53 & 0.2667 \\
 &  &  & 250 & 74.60 & 0.0907 & 55.79 & 1.1553 \\
 &  &  & 500 & 74.73 & 0.0621 & 55.91 & 0.2530 \\
 &  &  & 1000 & 74.75 & 0.0310 & 55.96 & 0.1029 \\
 &  &  & 2000 & 75.32 & 0.1101 & 55.82 & 0.3875 \\
  \cmidrule{4-8}
 &  & \multirow{10}{*}{XLM-R} & 10 & 72.10 & 0.2370 & 44.67 & 1.2159 \\
 &  &  & 25 & 72.69 & 0.2353 & 48.87 & 1.4164 \\
 &  &  & 50 & 72.86 & 0.1843 & 49.16 & 0.2496 \\
 &  &  & 75 & 72.74 & 0.1238 & 49.88 & 0.8218 \\
 &  &  & 100 & 73.03 & 0.1312 & 50.18 & 0.3240 \\
 &  &  & 150 & 73.05 & 0.3915 & 50.50 & 0.9673 \\
 &  &  & 250 & 72.99 & 0.2487 & 50.03 & 0.3608 \\
 &  &  & 500 & 72.98 & 0.0635 & 50.70 & 0.0975 \\
 &  &  & 1000 & 72.93 & 0.0945 & 50.38 & 0.3225 \\
 &  &  & 2000 & 73.41 & 0.1078 & 50.64 & 0.1503 \\
\\
\multirow{20}{*}{\textit{medium}} & \multirow{20}{*}{XLM-R} & \multirow{10}{*}{mBERT} & 10 & 77.40 & 0.2719 & 48.40 & 1.8551 \\*
 &  &  & 25 & 77.65 & 0.3435 & 51.10 & 1.7390 \\*
 &  &  & 50 & 77.89 & 0.1766 & 53.58 & 0.5910 \\
 &  &  & 75 & 77.89 & 0.1505 & 53.20 & 0.8143 \\
 &  &  & 100 & 77.95 & 0.0438 & 54.17 & 0.2357 \\
 &  &  & 150 & 78.02 & 0.1706 & 54.10 & 0.5164 \\
 &  &  & 250 & 77.92 & 0.0311 & 54.06 & 0.5838 \\
 &  &  & 500 & 78.06 & 0.0560 & 54.17 & 0.2143 \\
 &  &  & 1000 & 78.07 & 0.0402 & 53.92 & 0.2182 \\
 &  &  & 2000 & 78.70 & 0.0901 & 54.08 & 0.0361 \\
  \cmidrule{4-8}
 &  & \multirow{10}{*}{XLM-R} & 10 & 76.10 & 0.3626 & 45.51 & 1.8007 \\
 &  &  & 25 & 76.59 & 0.2389 & 48.73 & 1.6025 \\
 &  &  & 50 & 76.68 & 0.1508 & 49.21 & 0.6561 \\
 &  &  & 75 & 76.63 & 0.1566 & 50.16 & 0.5225 \\
 &  &  & 100 & 76.88 & 0.1513 & 50.96 & 0.2359 \\
 &  &  & 150 & 76.85 & 0.3256 & 51.24 & 0.9151 \\
 &  &  & 250 & 76.86 & 0.2388 & 51.49 & 0.6931 \\
 &  &  & 500 & 76.83 & 0.0592 & 51.89 & 0.3107 \\
 &  &  & 1000 & 76.71 & 0.1380 & 51.26 & 0.5007 \\
 &  &  & 2000 & 77.26 & 0.0542 & 51.80 & 0.4134 \\
  \cmidrule{1-8}
\multirow{40}{*}{\textit{all}} & \multirow{20}{*}{mBERT} & \multirow{10}{*}{mBERT} & 10 & 72.58 & 0.1915 & 39.17 & 1.4141 \\
 &  &  & 25 & 72.78 & 0.5296 & 41.88 & 1.6804 \\
 &  &  & 50 & 73.14 & 0.1272 & 43.94 & 0.3211 \\
 &  &  & 75 & 73.31 & 0.1674 & 43.95 & 0.5767 \\
 &  &  & 100 & 73.28 & 0.0378 & 43.88 & 0.7011 \\
 &  &  & 150 & 73.32 & 0.0831 & 44.61 & 0.6676 \\
 &  &  & 250 & 73.42 & 0.3133 & 44.29 & 0.6229 \\
 &  &  & 500 & 73.30 & 0.2252 & 44.22 & 0.3289 \\
 &  &  & 1000 & 73.35 & 0.0902 & 44.23 & 0.3403 \\
 &  &  & 2000 & 73.88 & 0.0583 & 44.16 & 0.3167 \\
  \cmidrule{4-8}
 &  & \multirow{10}{*}{XLM-R} & 10 & 70.86 & 0.2847 & 35.36 & 1.6877 \\
 &  &  & 25 & 71.14 & 0.5082 & 38.69 & 1.3061 \\
 &  &  & 50 & 71.64 & 0.2958 & 41.12 & 0.6685 \\
 &  &  & 75 & 71.50 & 0.2919 & 41.23 & 0.8403 \\
 &  &  & 100 & 71.66 & 0.1017 & 42.16 & 0.8231 \\
 &  &  & 150 & 71.59 & 0.0429 & 40.91 & 0.4555 \\
 &  &  & 250 & 71.53 & 0.1310 & 41.40 & 0.3779 \\
 &  &  & 500 & 71.43 & 0.2147 & 40.86 & 0.3017 \\
 &  &  & 1000 & 71.43 & 0.2595 & 40.76 & 0.0447 \\
 &  &  & 2000 & 71.84 & 0.0482 & 40.66 & 0.2772 \\
  \cmidrule{3-8}
 & \multirow{20}{*}{XLM-R} & \multirow{10}{*}{mBERT} & 10 & 74.98 & 0.2255 & 35.60 & 1.2733 \\
 &  &  & 25 & 75.02 & 0.6532 & 39.23 & 1.5762 \\
 &  &  & 50 & 75.33 & 0.0895 & 40.94 & 0.4181 \\
 &  &  & 75 & 75.60 & 0.1585 & 41.25 & 0.5508 \\
 &  &  & 100 & 75.30 & 0.1640 & 40.92 & 0.4279 \\
 &  &  & 150 & 75.62 & 0.1676 & 41.48 & 0.9218 \\
 &  &  & 250 & 75.50 & 0.2844 & 42.03 & 0.7664 \\
 &  &  & 500 & 75.52 & 0.3166 & 41.30 & 0.2536 \\
 &  &  & 1000 & 75.55 & 0.1301 & 41.68 & 0.2655 \\
 &  &  & 2000 & 76.15 & 0.1159 & 41.58 & 0.1400 \\
 \cmidrule{4-8}
 &  & \multirow{10}{*}{XLM-R} & 10 & 73.90 & 0.3098 & 33.60 & 0.5184 \\
 &  &  & 25 & 74.13 & 0.6088 & 36.48 & 1.2740 \\
 &  &  & 50 & 74.66 & 0.4388 & 38.75 & 1.2252 \\
 &  &  & 75 & 74.46 & 0.1867 & 39.36 & 0.5701 \\
 &  &  & 100 & 74.64 & 0.0280 & 40.01 & 1.3144 \\
 &  &  & 150 & 74.59 & 0.0899 & 39.33 & 0.3540 \\
 &  &  & 250 & 74.55 & 0.0918 & 39.93 & 0.6842 \\
 &  &  & 500 & 74.39 & 0.2216 & 39.55 & 0.3978 \\
 &  &  & 1000 & 74.44 & 0.2580 & 39.21 & 0.1768 \\
 &  &  & 2000 & 74.83 & 0.0892 & 39.23 & 0.4762 \\
\bottomrule
\end{xltabular}}

\clearpage

\section{Development Set Results}

\begin{table}[H]
\begin{adjustbox}{width=\linewidth}
\begin{tabular}{lllllll}
\toprule
Avg. of Accuracy @ 5 &  &  & NN-Rank \textit{k} &  &  &  \\
\toprule
Task Model & Source Split & Evaluation Split & 5 & 10 & 20 & 25 \\
\midrule
mBERT & train-all & dev-all & 72.54 & 72.49 & 72.52 & 72.42 \\
 &  & dev-large & 78.78 & 78.81 & 78.85 & 78.85 \\
 &  & dev-medium & 75.63 & 75.57 & 75.50 & 75.42 \\
 & train-large & dev-all & 74.08 & 73.91 & 73.78 & 73.75 \\
 &  & dev-large & 78.80 & 78.92 & 78.70 & 78.68 \\
 &  & dev-medium & 77.11 & 76.92 & 76.75 & 76.73 \\
 & train-medium & dev-all & 73.90 & 73.85 & 73.73 & 73.62 \\
 &  & dev-large & 79.85 & 79.96 & 79.89 & 79.80 \\
 &  & dev-medium & 77.05 & 76.96 & 76.80 & 76.68 \\
XLM-R & train-all & dev-all & 75.12 & 75.08 & 75.09 & 74.92 \\
 &  & dev-large & 80.19 & 80.23 & 80.33 & 80.33 \\
 &  & dev-medium & 78.19 & 78.15 & 78.12 & 78.01 \\
 & train-large & dev-all & 78.06 & 77.93 & 77.83 & 77.80 \\
 &  & dev-large & 81.56 & 81.69 & 81.55 & 81.53 \\
 &  & dev-medium & 81.22 & 81.07 & 80.94 & 80.91 \\
 & train-medium & dev-all & 77.45 & 77.36 & 77.18 & 77.11 \\
 &  & dev-large & 82.11 & 82.21 & 82.15 & 82.07 \\
 &  & dev-medium & 80.68 & 80.57 & 80.43 & 80.31 \\
\toprule
Avg. of NDCG @ 5 &  &  & NN-Rank \textit{k} &  &  &  \\
\midrule
Evaluation Model & Source Split & Evaluation Split & 5 & 10 & 20 & 25 \\
\midrule
mBERT & train-all & dev-all & 43.47 & 43.02 & 42.88 & 42.69 \\
 &  & dev-large & 52.66 & 52.21 & 52.24 & 52.08 \\
 &  & dev-medium & 46.30 & 45.89 & 45.73 & 45.49 \\
 & train-large & dev-all & 61.43 & 60.90 & 60.12 & 59.75 \\
 &  & dev-large & 69.46 & 69.15 & 68.36 & 67.88 \\
 &  & dev-medium & 64.35 & 63.41 & 62.41 & 62.19 \\
 & train-medium & dev-all & 53.81 & 53.67 & 53.39 & 52.94 \\
 &  & dev-large & 63.60 & 63.57 & 64.04 & 63.94 \\
 &  & dev-medium & 57.41 & 57.30 & 57.10 & 56.46 \\
XLM-R & train-all & dev-all & 41.42 & 41.16 & 41.19 & 40.91 \\
 &  & dev-large & 50.19 & 49.73 & 49.84 & 49.68 \\
 &  & dev-medium & 44.63 & 44.49 & 44.59 & 44.35 \\
 & train-large & dev-all & 60.95 & 60.21 & 59.38 & 59.06 \\
 &  & dev-large & 67.62 & 67.16 & 66.56 & 66.05 \\
 &  & dev-medium & 64.20 & 63.24 & 62.24 & 61.87 \\
 & train-medium & dev-all & 53.50 & 53.12 & 52.67 & 52.41 \\
 &  & dev-large & 63.68 & 63.66 & 63.57 & 63.49 \\
 &  & dev-medium & 57.29 & 56.91 & 56.58 & 56.19 \\
\bottomrule
\end{tabular}

\end{adjustbox}
   
    \caption{POS \nnr{} Development set results. Values are averaged over the two ranking models.}
    \label{tab:pos_nn_dev_results}
\end{table}

\begin{table}[H]
\begin{adjustbox}{width=\linewidth}
\begin{tabular}{lllllll}
\toprule
AVERAGE of F1@5 &  &  & NN-Rank \textit{k} &  &  &  \\
\toprule
Task Model & Source Split & Evaluation Split & 5 & 10 & 20 & 25 \\
\midrule
mBERT & train-all & dev-all & 58.41 & 58.09 & 57.72 & 57.64 \\
 &  & dev-large & 68.41 & 68.19 & 68.06 & 67.98 \\
 &  & dev-medium & 65.51 & 65.13 & 64.91 & 64.86 \\
 & train-large & dev-all & 60.33 & 60.20 & 60.10 & 60.07 \\
 &  & dev-large & 69.22 & 69.11 & 69.01 & 68.99 \\
 &  & dev-medium & 67.07 & 66.98 & 66.87 & 66.83 \\
 & train-medium & dev-all & 60.43 & 60.26 & 60.25 & 60.23 \\
 &  & dev-large & 69.34 & 69.18 & 69.10 & 69.05 \\
 &  & dev-medium & 67.08 & 66.89 & 66.81 & 66.77 \\
XLM-R & train-all & dev-all & 58.00 & 57.61 & 57.13 & 57.06 \\
 &  & dev-large & 66.64 & 66.40 & 66.20 & 66.18 \\
 &  & dev-medium & 63.96 & 63.52 & 63.26 & 63.24 \\
 & train-large & dev-all & 60.84 & 60.61 & 60.42 & 60.37 \\
 &  & dev-large & 68.21 & 68.14 & 68.01 & 67.97 \\
 &  & dev-medium & 66.28 & 66.19 & 66.03 & 65.98 \\
 & train-medium & dev-all & 61.09 & 60.90 & 60.84 & 60.82 \\
 &  & dev-large & 68.12 & 67.94 & 67.92 & 67.88 \\
 &  & dev-medium & 66.31 & 66.04 & 65.98 & 65.94 \\
\toprule
Avg. of NDCG@5 &  &  &NN-Rank \textit{k}&  &  &  \\
\toprule
Task Model & Source Split & Evaluation Split & 5 & 10 & 20 & 25 \\
\midrule
mBERT & train-all & dev-all & {40.91} & {40.57} & {39.89} & {39.37} \\
 &  & dev-large & {50.46} & {50.19} & {49.93} & {49.41} \\
 &  & dev-medium & {45.76} & {45.34} & {44.80} & {44.35} \\
 & train-large & dev-all & {46.09} & {45.49} & {45.10} & {44.92} \\
 &  & dev-large & {53.33} & {52.58} & {52.93} & {52.90} \\
 &  & dev-medium & {50.48} & {49.68} & {49.31} & {48.99} \\
 & train-medium & dev-all & {45.39} & {44.85} & {45.06} & {44.81} \\
 &  & dev-large & {52.78} & {51.96} & {52.18} & {51.98} \\
 &  & dev-medium & {49.19} & {48.45} & {48.35} & {48.09} \\
XLM-R & train-all & dev-all & {41.36} & {40.90} & {40.14} & {40.15} \\
 &  & dev-large & {54.75} & {54.68} & {53.88} & {53.90} \\
 &  & dev-medium & {47.43} & {46.97} & {46.28} & {46.35} \\
 & train-large & dev-all & {47.35} & {46.48} & {46.26} & {45.90} \\
 &  & dev-large & {60.04} & {59.25} & {59.05} & {58.70} \\
 &  & dev-medium & {53.54} & {52.98} & {52.85} & {52.50} \\
 & train-medium & dev-all & {46.08} & {45.90} & {45.79} & {45.73} \\
 &  & dev-large & {57.79} & {57.49} & {57.58} & {57.65} \\
 &  & dev-medium & {51.41} & {50.97} & {50.77} & {50.77} \\
 \bottomrule
\end{tabular}
\end{adjustbox}
   
    \caption{NER \nnr{} development set results. Values are averaged over the two ranking models.}
    \label{tab:ner_nn_dev_results}
\end{table}

\clearpage

\section{Language Tables}
\small{
\begin{xltabular}{\textwidth}{lllll}
\toprule
UD\_ISO & ISO-3 & Language Family & Treebank & Language Splits \\
\toprule
\endfirsthead
\toprule
UD\_ISO & ISO-3 & Language Family & Treebank & Language Splits \\
\toprule
\endhead
\bottomrule
&&&\multicolumn{2}{c}{\textit{Continued on next page}}  \\
\bottomrule
\endfoot
\bottomrule
\\ 
\caption{POS Train Languages and Datasets. There are 78 datasets in the \textit{train-all}, 42 in train-medium, and 25 in train-large. The number of unique languages in train-all is 51, 28 in train-medium, and 20 in train-large. In train-all, there are languages from 11 language families and one language isolate, however this distribution is heavily biased towards Indo-European languages. } \label{tab:pos_train_langs}
\endlastfoot
af & afr & Indo-European & UD\_Afrikaans-AfriBooms & train\_all \\
hy & hye & Indo-European & UD\_Armenian-ArmTDP & train\_all \\
eu & eus & - & UD\_Basque-BDT & train\_all \\
zh & zho & Sino-Tibetan & UD\_Chinese-GSD & train\_all \\
cop & cop & Afro-Asiatic & UD\_Coptic-Scriptorium & train\_all \\
cs & ces & Indo-European & UD\_Czech-CLTT & train\_all \\
da & dan & Indo-European & UD\_Danish-DDT & train\_all \\
en & eng & Indo-European & UD\_English-LinES & train\_all \\
en & eng & Indo-European & UD\_English-ParTUT & train\_all \\
fr & fra & Indo-European & UD\_French-ParTUT & train\_all \\
fr & fra & Indo-European & UD\_French-Sequoia & train\_all \\
gl & glg & Indo-European & UD\_Galician-CTG & train\_all \\
gl & glg & Indo-European & UD\_Galician-TreeGal & train\_all \\
got & got & Indo-European & UD\_Gothic-PROIEL & train\_all \\
el & ell & Indo-European & UD\_Greek-GDT & train\_all \\
he & heb & Afro-Asiatic & UD\_Hebrew-HTB & train\_all \\
hu & hun & Uralic & UD\_Hungarian-Szeged & train\_all \\
ga & gle & Indo-European & UD\_Irish-IDT & train\_all \\
it & ita & Indo-European & UD\_Italian-ParTUT & train\_all \\
it & ita & Indo-European & UD\_Italian-PoSTWITA & train\_all \\
ko & kor & Koreanic & UD\_Korean-GSD & train\_all \\
la & lat & Indo-European & UD\_Latin-Perseus & train\_all \\
sme & sme & Uralic & UD\_North\_Sami-Giella & train\_all \\
fa & fas & Indo-European & UD\_Persian-Seraji & train\_all \\
pl & pol & Indo-European & UD\_Polish-LFG & train\_all \\
ru & rus & Indo-European & UD\_Russian-GSD & train\_all \\
sr & srp & Indo-European & UD\_Serbian-SET & train\_all \\
sk & slk & Indo-European & UD\_Slovak-SNK & train\_all \\
sl & slv & Indo-European & UD\_Slovenian-SST & train\_all \\
sv & swe & Indo-European & UD\_Swedish-LinES & train\_all \\
sv & swe & Indo-European & UD\_Swedish-Talbanken & train\_all \\
ta & tam & Dravidian & UD\_Tamil-TTB & train\_all \\
tr & tur & Turkic & UD\_Turkish-IMST & train\_all \\
ur & urd & Indo-European & UD\_Urdu-UDTB & train\_all \\
ug & uig & Turkic & UD\_Uyghur-UDT & train\_all \\
vi & vie & Austroasiatic & UD\_Vietnamese-VTB & train\_all \\
grc & grc & Indo-European & UD\_Ancient\_Greek-Perseus & train\_all, train\_medium \\
grc & grc & Indo-European & UD\_Ancient\_Greek-PROIEL & train\_all, train\_medium \\
ar & ara & Afroasiatic & UD\_Arabic-NYUAD & train\_all, train\_medium \\
bg & bul & Indo-European & UD\_Bulgarian-BTB & train\_all, train\_medium \\
hr & hrv & Indo-European & UD\_Croatian-SET & train\_all, train\_medium \\
cs & ces & Indo-European & UD\_Czech-FicTree & train\_all, train\_medium \\
nl & nld & Indo-European & UD\_Dutch-Alpino & train\_all, train\_medium \\
fi & fin & Uralic & UD\_Finnish-FTB & train\_all, train\_medium \\
id & ind & Austronesian & UD\_Indonesian-GSD & train\_all, train\_medium \\
ja & jpn & Japonic & UD\_Japanese-BCCWJ & train\_all, train\_medium \\
ja & jpn & Japonic & UD\_Japanese-GSD & train\_all, train\_medium \\
ko & kor & Koreanic & UD\_Korean-Kaist & train\_all, train\_medium \\
la & lat & Indo-European & UD\_Latin-PROIEL & train\_all, train\_medium \\
cu & chu & Indo-European & UD\_Old\_Church\_Slavonic-PROIEL & train\_all, train\_medium \\
pt & por & Indo-European & UD\_Portuguese-Bosque & train\_all, train\_medium \\
ro & ron & Indo-European & UD\_Romanian-RRT & train\_all, train\_medium \\
uk & ukr & Indo-European & UD\_Ukrainian-IU & train\_all, train\_medium \\
ar & ara & Afroasiatic & UD\_Arabic-PADT & train\_all, train\_medium, train\_large \\
be & bel & Indo-European & UD\_Belarusian-HSE & train\_all, train\_medium, train\_large \\
ca & cat & Indo-European & UD\_Catalan-AnCora & train\_all, train\_medium, train\_large \\
cs & ces & Indo-European & UD\_Czech-CAC & train\_all, train\_medium, train\_large \\
cs & ces & Indo-European & UD\_Czech-PDT & train\_all, train\_medium, train\_large \\
nl & nld & Indo-European & UD\_Dutch-LassySmall & train\_all, train\_medium, train\_large \\
en & eng & Indo-European & UD\_English-EWT & train\_all, train\_medium, train\_large \\
en & eng & Indo-European & UD\_English-GUM & train\_all, train\_medium, train\_large \\
et & est & Uralic & UD\_Estonian-EDT & train\_all, train\_medium, train\_large \\
fi & fin & Uralic & UD\_Finnish-TDT & train\_all, train\_medium, train\_large \\
fr & fra & Indo-European & UD\_French-GSD & train\_all, train\_medium, train\_large \\
de & deu & Indo-European & UD\_German-GSD & train\_all, train\_medium, train\_large \\
hi & hin & Indo-European & UD\_Hindi-HDTB & train\_all, train\_medium, train\_large \\
it & ita & Indo-European & UD\_Italian-ISDT & train\_all, train\_medium, train\_large \\
la & lat & Indo-European & UD\_Latin-ITTB & train\_all, train\_medium, train\_large \\
lv & lav & Indo-European & UD\_Latvian-LVTB & train\_all, train\_medium, train\_large \\
no & nor & Indo-European & UD\_Norwegian-Bokmaal & train\_all, train\_medium, train\_large \\
no & nor & Indo-European & UD\_Norwegian-Nynorsk & train\_all, train\_medium, train\_large \\
pt & por & Indo-European & UD\_Portuguese-GSD & train\_all, train\_medium, train\_large \\
ro & ron & Indo-European & UD\_Romanian-Nonstandard & train\_all, train\_medium, train\_large \\
ru & rus & Indo-European & UD\_Russian-SynTagRus & train\_all, train\_medium, train\_large \\
ru & rus & Indo-European & UD\_Russian-Taiga & train\_all, train\_medium, train\_large \\
sl & slv & Indo-European & UD\_Slovenian-SSJ & train\_all, train\_medium, train\_large \\
es & spa & Indo-European & UD\_Spanish-AnCora & train\_all, train\_medium, train\_large \\
es & spa & Indo-European & UD\_Spanish-GSD & train\_all, train\_medium, train\_large \\
\bottomrule
\end{xltabular}}

\small{
\begin{xltabular}{\textwidth}{lllll}
\toprule
UD\_ISO & ISO 639-3 & Language Family & Treebank & Language Splits \\
\bottomrule
\endfirsthead
\toprule
UD\_ISO & ISO 639-3 & Language Family & Treebank & Language Splits \\
\bottomrule
\endhead
\bottomrule
&&&\multicolumn{2}{c}{\textit{Continued on next page}}  \\
\bottomrule
\endfoot
\bottomrule
\\ 
\caption{POS Test Languages and Datasets. There are 118 total dataset in test-all, 83 in test-medium, and 25 in test-large. There are 56 unique languages in test-all, 46 in test-medium, and 21 in test-large. Languages cover 12 language families and 1 language isolate.} \label{tab:pos_test_langs}
\endlastfoot
hy & hye & Indo-European & UD\_Armenian-ArmTDP & test\_all \\
zh & zho & Tino-Sibetan & UD\_Chinese-GSD & test\_all \\
zh & zho & Tino-Sibetan & UD\_Chinese-GSDSimp & test\_all \\
da & dan & Indo-European & UD\_Danish-DDT & test\_all \\
en & eng & Indo-European & UD\_English-Atis & test\_all \\
en & eng & Indo-European & UD\_English-ESLSpok & test\_all \\
en & eng & Indo-European & UD\_English-GUMReddit & test\_all \\
en & eng & Indo-European & UD\_English-ParTUT & test\_all \\
fo & fao & Indo-European & UD\_Faroese-FarPaHC & test\_all \\
fr & fra & Indo-European & UD\_French-GSD & test\_all \\
fr & fra & Indo-European & UD\_French-ParisStories & test\_all \\
fr & fra & Indo-European & UD\_French-ParTUT & test\_all \\
fr & fra & Indo-European & UD\_French-Rhapsodie & test\_all \\
fr & fra & Indo-European & UD\_French-Sequoia & test\_all \\
he & heb & Afro-Asiatic & UD\_Hebrew-HTB & test\_all \\
he & heb & Afro-Asiatic & UD\_Hebrew-IAHLTknesset & test\_all \\
he & heb & Afro-Asiatic & UD\_Hebrew-IAHLTwiki & test\_all \\
is & isl & Indo-European & UD\_Icelandic-Modern & test\_all \\
ga & gle & Indo-European & UD\_Irish-IDT & test\_all \\
it & ita & Indo-European & UD\_Italian-ISDT & test\_all \\
it & ita & Indo-European & UD\_Italian-MarkIT & test\_all \\
it & ita & Indo-European & UD\_Italian-Old & test\_all \\
it & ita & Indo-European & UD\_Italian-ParTUT & test\_all \\
it & ita & Indo-European & UD\_Italian-TWITTIRO & test\_all \\
ko & kor & Koreanic & UD\_Korean-GSD & test\_all \\
ko & kor & Koreanic & UD\_Korean-KSL & test\_all \\
lt & lit & Indo-European & UD\_Lithuanian-HSE & test\_all \\
mt & mlt & Afro-Asiatic & UD\_Maltese-MUDT & test\_all \\
gd & gla & Indo-European & UD\_Scottish\_Gaelic-ARCOSG & test\_all \\
sl & slv & Indo-European & UD\_Slovenian-SST & test\_all \\
ta & tam & Dravidian & UD\_Tamil-TTB & test\_all \\
tr & tur & Turkic & UD\_Turkish-Atis & test\_all \\
tr & tur & Turkic & UD\_Turkish-FrameNet & test\_all \\
vi & vie & Austroasiatic & UD\_Vietnamese-VTB & test\_all \\
wo & wol & Atlantic-Congo & UD\_Wolof-WTB & test\_all \\
af & afr & Indo-European & UD\_Afrikaans-AfriBooms & test\_all, test\_medium \\
hy & hye & Indo-European & UD\_Armenian-BSUT & test\_all, test\_medium \\
bg & bul & Indo-European & UD\_Bulgarian-BTB & test\_all, test\_medium \\
hr & hrv & Indo-European & UD\_Croatian-SET & test\_all, test\_medium \\
cs & ces & Indo-European & UD\_Czech-CAC & test\_all, test\_medium \\
cs & ces & Indo-European & UD\_Czech-CLTT & test\_all, test\_medium \\
cs & ces & Indo-European & UD\_Czech-FicTree & test\_all, test\_medium \\
nl & nld & Indo-European & UD\_Dutch-Alpino & test\_all, test\_medium \\
en & eng & Indo-European & UD\_English-LinES & test\_all, test\_medium \\
et & est & Uralic & UD\_Estonian-EWT & test\_all, test\_medium \\
fi & fin & Uralic & UD\_Finnish-FTB & test\_all, test\_medium \\
fi & fin & Uralic & UD\_Finnish-TDT & test\_all, test\_medium \\
ka & kat & Kartvelian & UD\_Georgian-GLC & test\_all, test\_medium \\
de & deu & Indo-European & UD\_German-GSD & test\_all, test\_medium \\
el & ell & Indo-European & UD\_Greek-GDT & test\_all, test\_medium \\
hu & hun & Uralic & UD\_Hungarian-Szeged & test\_all, test\_medium \\
is & isl & Indo-European & UD\_Icelandic-GC & test\_all, test\_medium \\
id & ind & Austronesian & UD\_Indonesian-GSD & test\_all, test\_medium \\
ga & gle & Indo-European & UD\_Irish-TwittIrish & test\_all, test\_medium \\
it & ita & Indo-European & UD\_Italian-PoSTWITA & test\_all, test\_medium \\
it & ita & Indo-European & UD\_Italian-VIT & test\_all, test\_medium \\
ja & jpn & Japonic & UD\_Japanese-GSD & test\_all, test\_medium \\
ja & jpn & Japonic & UD\_Japanese-GSDLUW & test\_all, test\_medium \\
ko & kor & Koreanic & UD\_Korean-Kaist & test\_all, test\_medium \\
la & lat & Indo-European & UD\_Latin-LLCT & test\_all, test\_medium \\
la & lat & Indo-European & UD\_Latin-PROIEL & test\_all, test\_medium \\
la & lat & Indo-European & UD\_Latin-UDante & test\_all, test\_medium \\
lt & lit & Indo-European & UD\_Lithuanian-ALKSNIS & test\_all, test\_medium \\
cu & chu & Indo-European & UD\_Old\_Church\_Slavonic-PROIEL & test\_all, test\_medium \\
fa & fas & Indo-European & UD\_Persian-PerDT & test\_all, test\_medium \\
fa & fas & Indo-European & UD\_Persian-Seraji & test\_all, test\_medium \\
pl & pol & Indo-European & UD\_Polish-LFG & test\_all, test\_medium \\
pt & por & Indo-European & UD\_Portuguese-Bosque & test\_all, test\_medium \\
pt & por & Indo-European & UD\_Portuguese-GSD & test\_all, test\_medium \\
pt & por & Indo-European & UD\_Portuguese-PetroGold & test\_all, test\_medium \\
ro & ron & Indo-European & UD\_Romanian-Nonstandard & test\_all, test\_medium \\
ro & ron & Indo-European & UD\_Romanian-RRT & test\_all, test\_medium \\
ro & ron & Indo-European & UD\_Romanian-SiMoNERo & test\_all, test\_medium \\
ru & rus & Indo-European & UD\_Russian-GSD & test\_all, test\_medium \\
ru & rus & Indo-European & UD\_Russian-Poetry & test\_all, test\_medium \\
ru & rus & Indo-European & UD\_Russian-Taiga & test\_all, test\_medium \\
sa & san & Indo-European & UD\_Sanskrit-Vedic & test\_all, test\_medium \\
sr & srp & Indo-European & UD\_Serbian-SET & test\_all, test\_medium \\
sk & slk & Indo-European & UD\_Slovak-SNK & test\_all, test\_medium \\
es & spa & Indo-European & UD\_Spanish-GSD & test\_all, test\_medium \\
sv & swe & Indo-European & UD\_Swedish-LinES & test\_all, test\_medium \\
tr & tur & Turkic & UD\_Turkish-BOUN & test\_all, test\_medium \\
tr & tur & Turkic & UD\_Turkish-IMST & test\_all, test\_medium \\
tr & tur & Turkic & UD\_Turkish-Kenet & test\_all, test\_medium \\
tr & tur & Turkic & UD\_Turkish-Penn & test\_all, test\_medium \\
tr & tur & Turkic & UD\_Turkish-Tourism & test\_all, test\_medium \\
uk & ukr & Indo-European & UD\_Ukrainian-IU & test\_all, test\_medium \\
uk & ukr & Indo-European & UD\_Ukrainian-ParlaMint & test\_all, test\_medium \\
ur & urd & Indo-European & UD\_Urdu-UDTB & test\_all, test\_medium \\
ug & uig & Turkic & UD\_Uyghur-UDT & test\_all, test\_medium \\
cy & cym & Indo-European & UD\_Welsh-CCG & test\_all, test\_medium \\
ar & ara & Afro-Asiatic & UD\_Arabic-PADT & test\_all, test\_medium, test\_large \\
eu & eus & - & UD\_Basque-BDT & test\_all, test\_medium, test\_large \\
be & bel & Indo-European & UD\_Belarusian-HSE & test\_all, test\_medium, test\_large \\
ca & cat & Indo-European & UD\_Catalan-AnCora & test\_all, test\_medium, test\_large \\
cs & ces & Indo-European & UD\_Czech-PDT & test\_all, test\_medium, test\_large \\
nl & nld & Indo-European & UD\_Dutch-LassySmall & test\_all, test\_medium, test\_large \\
en & eng & Indo-European & UD\_English-EWT & test\_all, test\_medium, test\_large \\
en & eng & Indo-European & UD\_English-GUM & test\_all, test\_medium, test\_large \\
et & est & Uralic & UD\_Estonian-EDT & test\_all, test\_medium, test\_large \\
gl & glg & Indo-European & UD\_Galician-CTG & test\_all, test\_medium, test\_large \\
de & deu & Indo-European & UD\_German-HDT & test\_all, test\_medium, test\_large \\
hi & hin & Indo-European & UD\_Hindi-HDTB & test\_all, test\_medium, test\_large \\
is & isl & Indo-European & UD\_Icelandic-IcePaHC & test\_all, test\_medium, test\_large \\
la & lat & Indo-European & UD\_Latin-ITTB & test\_all, test\_medium, test\_large \\
lv & lav & Indo-European & UD\_Latvian-LVTB & test\_all, test\_medium, test\_large \\
no & nor & Indo-European & UD\_Norwegian-Bokmaal & test\_all, test\_medium, test\_large \\
no & nor & Indo-European & UD\_Norwegian-Nynorsk & test\_all, test\_medium, test\_large \\
pl & pol & Indo-European & UD\_Polish-PDB & test\_all, test\_medium, test\_large \\
pt & por & Indo-European & UD\_Portuguese-CINTIL & test\_all, test\_medium, test\_large \\
pt & por & Indo-European & UD\_Portuguese-Porttinari & test\_all, test\_medium, test\_large \\
ru & rus & Indo-European & UD\_Russian-SynTagRus & test\_all, test\_medium, test\_large \\
sl & slv & Indo-European & UD\_Slovenian-SSJ & test\_all, test\_medium, test\_large \\
es & spa & Indo-European & UD\_Spanish-AnCora & test\_all, test\_medium, test\_large \\
sv & swe & Indo-European & UD\_Swedish-Talbanken & test\_all, test\_medium, test\_large \\
\bottomrule
\end{xltabular}}

\small{
\begin{xltabular}{\textwidth}{lllll}
\toprule
UD\_ISO & ISO-3 & Language Family & treebank & Language Split \\
\bottomrule
\endfirsthead
\toprule
UD\_ISO & ISO-3 & Language Family & treebank & Language Split \\
\bottomrule
\endhead
\bottomrule
&&&\multicolumn{2}{c}{\textit{Continued on next page}}  \\
\bottomrule
\endfoot
\bottomrule
\\ 
\caption{NER Train Languages and Datasets. There are 51 datasets in train-all, 44 in train-medium, and 37 in train-large. As the WikiANN dataset only has one dataset per language, these counts represent unique languages as well. There are languages from 10 different languages families, 1 artificial language, and 1 language isolate. } \label{tab:ner_train_langs}
\endlastfoot
ckb & ckb & Indo-European & Central Kurdish & train\_all \\
la & lat & Indo-European & Latin & train\_all \\
br & bre & Indo-European & Breton & train\_all \\
hi & hin & Indo-European & Hindi & train\_all \\
ga & gle & Indo-European & Irish & train\_all \\
af & afr & Indo-European & Afrikaans & train\_all \\
tt & tat & Turkic & Tatar & train\_all \\
cy & cym & Indo-European & Welsh & train\_all,train\_medium \\
eu & eus & - & Basque & train\_all,train\_medium \\
lv & lav & Indo-European & Latvian & train\_all,train\_medium \\
tl & tgl & Austronesian & Tagalog & train\_all,train\_medium \\
mk & mkd & Indo-European & Macedonian & train\_all,train\_medium \\
bn & ben & Indo-European & Bengali & train\_all,train\_medium \\
lt & lit & Indo-European & Lithuanian & train\_all,train\_medium \\
it & ita & Indo-European & Italian & train\_all,train\_medium,train\_large \\
sr & srp & Indo-European & Serbian Standard & train\_all,train\_medium,train\_large \\
sl & slv & Indo-European & Slovenian & train\_all,train\_medium,train\_large \\
ko & kor & Koreanic & Korean & train\_all,train\_medium,train\_large \\
eo & epo & Artificial Language & Esperanto & train\_all,train\_medium,train\_large \\
pt & por & Indo-European & Portuguese & train\_all,train\_medium,train\_large \\
ta & tam & Dravidian & Tamil & train\_all,train\_medium,train\_large \\
es & spa & Indo-European & Spanish & train\_all,train\_medium,train\_large \\
et & est & Uralic & Estonian & train\_all,train\_medium,train\_large \\
ja & jpn & Japonic & Japanese & train\_all,train\_medium,train\_large \\
fi & fin & Uralic & Finnish & train\_all,train\_medium,train\_large \\
fr & fra & Indo-European & French & train\_all,train\_medium,train\_large \\
be & bel & Indo-European & Belarusian & train\_all,train\_medium,train\_large \\
nl & nld & Indo-European & Dutch & train\_all,train\_medium,train\_large \\
uk & ukr & Indo-European & Ukrainian & train\_all,train\_medium,train\_large \\
ur & urd & Indo-European & Urdu & train\_all,train\_medium,train\_large \\
de & deu & Indo-European & German & train\_all,train\_medium,train\_large \\
id & ind & Austronesian & Standard Indonesian & train\_all,train\_medium,train\_large \\
el & ell & Indo-European & Modern Greek & train\_all,train\_medium,train\_large \\
ru & rus & Indo-European & Russian & train\_all,train\_medium,train\_large \\
pl & pol & Indo-European & Polish & train\_all,train\_medium,train\_large \\
da & dan & Indo-European & Danish & train\_all,train\_medium,train\_large \\
bg & bul & Indo-European & Bulgarian & train\_all,train\_medium,train\_large \\
vi & vie & Austroasiatic & Vietnamese & train\_all,train\_medium,train\_large \\
sv & swe & Indo-European & Swedish & train\_all,train\_medium,train\_large \\
hu & hun & Uralic & Hungarian & train\_all,train\_medium,train\_large \\
zh & zho & Sino-Tibetan & Chinese & train\_all,train\_medium,train\_large \\
hy & hye & Indo-European & Eastern Armenian & train\_all,train\_medium,train\_large \\
th & tha & Tai-Kadai & Thai & train\_all,train\_medium,train\_large \\
nn & nno & Indo-European & Norwegian Nynorsk & train\_all,train\_medium,train\_large \\
ro & ron & Indo-European & Romanian & train\_all,train\_medium,train\_large \\
ca & cat & Indo-European & Catalan & train\_all,train\_medium,train\_large \\
tr & tur & Turkic & Turkish & train\_all,train\_medium,train\_large \\
sk & slk & Indo-European & Slovak & train\_all,train\_medium,train\_large \\
cs & ces & Indo-European & Czech & train\_all,train\_medium,train\_large \\
hr & hrv & Indo-European & Croatian Standard & train\_all,train\_medium,train\_large \\
ms & msa & Austronesian & Malay & train\_all,train\_medium,train\_large \\
\bottomrule
\end{xltabular}}

\small{
\begin{xltabular}{\textwidth}{lllll}
\toprule
UD\_ISO & ISO-3 & Language Family & treebank & Language Split \\
\bottomrule
\endfirsthead
\toprule
UD\_ISO & ISO-3 & Language Family & treebank & Language Splits \\
\bottomrule
\endhead
\bottomrule
&&&\multicolumn{2}{c}{\textit{Continued on next page}}  \\
\bottomrule
\endfoot
\bottomrule
\\ 
\caption{NER Test Languages and Datasets. In test-all, there are 72 dataset, 51 in test-medium, and 34 in test-large. Languages come from 13 language families, in addition to one language isolate and one artificial language. } \label{tab:ner_test_langs}
\endlastfoot
am & amh & Afro-Asiatic & Amharic & test\_all \\
my & mya & Sino-Tibetan & Burmese & test\_all \\
ceb & ceb & Austronesian & Cebuano & test\_all \\
km & khm & Austroasiatic & Central Khmer & test\_all \\
ce & che & Nakh-Daghestanian & Chechen & test\_all \\
crh & crh & Turkic & Crimean Tatar & test\_all \\
ne & nep & Indo-European & Eastern Pahari & test\_all \\
fo & fao & Indo-European & Faroese & test\_all \\
ig & ibo & Atlantic-Congo & Igbo & test\_all \\
ilo & ilo & Austronesian & Iloko & test\_all \\
jv & jav & Austronesian & Javanese & test\_all \\
rw & kin & Atlantic-Congo & Kinyarwanda & test\_all \\
mg & mlg & Austronesian & Malagasy & test\_all \\
mi & mri & Austronesian & Maori & test\_all \\
pdc & pdc & Indo-European & Pennsylvania German & test\_all \\
gd & gla & Indo-European & Scottish Gaelic & test\_all \\
sd & snd & Indo-European & Sindhi & test\_all \\
so & som & Afro-Asiatic & Somali & test\_all \\
tg & tgk & Indo-European & Tajik & test\_all \\
ug & uig & Turkic & Uighur & test\_all \\
yo & yor & Atlantic-Congo & Yoruba & test\_all \\
af & afr & Indo-European & Afrikaans & test\_all,test\_medium \\
be & bel & Indo-European & Belarusian & test\_all,test\_medium \\
bn & ben & Indo-European & Bengali & test\_all,test\_medium \\
br & bre & Indo-European & Breton & test\_all,test\_medium \\
ckb & ckb & Indo-European & Central Kurdish & test\_all,test\_medium \\
hy & hye & Indo-European & Eastern Armenian & test\_all,test\_medium \\
hi & hin & Indo-European & Hindi & test\_all,test\_medium \\
ga & gle & Indo-European & Irish & test\_all,test\_medium \\
la & lat & Indo-European & Latin & test\_all,test\_medium \\
mk & mkd & Indo-European & Macedonian & test\_all,test\_medium \\
ms & msa & Austronesian & Malay & test\_all,test\_medium \\
nn & nno & Indo-European & Norwegian Nynorsk & test\_all,test\_medium \\
tl & tgl & Austronesian & Tagalog & test\_all,test\_medium \\
ta & tam & Dravidian & Tamil & test\_all,test\_medium \\
tt & tat & Turkic & Tatar & test\_all,test\_medium \\
ur & urd & Indo-European & Urdu & test\_all,test\_medium \\
cy & cym & Indo-European & Welsh & test\_all,test\_medium \\
eu & eus & - & Basque & test\_all,test\_medium,test\_large \\
bg & bul & Indo-European & Bulgarian & test\_all,test\_medium,test\_large \\
ca & cat & Indo-European & Catalan & test\_all,test\_medium,test\_large \\
zh & zho & Sino-Tibetan & Chinese & test\_all,test\_medium,test\_large \\
hr & hrv & Indo-European & Croatian Standard & test\_all,test\_medium,test\_large \\
cs & ces & Indo-European & Czech & test\_all,test\_medium,test\_large \\
da & dan & Indo-European & Danish & test\_all,test\_medium,test\_large \\
nl & nld & Indo-European & Dutch & test\_all,test\_medium,test\_large \\
eo & epo & Artificial Language & Esperanto & test\_all,test\_medium,test\_large \\
et & est & Uralic & Estonian & test\_all,test\_medium,test\_large \\
fi & fin & Uralic & Finnish & test\_all,test\_medium,test\_large \\
fr & fra & Indo-European & French & test\_all,test\_medium,test\_large \\
de & deu & Indo-European & German & test\_all,test\_medium,test\_large \\
hu & hun & Uralic & Hungarian & test\_all,test\_medium,test\_large \\
it & ita & Indo-European & Italian & test\_all,test\_medium,test\_large \\
ja & jpn & Japonic & Japanese & test\_all,test\_medium,test\_large \\
ko & kor & Koreanic & Korean & test\_all,test\_medium,test\_large \\
lv & lav & Indo-European & Latvian & test\_all,test\_medium,test\_large \\
lt & lit & Indo-European & Lithuanian & test\_all,test\_medium,test\_large \\
el & ell & Indo-European & Modern Greek & test\_all,test\_medium,test\_large \\
pl & pol & Indo-European & Polish & test\_all,test\_medium,test\_large \\
pt & por & Indo-European & Portuguese & test\_all,test\_medium,test\_large \\
ro & ron & Indo-European & Romanian & test\_all,test\_medium,test\_large \\
ru & rus & Indo-European & Russian & test\_all,test\_medium,test\_large \\
sr & srp & Indo-European & Serbian Standard & test\_all,test\_medium,test\_large \\
sk & slk & Indo-European & Slovak & test\_all,test\_medium,test\_large \\
sl & slv & Indo-European & Slovenian & test\_all,test\_medium,test\_large \\
es & spa & Indo-European & Spanish & test\_all,test\_medium,test\_large \\
id & ind & Austronesian & Standard Indonesian & test\_all,test\_medium,test\_large \\
sv & swe & Indo-European & Swedish & test\_all,test\_medium,test\_large \\
th & tha & Tai-Kadai & Thai & test\_all,test\_medium,test\_large \\
tr & tur & Turkic & Turkish & test\_all,test\_medium,test\_large \\
uk & ukr & Indo-European & Ukrainian & test\_all,test\_medium,test\_large \\
vi & vie & Austroasiatic & Vietnamese & test\_all,test\_medium,test\_large \\
\bottomrule
\end{xltabular}}

\small{
\begin{xltabular}{\textwidth}{lllll}
\toprule
Treebank & UD File & Bible ISO & Bible File \\
\toprule
\endfirsthead
\toprule
Treebank & UD File & Bible ISO & Bible File \\
\toprule
\endhead
\toprule
&&&\multicolumn{2}{c}{\textit{Continued on next page}}  \\
\toprule
\endfoot
\bottomrule
\\ 
\caption{Mapping from UD datasets to Bible datasets.} \label{tab:pos_bible_map}
\endlastfoot
UD\_Lithuanian-ALKSNIS & lt\_alksnis-ud-dev & lit & lit-x-bible-lit-v1.txt \\
UD\_English-LinES & en\_lines-ud-dev & eng & eng-x-bible-books-v1.txt \\
UD\_Portuguese-PetroGold & pt\_petrogold-ud-dev & por & por-x-bible-almeidaatualizada-v1.txt \\
UD\_Czech-FicTree & cs\_fictree-ud-dev & ces & ces-x-bible-preklad-v1.txt \\
UD\_Portuguese-CINTIL & pt\_cintil-ud-dev & por & por-x-bible-almeidaatualizada-v1.txt \\
UD\_Czech-CLTT & cs\_cltt-ud-dev & ces & ces-x-bible-preklad-v1.txt \\
UD\_Romanian-Nonstandard & ro\_nonstandard-ud-dev & ron & ron-x-bible-cornilescu-v1.txt \\
UD\_English-ParTUT & en\_partut-ud-dev & eng & eng-x-bible-books-v1.txt \\
UD\_Maltese-MUDT & mt\_mudt-ud-dev & mlt & mlt-x-bible-mlt-v1.txt \\
UD\_Polish-PDB & pl\_pdb-ud-dev & pol & pol-x-bible-gdansk-v1.txt \\
UD\_Icelandic-Modern & is\_modern-ud-dev & isl & isl-x-bible-isl-v1.txt \\
UD\_Dutch-Alpino & nl\_alpino-ud-dev & nld & nld-x-bible-2004-v1.txt \\
UD\_English-GUM & en\_gum-ud-dev & eng & eng-x-bible-books-v1.txt \\
UD\_Turkish-Kenet & tr\_kenet-ud-dev & tur & tur-x-bible-tur-v1.txt \\
UD\_Italian-Old & it\_old-ud-dev & ita & ita-x-bible-2009-v1.txt \\
UD\_Russian-Taiga & ru\_taiga-ud-dev & rus & rus-x-bible-synodal-v1.txt \\
UD\_Ukrainian-IU & uk\_iu-ud-dev & ukr & ukr-x-bible-2007-v1.txt \\
UD\_Hindi-HDTB & hi\_hdtb-ud-dev & hin & hin-HNDSKV.txt \\
UD\_Wolof-WTB & wo\_wtb-ud-dev & wol & wol-x-bible-wol-v1.txt \\
UD\_Korean-GSD & ko\_gsd-ud-dev & kor & kor-x-bible-latinscript-v1.txt \\
UD\_Estonian-EDT & et\_edt-ud-dev & est & est-x-bible-portions-v1.txt \\
UD\_Persian-PerDT & fa\_perdt-ud-dev & fas & fas-x-bible-1995-v1.txt \\
UD\_French-GSD & fr\_gsd-ud-dev & fra & fra-FRNPDC.txt \\
UD\_Latin-ITTB & la\_ittb-ud-dev & lat & lat-LTNNVV.txt \\
UD\_Vietnamese-VTB & vi\_vtb-ud-dev & vie & vie-VIEVOV.txt \\
UD\_Latvian-LVTB & lv\_lvtb-ud-dev & lav & lav-x-bible-ljd-youversion-v1.txt \\
UD\_Finnish-FTB & fi\_ftb-ud-dev & fin & fin-x-bible-1766-v1.txt \\
UD\_Icelandic-IcePaHC & is\_icepahc-ud-dev & isl & isl-x-bible-isl-v1.txt \\
UD\_Latin-PROIEL & la\_proiel-ud-dev & lat & lat-LTNNVV.txt \\
UD\_Romanian-RRT & ro\_rrt-ud-dev & ron & ron-x-bible-cornilescu-v1.txt \\
UD\_Czech-CAC & cs\_cac-ud-dev & ces & ces-x-bible-preklad-v1.txt \\
UD\_English-ESLSpok & en\_eslspok-ud-dev & eng & eng-x-bible-books-v1.txt \\
UD\_Russian-SynTagRus & ru\_syntagrus-ud-dev & rus & rus-x-bible-synodal-v1.txt \\
UD\_Italian-ParTUT & it\_partut-ud-dev & ita & ita-x-bible-2009-v1.txt \\
UD\_Turkish-IMST & tr\_imst-ud-dev & tur & tur-x-bible-tur-v1.txt \\
UD\_Swedish-LinES & sv\_lines-ud-dev & swe & swe-SWESFV.txt \\
UD\_Russian-GSD & ru\_gsd-ud-dev & rus & rus-x-bible-synodal-v1.txt \\
UD\_Icelandic-GC & is\_gc-ud-dev & isl & isl-x-bible-isl-v1.txt \\
UD\_Persian-Seraji & fa\_seraji-ud-dev & fas & fas-x-bible-1995-v1.txt \\
UD\_Latin-UDante & la\_udante-ud-dev & lat & lat-LTNNVV.txt \\
UD\_Greek-GDT & el\_gdt-ud-dev & ell & ell-x-bible-hellenic1-v1.txt \\
UD\_Norwegian-Bokmaal & no\_bokmaal-ud-dev & nor & nor-x-bible-student-v1.txt \\
UD\_Turkish-FrameNet & tr\_framenet-ud-dev & tur & tur-x-bible-tur-v1.txt \\
UD\_Swedish-Talbanken & sv\_talbanken-ud-dev & swe & swe-SWESFV.txt \\
UD\_Danish-DDT & da\_ddt-ud-dev & dan & dan-x-bible-1931-v1.txt \\
UD\_Italian-ISDT & it\_isdt-ud-dev & ita & ita-x-bible-2009-v1.txt \\
UD\_Slovak-SNK & sk\_snk-ud-dev & slk & slk-x-bible-standard-v1.txt \\
UD\_Latin-LLCT & la\_llct-ud-dev & lat & lat-LTNNVV.txt \\
UD\_English-EWT & en\_ewt-ud-dev & eng & eng-x-bible-books-v1.txt \\
UD\_Welsh-CCG & cy\_ccg-ud-dev & cym & cym-x-bible-revised2004-v1.txt \\
UD\_Portuguese-DANTEStocks & pt\_dantestocks-ud-dev & por & por-x-bible-almeidaatualizada-v1.txt \\
UD\_Hebrew-IAHLTknesset & he\_iahltknesset-ud-dev & heb & heb-x-bible-2009-v1.txt \\
UD\_Portuguese-Porttinari & pt\_porttinari-ud-dev & por & por-x-bible-almeidaatualizada-v1.txt \\
UD\_Hungarian-Szeged & hu\_szeged-ud-dev & hun & hun-x-bible-revised-v1.txt \\
UD\_Russian-Poetry & ru\_poetry-ud-dev & rus & rus-x-bible-synodal-v1.txt \\
UD\_Catalan-AnCora & ca\_ancora-ud-dev & cat & cat-x-bible-cat-v1.txt \\
UD\_French-ParTUT & fr\_partut-ud-dev & fra & fra-FRNPDC.txt \\
UD\_Italian-VIT & it\_vit-ud-dev & ita & ita-x-bible-2009-v1.txt \\
UD\_German-GSD & de\_gsd-ud-dev & deu & deu-x-bible-freebible-v1.txt \\
UD\_Armenian-BSUT & hy\_bsut-ud-dev & hye & hye-x-bible-eastern-v1.txt \\
UD\_Lithuanian-HSE & lt\_hse-ud-dev & lit & lit-x-bible-lit-v1.txt \\
UD\_English-GUMReddit & en\_gumreddit-ud-dev & eng & eng-x-bible-books-v1.txt \\
UD\_Italian-PoSTWITA & it\_postwita-ud-dev & ita & ita-x-bible-2009-v1.txt \\
UD\_Korean-KSL & ko\_ksl-ud-dev & kor & kor-x-bible-latinscript-v1.txt \\
UD\_Spanish-AnCora & es\_ancora-ud-dev & spa & spa-SPNBDA.txt \\
UD\_Portuguese-GSD & pt\_gsd-ud-dev & por & por-x-bible-almeidaatualizada-v1.txt \\
UD\_Portuguese-Bosque & pt\_bosque-ud-dev & por & por-x-bible-almeidaatualizada-v1.txt \\
UD\_Polish-LFG & pl\_lfg-ud-dev & pol & pol-x-bible-gdansk-v1.txt \\
UD\_Czech-PDT & cs\_pdt-ud-dev & ces & ces-x-bible-preklad-v1.txt \\
UD\_Turkish-Atis & tr\_atis-ud-dev & tur & tur-x-bible-tur-v1.txt \\
UD\_Finnish-TDT & fi\_tdt-ud-dev & fin & fin-x-bible-1766-v1.txt \\
UD\_Italian-MarkIT & it\_markit-ud-dev & ita & ita-x-bible-2009-v1.txt \\
UD\_Romanian-SiMoNERo & ro\_simonero-ud-dev & ron & ron-x-bible-cornilescu-v1.txt \\
UD\_German-HDT & de\_hdt-ud-dev & deu & deu-x-bible-freebible-v1.txt \\
UD\_Hebrew-IAHLTwiki & he\_iahltwiki-ud-dev & heb & heb-x-bible-2009-v1.txt \\
UD\_French-Sequoia & fr\_sequoia-ud-dev & fra & fra-FRNPDC.txt \\
UD\_Estonian-EWT & et\_ewt-ud-dev & est & est-x-bible-portions-v1.txt \\
UD\_Uyghur-UDT & ug\_udt-ud-dev & uig & uig-x-bible-uig-v1.txt \\
UD\_Italian-TWITTIRO & it\_twittiro-ud-dev & ita & ita-x-bible-2009-v1.txt \\
UD\_Slovenian-SSJ & sl\_ssj-ud-dev & slv & slv-x-bible-slv-v1.txt \\
UD\_English-Atis & en\_atis-ud-dev & eng & eng-x-bible-books-v1.txt \\
UD\_Armenian-ArmTDP & hy\_armtdp-ud-dev & hye & hye-x-bible-eastern-v1.txt \\
UD\_Korean-Kaist & ko\_kaist-ud-dev & kor & kor-x-bible-latinscript-v1.txt \\
UD\_Serbian-SET & sr\_set-ud-dev & srp & srp-x-bible-srp-v1.txt \\
UD\_Slovenian-SST & sl\_sst-ud-dev & slv & slv-x-bible-slv-v1.txt \\
UD\_Hebrew-HTB & he\_htb-ud-dev & heb & heb-x-bible-2009-v1.txt \\
UD\_Old\_Church\_Slavonic-PROIEL & cu\_proiel-ud-dev & chu & chu-x-bible-chu-v1.txt \\
UD\_Urdu-UDTB & ur\_udtb-ud-dev & urd & urd-x-bible-revised2010-v1.txt \\
UD\_Norwegian-Nynorsk & no\_nynorsk-ud-dev & nor & nor-x-bible-student-v1.txt \\
UD\_Turkish-BOUN & tr\_boun-ud-dev & tur & tur-x-bible-tur-v1.txt \\
UD\_Bulgarian-BTB & bg\_btb-ud-dev & bul & bul-x-bible-veren-v1.txt \\
UD\_Indonesian-GSD & id\_gsd-ud-dev & ind & ind-x-bible-suciinjil-v1.txt \\
UD\_Dutch-LassySmall & nl\_lassysmall-ud-dev & nld & nld-x-bible-2004-v1.txt \\
UD\_Turkish-Penn & tr\_penn-ud-dev & tur & tur-x-bible-tur-v1.txt \\
UD\_Georgian-GLC & ka\_glc-ud-dev & kat & kat-x-bible-kat-v1.txt \\
UD\_Ukrainian-ParlaMint & uk\_parlamint-ud-dev & ukr & ukr-x-bible-2007-v1.txt \\
UD\_Afrikaans-AfriBooms & af\_afribooms-ud-dev & afr & afr-x-bible-1953-v1.txt \\
UD\_Spanish-GSD & es\_gsd-ud-dev & spa & spa-SPNBDA.txt \\
UD\_Basque-BDT & eu\_bdt-ud-dev & eus & eus-x-bible-Hautin1571-v1.txt \\
UD\_French-ParisStories & fr\_parisstories-ud-dev & fra & fra-FRNPDC.txt \\
UD\_French-Rhapsodie & fr\_rhapsodie-ud-dev & fra & fra-FRNPDC.txt \\
UD\_Tamil-TTB & ta\_ttb-ud-dev & tam & tam-x-bible-tam-v1.txt \\
UD\_Croatian-SET & hr\_set-ud-dev & hrv & hrv-x-bible-hrv-v1.txt \\
UD\_Turkish-Tourism & tr\_tourism-ud-dev & tur & tur-x-bible-tur-v1.txt \\
UD\_English-LinES & en\_lines-ud-train & eng & eng-x-bible-books-v1.txt \\
UD\_Czech-FicTree & cs\_fictree-ud-train & ces & ces-x-bible-preklad-v1.txt \\
UD\_Czech-CLTT & cs\_cltt-ud-train & ces & ces-x-bible-preklad-v1.txt \\
UD\_Romanian-Nonstandard & ro\_nonstandard-ud-train & ron & ron-x-bible-cornilescu-v1.txt \\
UD\_English-ParTUT & en\_partut-ud-train & eng & eng-x-bible-books-v1.txt \\
UD\_Dutch-Alpino & nl\_alpino-ud-train & nld & nld-x-bible-2004-v1.txt \\
UD\_English-GUM & en\_gum-ud-train & eng & eng-x-bible-books-v1.txt \\
UD\_Russian-Taiga & ru\_taiga-ud-train & rus & rus-x-bible-synodal-v1.txt \\
UD\_Ukrainian-IU & uk\_iu-ud-train & ukr & ukr-x-bible-2007-v1.txt \\
UD\_Hindi-HDTB & hi\_hdtb-ud-train & hin & hin-HNDSKV.txt \\
UD\_Korean-GSD & ko\_gsd-ud-train & kor & kor-x-bible-latinscript-v1.txt \\
UD\_Estonian-EDT & et\_edt-ud-train & est & est-x-bible-portions-v1.txt \\
UD\_French-GSD & fr\_gsd-ud-train & fra & fra-FRNPDC.txt \\
UD\_Latin-ITTB & la\_ittb-ud-train & lat & lat-LTNNVV.txt \\
UD\_Vietnamese-VTB & vi\_vtb-ud-train & vie & vie-VIEVOV.txt \\
UD\_Latvian-LVTB & lv\_lvtb-ud-train & lav & lav-x-bible-ljd-youversion-v1.txt \\
UD\_Finnish-FTB & fi\_ftb-ud-train & fin & fin-x-bible-1766-v1.txt \\
UD\_Latin-PROIEL & la\_proiel-ud-train & lat & lat-LTNNVV.txt \\
UD\_Romanian-RRT & ro\_rrt-ud-train & ron & ron-x-bible-cornilescu-v1.txt \\
UD\_Czech-CAC & cs\_cac-ud-train & ces & ces-x-bible-preklad-v1.txt \\
UD\_Russian-SynTagRus & ru\_syntagrus-ud-train & rus & rus-x-bible-synodal-v1.txt \\
UD\_Italian-ParTUT & it\_partut-ud-train & ita & ita-x-bible-2009-v1.txt \\
UD\_Turkish-IMST & tr\_imst-ud-train & tur & tur-x-bible-tur-v1.txt \\
UD\_Swedish-LinES & sv\_lines-ud-train & swe & swe-SWESFV.txt \\
UD\_Russian-GSD & ru\_gsd-ud-train & rus & rus-x-bible-synodal-v1.txt \\
UD\_Persian-Seraji & fa\_seraji-ud-train & fas & fas-x-bible-1995-v1.txt \\
UD\_Greek-GDT & el\_gdt-ud-train & ell & ell-x-bible-hellenic1-v1.txt \\
UD\_Norwegian-Bokmaal & no\_bokmaal-ud-train & nor & nor-x-bible-student-v1.txt \\
UD\_Swedish-Talbanken & sv\_talbanken-ud-train & swe & swe-SWESFV.txt \\
UD\_Danish-DDT & da\_ddt-ud-train & dan & dan-x-bible-1931-v1.txt \\
UD\_Italian-ISDT & it\_isdt-ud-train & ita & ita-x-bible-2009-v1.txt \\
UD\_Slovak-SNK & sk\_snk-ud-train & slk & slk-x-bible-standard-v1.txt \\
UD\_English-EWT & en\_ewt-ud-train & eng & eng-x-bible-books-v1.txt \\
UD\_Hungarian-Szeged & hu\_szeged-ud-train & hun & hun-x-bible-revised-v1.txt \\
UD\_Catalan-AnCora & ca\_ancora-ud-train & cat & cat-x-bible-cat-v1.txt \\
UD\_French-ParTUT & fr\_partut-ud-train & fra & fra-FRNPDC.txt \\
UD\_German-GSD & de\_gsd-ud-train & deu & deu-x-bible-freebible-v1.txt \\
UD\_Italian-PoSTWITA & it\_postwita-ud-train & ita & ita-x-bible-2009-v1.txt \\
UD\_Spanish-AnCora & es\_ancora-ud-train & spa & spa-SPNBDA.txt \\
UD\_Portuguese-GSD & pt\_gsd-ud-train & por & por-x-bible-almeidaatualizada-v1.txt \\
UD\_Portuguese-Bosque & pt\_bosque-ud-train & por & por-x-bible-almeidaatualizada-v1.txt \\
UD\_Polish-LFG & pl\_lfg-ud-train & pol & pol-x-bible-gdansk-v1.txt \\
UD\_Latin-Perseus & la\_perseus-ud-train & lat & lat-LTNNVV.txt \\
UD\_Czech-PDT & cs\_pdt-ud-train & ces & ces-x-bible-preklad-v1.txt \\
UD\_Finnish-TDT & fi\_tdt-ud-train & fin & fin-x-bible-1766-v1.txt \\
UD\_French-Sequoia & fr\_sequoia-ud-train & fra & fra-FRNPDC.txt \\
UD\_Uyghur-UDT & ug\_udt-ud-train & uig & uig-x-bible-uig-v1.txt \\
UD\_Slovenian-SSJ & sl\_ssj-ud-train & slv & slv-x-bible-slv-v1.txt \\
UD\_Armenian-ArmTDP & hy\_armtdp-ud-train & hye & hye-x-bible-eastern-v1.txt \\
UD\_Korean-Kaist & ko\_kaist-ud-train & kor & kor-x-bible-latinscript-v1.txt \\
UD\_Serbian-SET & sr\_set-ud-train & srp & srp-x-bible-srp-v1.txt \\
UD\_Slovenian-SST & sl\_sst-ud-train & slv & slv-x-bible-slv-v1.txt \\
UD\_Hebrew-HTB & he\_htb-ud-train & heb & heb-x-bible-2009-v1.txt \\
UD\_Old\_Church\_Slavonic-PROIEL & cu\_proiel-ud-train & chu & chu-x-bible-chu-v1.txt \\
UD\_Urdu-UDTB & ur\_udtb-ud-train & urd & urd-x-bible-revised2010-v1.txt \\
UD\_Norwegian-Nynorsk & no\_nynorsk-ud-train & nor & nor-x-bible-student-v1.txt \\
UD\_Bulgarian-BTB & bg\_btb-ud-train & bul & bul-x-bible-veren-v1.txt \\
UD\_Indonesian-GSD & id\_gsd-ud-train & ind & ind-x-bible-suciinjil-v1.txt \\
UD\_Dutch-LassySmall & nl\_lassysmall-ud-train & nld & nld-x-bible-2004-v1.txt \\
UD\_Afrikaans-AfriBooms & af\_afribooms-ud-train & afr & afr-x-bible-1953-v1.txt \\
UD\_Spanish-GSD & es\_gsd-ud-train & spa & spa-SPNBDA.txt \\
UD\_Basque-BDT & eu\_bdt-ud-train & eus & eus-x-bible-Hautin1571-v1.txt \\
UD\_Tamil-TTB & ta\_ttb-ud-train & tam & tam-x-bible-tam-v1.txt \\
UD\_Croatian-SET & hr\_set-ud-train & hrv & hrv-x-bible-hrv-v1.txt \\
\bottomrule
\end{xltabular}}

\small{
\begin{xltabular}{\textwidth}{lll}
\toprule
Rahimi ISO & Bible ISO & Bible File \\
\toprule
\endfirsthead
\toprule
Rahimi ISO & Bible ISO & Bible File \\
\toprule
\endhead
\toprule
&\multicolumn{2}{c}{\textit{Continued on next page}}  \\
\toprule
\endfoot
\bottomrule
\\ 
\caption{Mapping from \cite{rahimiMassivelyMultilingualTransfer2019} datasets to Bible datasets.} \label{tab:ner_bible_data_map}
\endlastfoot

it & ita & ita-x-bible-riveduta-v1.txt \\
sr & srp & srp-x-bible-srp-v1.txt \\
sl & slv & slv-x-bible-slv-v1.txt \\
so & som & som-SOMSIM.txt \\
ko & kor & kor-x-bible-kor-v1.txt \\
crh & crh & crh-CRHIBT.txt \\
cy & cym & cym-x-bible-colloquial2013-v1.txt \\
eo & epo & epo-x-bible-epo-v1.txt \\
pt & por & por-PORARC.txt \\
ta & tam & tam-TCVWTC.txt \\
es & spa & spa-SPNWTC.txt \\
la & lat & lat-x-bible-vulgataclementina-v1.txt \\
ceb & ceb & ceb-x-bible-popular-v1.txt \\
et & est & est-x-bible-portions-v1.txt \\
yo & yor & yor-x-bible-yor-v1.txt \\
br & bre & bre-x-bible-bre-v1.txt \\
fi & fin & fin-x-bible-1766-v1.txt \\
eu & eus & eus-x-bible-batua-v1.txt \\
hi & hin & hin-HNDSKV.txt \\
fr & fra & fra-x-bible-kingjames-v1.txt \\
ug & uig & uig-UI1UMK.txt \\
lv & lav & lav-x-bible-1997-v1.txt \\
ilo & ilo & ilo-x-bible-ilo-v1.txt \\
ce & che & che-CHEIBT.txt \\
tl & tgl & tgl-TGLPBS.txt \\
nl & nld & nld-x-bible-2007-v1.txt \\
rw & kin & kin-x-bible-bird-youversion-v1.txt \\
mg & mlg & mlg-MLGRCV.txt \\
uk & ukr & ukr-x-bible-2009-v1.txt \\
mk & mkd & mkd-x-bible-2004-v1.txt \\
ur & urd & urd-x-bible-devanagari-v1.txt \\
de & deu & deu-x-bible-greber-v1.txt \\
id & ind & ind-INZNTV.txt \\
el & ell & ell-x-bible-hellenic1-v1.txt \\
am & amh & amh-x-bible-amh-v1.txt \\
ru & rus & rus-x-bible-kulakov-v1.txt \\
af & afr & afr-x-bible-boodskap-v1.txt \\
pl & pol & pol-x-bible-gdansk-v1.txt \\
da & dan & dan-x-bible-1931-v1.txt \\
bg & bul & bul-x-bible-veren-v1.txt \\
my & mya & mya-x-bible-common-v1.txt \\
vi & vie & vie-x-bible-bd2011-youversion-v1.txt \\
tt & tat & tat-TTRIBT.txt \\
tg & tgk & tgk-TGKIBT.txt \\
sv & swe & swe-SWESFV.txt \\
hu & hun & hun-x-bible-revised-v1.txt \\
hy & hye & hye-x-bible-eastern-v1.txt \\
th & tha & tha-THATSV.txt \\
ig & ibo & ibo-x-bible-ibo-v1.txt \\
jv & jav & jav-x-bible-jav-v1.txt \\
nn & nno & nno-x-bible-2011-v1.txt \\
bn & ben & ben-x-bible-common-v1.txt \\
mi & mri & mri-x-bible-mri-v1.txt \\
lt & lit & lit-x-bible-1999-v1.txt \\
ro & ron & ron-RONBSR.txt \\
ca & cat & cat-x-bible-cat-v1.txt \\
tr & tur & tur-TRKBST.txt \\
sk & slk & slk-x-bible-standard-v1.txt \\
cs & ces & ces-x-bible-novakarlica-v1.txt \\
hr & hrv & hrv-x-bible-hrv-v1.txt \\
km & khm & khm-x-bible-2011-v1.txt \\
ms & msa & msa-x-bible-1996-v1.txt \\
\bottomrule
\end{xltabular}}

\small{
\begin{xltabular}{\textwidth}{llll}
\toprule
Source & \multicolumn{2}{l}{Target Tokens} & \multicolumn{1}{l}{Num. Examples} \\
 & \multicolumn{1}{l}{mBERT} & \multicolumn{1}{l}{XLM-R} & \multicolumn{1}{l}{} \\
\toprule
\endfirsthead
\toprule
Source & \multicolumn{2}{l}{Target Tokens} & \multicolumn{1}{l}{Num. Examples} \\
 & \multicolumn{1}{l}{mBERT} & \multicolumn{1}{l}{XLM-R} & \multicolumn{1}{l}{} \\
\toprule
\endhead
\toprule
&\multicolumn{2}{c}{\textit{Continued on next page}}  \\
\toprule
\endfoot
\bottomrule
\\ 
\caption{POS Target Token Counts} \label{tab:pos_target_token_counts}
\endlastfoot

af\_afribooms-ud-dev & 8521 & 7515 & 398 \\
ar\_nyuad-ud-dev & 40000 & 40000 & 1000 \\
ar\_padt-ud-dev & 24390 & 18741 & 1000 \\
be\_hse-ud-dev & 26740 & 20806 & 1000 \\
bg\_btb-ud-dev & 25337 & 19559 & 1000 \\
ca\_ancora-ud-dev & 20738 & 19755 & 1000 \\
cs\_cac-ud-dev & 22820 & 18853 & 856 \\
cs\_cltt-ud-dev & 24100 & 19053 & 917 \\
cs\_fictree-ud-dev & 27022 & 23007 & 1000 \\
cs\_pdt-ud-dev & 21609 & 17678 & 1000 \\
cu\_proiel-ud-dev & 22228 & 39918 & 1000 \\
cy\_ccg-ud-dev & 16210 & 13478 & 564 \\
da\_ddt-ud-dev & 15924 & 14106 & 711 \\
de\_gsd-ud-dev & 17983 & 16517 & 944 \\
de\_hdt-ud-dev & 18014 & 18195 & 1000 \\
el\_gdt-ud-dev & 24439 & 16737 & 799 \\
en\_atis-ud-dev & 8875 & 9388 & 478 \\
en\_eslspok-ud-dev & 2420 & 2697 & 122 \\
en\_ewt-ud-dev & 17333 & 16780 & 1000 \\
en\_gum-ud-dev & 15137 & 15423 & 1000 \\
en\_gumreddit-ud-dev & 1755 & 1555 & 71 \\
en\_lines-ud-dev & 18652 & 18990 & 1000 \\
en\_partut-ud-dev & 3181 & 3351 & 183 \\
es\_ancora-ud-dev & 19084 & 18510 & 1000 \\
es\_gsd-ud-dev & 18506 & 18786 & 1000 \\
et\_edt-ud-dev & 25452 & 21031 & 1000 \\
et\_ewt-ud-dev & 18747 & 15140 & 699 \\
eu\_bdt-ud-dev & 23246 & 20001 & 1000 \\
fa\_perdt-ud-dev & 27040 & 22247 & 1000 \\
fa\_seraji-ud-dev & 24608 & 20241 & 962 \\
fi\_ftb-ud-dev & 23579 & 20469 & 1000 \\
fi\_tdt-ud-dev & 23211 & 18989 & 1000 \\
fo\_farpahc-ud-dev & 15190 & 13542 & 521 \\
fr\_gsd-ud-dev & 19507 & 20455 & 1000 \\
fr\_parisstories-ud-dev & 12645 & 12226 & 521 \\
fr\_partut-ud-dev & 2408 & 2401 & 124 \\
fr\_rhapsodie-ud-dev & 16030 & 15763 & 684 \\
fr\_sequoia-ud-dev & 13779 & 13514 & 679 \\
ga\_idt-ud-dev & 18391 & 15299 & 643 \\
ga\_twittirish-ud-dev & 31893 & 26959 & 1000 \\
gd\_arcosg-ud-dev & 20167 & 17136 & 669 \\
gl\_ctg-ud-dev & 18628 & 17014 & 1000 \\
he\_htb-ud-dev & 16549 & 13905 & 522 \\
he\_iahltknesset-ud-dev & 8838 & 7217 & 317 \\
he\_iahltwiki-ud-dev & 13857 & 11748 & 514 \\
hi\_hdtb-ud-dev & 30255 & 21135 & 1000 \\
hr\_set-ud-dev & 22897 & 19128 & 1000 \\
hu\_szeged-ud-dev & 24410 & 18650 & 1000 \\
hy\_armtdp-ud-dev & 13562 & 9144 & 456 \\
hy\_bsut-ud-dev & 26762 & 16933 & 977 \\
id\_gsd-ud-dev & 17783 & 16168 & 978 \\
is\_gc-ud-dev & 21888 & 16661 & 786 \\
is\_icepahc-ud-dev & 29366 & 23409 & 1000 \\
is\_modern-ud-dev & 15990 & 12184 & 565 \\
it\_isdt-ud-dev & 15137 & 14639 & 782 \\
it\_markit-ud-dev & 12895 & 12992 & 705 \\
it\_old-ud-dev & 16504 & 16190 & 686 \\
it\_partut-ud-dev & 3812 & 3665 & 200 \\
it\_postwita-ud-dev & 21299 & 19585 & 854 \\
it\_twittiro-ud-dev & 4871 & 4577 & 202 \\
it\_vit-ud-dev & 18397 & 17271 & 1000 \\
ja\_bccwj-ud-dev & 80653 & 6395 & 1000 \\
ja\_bccwjluw-ud-dev & 80586 & 6443 & 1000 \\
ja\_gsd-ud-dev & 16004 & 12800 & 538 \\
ja\_gsdluw-ud-dev & 16004 & 12800 & 538 \\
ka\_glc-ud-dev & 31320 & 18913 & 895 \\
ko\_gsd-ud-dev & 27817 & 24689 & 561 \\
ko\_kaist-ud-dev & 48999 & 42660 & 1000 \\
ko\_ksl-ud-dev & 13806 & 11485 & 288 \\
la\_ittb-ud-dev & 22941 & 19384 & 1000 \\
la\_llct-ud-dev & 24171 & 20836 & 1000 \\
la\_proiel-ud-dev & 23033 & 19301 & 1000 \\
la\_udante-ud-dev & 20738 & 18207 & 887 \\
lt\_alksnis-ud-dev & 25320 & 19442 & 985 \\
lt\_hse-ud-dev & 2205 & 1822 & 108 \\
lv\_lvtb-ud-dev & 24574 & 19819 & 1000 \\
mt\_mudt-ud-dev & 22456 & 22470 & 715 \\
nl\_alpino-ud-dev & 16672 & 16126 & 825 \\
nl\_lassysmall-ud-dev & 20204 & 19797 & 1000 \\
no\_bokmaal-ud-dev & 20175 & 17823 & 1000 \\
no\_nynorsk-ud-dev & 21417 & 20205 & 1000 \\
pl\_lfg-ud-dev & 24459 & 20574 & 921 \\
pl\_pdb-ud-dev & 25627 & 21685 & 1000 \\
pt\_bosque-ud-dev & 20164 & 19222 & 1000 \\
pt\_cintil-ud-dev & 19545 & 20356 & 1000 \\
pt\_dantestocks-ud-dev & 20825 & 18335 & 673 \\
pt\_gsd-ud-dev & 19702 & 18554 & 1000 \\
pt\_petrogold-ud-dev & 19890 & 18738 & 1000 \\
pt\_porttinari-ud-dev & 20162 & 19148 & 1000 \\
ro\_nonstandard-ud-dev & 30543 & 26920 & 1000 \\
ro\_rrt-ud-dev & 23287 & 20163 & 1000 \\
ro\_simonero-ud-dev & 23378 & 19727 & 1000 \\
ru\_gsd-ud-dev & 21848 & 19077 & 932 \\
ru\_poetry-ud-dev & 19372 & 16107 & 869 \\
ru\_syntagrus-ud-dev & 22194 & 19571 & 1000 \\
ru\_taiga-ud-dev & 19948 & 17288 & 755 \\
sa\_vedic-ud-dev & 35281 & 32561 & 1000 \\
sk\_snk-ud-dev & 25869 & 21288 & 985 \\
sl\_ssj-ud-dev & 23134 & 19058 & 1000 \\
sl\_sst-ud-dev & 15180 & 15014 & 615 \\
sr\_set-ud-dev & 20676 & 17723 & 896 \\
sv\_lines-ud-dev & 22032 & 19407 & 1000 \\
sv\_talbanken-ud-dev & 16240 & 13716 & 794 \\
ta\_ttb-ud-dev & 3597 & 2188 & 120 \\
tr\_atis-ud-dev & 10765 & 8658 & 481 \\
tr\_boun-ud-dev & 25020 & 19336 & 984 \\
tr\_framenet-ud-dev & 2998 & 2616 & 115 \\
tr\_imst-ud-dev & 21716 & 16889 & 863 \\
tr\_kenet-ud-dev & 25741 & 22390 & 1000 \\
tr\_penn-ud-dev & 14326 & 11950 & 594 \\
tr\_tourism-ud-dev & 38899 & 28671 & 801 \\
ug\_udt-ud-dev & 25272 & 22254 & 863 \\
uk\_iu-ud-dev & 23320 & 18934 & 1000 \\
uk\_parlamint-ud-dev & 20090 & 15537 & 997 \\
ur\_udtb-ud-dev & 25949 & 19464 & 847 \\
vi\_vtb-ud-dev & 22978 & 22816 & 1000 \\
wo\_wtb-ud-dev & 17495 & 17275 & 553 \\
zh\_gsd-ud-dev & 19099 & 14906 & 248 \\
zh\_gsdsimp-ud-dev & 19099 & 14553 & 248\\
\bottomrule
\end{xltabular}}

\small{
\begin{xltabular}{\textwidth}{llll}
\toprule
Source & \multicolumn{2}{l}{Source Tokens} & \multicolumn{1}{l}{Num. Examples} \\
 & \multicolumn{1}{l}{mBERT} & \multicolumn{1}{l}{XLM-R} & \multicolumn{1}{l}{} \\
\toprule
\endfirsthead
\toprule
Source & \multicolumn{2}{l}{Source Tokens} & \multicolumn{1}{l}{Num. Examples} \\
 & \multicolumn{1}{l}{mBERT} & \multicolumn{1}{l}{XLM-R} & \multicolumn{1}{l}{} \\
\toprule
\endhead
\toprule
&\multicolumn{2}{c}{\textit{Continued on next page}}  \\
\toprule
\endfoot
\bottomrule
\\ 
\caption{POS Source Token Counts} \label{tab:pos_source_token_counts}
\endlastfoot
af\_afribooms-ud-train & 22032 & 19344 & 1000 \\
ar\_nyuad-ud-train & 40000 & 40000 & 1000 \\
ar\_padt-ud-train & 24926 & 19355 & 1000 \\
be\_hse-ud-train & 30807 & 23987 & 1000 \\
bg\_btb-ud-train & 27479 & 21340 & 1000 \\
ca\_ancora-ud-train & 20705 & 19615 & 1000 \\
cop\_scriptorium-ud-train & 12065 & 22713 & 1000 \\
cs\_cac-ud-train & 25907 & 20818 & 1000 \\
cs\_cltt-ud-train & 26619 & 21440 & 1000 \\
cs\_fictree-ud-train & 26956 & 22541 & 1000 \\
cs\_pdt-ud-train & 21379 & 17393 & 1000 \\
cu\_proiel-ud-train & 31190 & 39693 & 1000 \\
da\_ddt-ud-train & 22707 & 20025 & 1000 \\
de\_gsd-ud-train & 19558 & 17229 & 1000 \\
el\_gdt-ud-train & 31139 & 20363 & 1000 \\
en\_ewt-ud-train & 17188 & 17915 & 1000 \\
en\_gum-ud-train & 15677 & 16395 & 1000 \\
en\_lines-ud-train & 18533 & 18951 & 1000 \\
en\_partut-ud-train & 16296 & 16640 & 1000 \\
es\_ancora-ud-train & 18108 & 17223 & 1000 \\
es\_gsd-ud-train & 18481 & 18543 & 1000 \\
et\_edt-ud-train & 25274 & 20247 & 1000 \\
eu\_bdt-ud-train & 23293 & 20023 & 1000 \\
fa\_seraji-ud-train & 26436 & 22096 & 1000 \\
fi\_ftb-ud-train & 23958 & 21113 & 1000 \\
fi\_tdt-ud-train & 24186 & 19118 & 1000 \\
fr\_gsd-ud-train & 19488 & 20265 & 1000 \\
fr\_partut-ud-train & 18435 & 18196 & 1000 \\
fr\_sequoia-ud-train & 20856 & 20610 & 1000 \\
ga\_idt-ud-train & 28360 & 23489 & 1000 \\
gl\_ctg-ud-train & 18809 & 17081 & 1000 \\
gl\_treegal-ud-train & 19181 & 17820 & 991 \\
got\_proiel-ud-train & 29054 & 27840 & 1000 \\
grc\_perseus-ud-train & 39994 & 44806 & 1000 \\
grc\_proiel-ud-train & 41031 & 41922 & 1000 \\
he\_htb-ud-train & 32107 & 26899 & 1000 \\
hi\_hdtb-ud-train & 30541 & 22278 & 1000 \\
hr\_set-ud-train & 23100 & 19243 & 1000 \\
hu\_szeged-ud-train & 24914 & 20141 & 1000 \\
hy\_armtdp-ud-train & 30744 & 21104 & 1000 \\
id\_gsd-ud-train & 18133 & 16515 & 1000 \\
it\_isdt-ud-train & 20357 & 19516 & 1000 \\
it\_partut-ud-train & 17912 & 16512 & 1000 \\
it\_postwita-ud-train & 24636 & 22727 & 1000 \\
ja\_bccwj-ud-train & 80347 & 6827 & 1000 \\
ja\_gsd-ud-train & 29394 & 23727 & 1000 \\
ko\_gsd-ud-train & 49639 & 44018 & 1000 \\
ko\_kaist-ud-train & 48678 & 43974 & 1000 \\
la\_ittb-ud-train & 22431 & 19385 & 1000 \\
la\_perseus-ud-train & 24084 & 21151 & 1000 \\
la\_proiel-ud-train & 23178 & 19169 & 1000 \\
lv\_lvtb-ud-train & 26548 & 21311 & 1000 \\
nl\_alpino-ud-train & 20144 & 19048 & 1000 \\
nl\_lassysmall-ud-train & 19747 & 19830 & 1000 \\
no\_bokmaal-ud-train & 19767 & 17679 & 1000 \\
no\_nynorsk-ud-train & 21361 & 19908 & 1000 \\
pl\_lfg-ud-train & 26914 & 22629 & 1000 \\
pt\_bosque-ud-train & 20416 & 19109 & 1000 \\
pt\_gsd-ud-train & 19664 & 18483 & 1000 \\
ro\_nonstandard-ud-train & 32138 & 28198 & 1000 \\
ro\_rrt-ud-train & 25965 & 22750 & 1000 \\
ru\_gsd-ud-train & 23059 & 20180 & 1000 \\
ru\_syntagrus-ud-train & 25673 & 21795 & 1000 \\
ru\_taiga-ud-train & 21552 & 17830 & 1000 \\
sk\_snk-ud-train & 28149 & 23430 & 1000 \\
sl\_ssj-ud-train & 23635 & 19730 & 1000 \\
sl\_sst-ud-train & 24226 & 24141 & 1000 \\
sme\_giella-ud-train & 31075 & 30260 & 1000 \\
sr\_set-ud-train & 23153 & 19811 & 1000 \\
sv\_lines-ud-train & 21788 & 19220 & 1000 \\
sv\_talbanken-ud-train & 19128 & 16001 & 1000 \\
ta\_ttb-ud-train & 18592 & 11484 & 609 \\
tr\_imst-ud-train & 25381 & 19909 & 1000 \\
ug\_udt-ud-train & 29758 & 25737 & 1000 \\
uk\_iu-ud-train & 25311 & 21342 & 1000 \\
ur\_udtb-ud-train & 30512 & 22656 & 1000 \\
vi\_vtb-ud-train & 22842 & 22841 & 1000 \\
zh\_gsd-ud-train & 76547 & 60503 & 1000 \\
\bottomrule
\end{xltabular}}

\small{
\begin{xltabular}{\textwidth}{llll}
\toprule
Target Lang & \multicolumn{2}{l}{Target Tokens} & \multicolumn{1}{l}{Num. Lines} \\
 & \multicolumn{1}{l}{mBERT} & \multicolumn{1}{l}{XLM-R} & \multicolumn{1}{l}{} \\
\toprule
\endfirsthead
\toprule
Target Lang & \multicolumn{2}{l}{Target Tokens} & \multicolumn{1}{l}{Num. Lines} \\
 & \multicolumn{1}{l}{mBERT} & \multicolumn{1}{l}{XLM-R} & \multicolumn{1}{l}{} \\
\toprule
\endhead
\toprule
&\multicolumn{2}{c}{\textit{Continued on next page}}  \\
\toprule
\endfoot
\bottomrule
\\ 
\caption{NER Target Token Counts} \label{tab:ner_target_token_counts}
\endlastfoot
af\_dev & 17736 & 18092 & 1000 \\
am\_dev & 669 & 1296 & 100 \\
be\_dev & 16748 & 14468 & 1000 \\
bg\_dev & 15589 & 14525 & 1000 \\
bn\_dev & 13388 & 10384 & 1000 \\
br\_dev & 12928 & 13464 & 1000 \\
ca\_dev & 9460 & 9566 & 1000 \\
ce\_dev & 2137 & 2838 & 100 \\
ceb\_dev & 1106 & 1372 & 100 \\
ckb\_dev & 8342 & 16355 & 1000 \\
crh\_dev & 1376 & 1284 & 100 \\
cs\_dev & 14561 & 14109 & 1000 \\
cy\_dev & 15155 & 14544 & 1000 \\
da\_dev & 13352 & 13423 & 1000 \\
de\_dev & 14628 & 16019 & 1000 \\
el\_dev & 23649 & 18368 & 1000 \\
eo\_dev & 12316 & 11864 & 1000 \\
es\_dev & 10755 & 10384 & 1000 \\
et\_dev & 16242 & 15177 & 1000 \\
eu\_dev & 15957 & 17482 & 1000 \\
fi\_dev & 16814 & 16361 & 1000 \\
fo\_dev & 1832 & 1760 & 100 \\
fr\_dev & 10486 & 11192 & 1000 \\
ga\_dev & 15008 & 14162 & 1000 \\
gd\_dev & 1634 & 1537 & 100 \\
hi\_dev & 14286 & 10476 & 1000 \\
hr\_dev & 14429 & 13888 & 1000 \\
hu\_dev & 16757 & 15377 & 1000 \\
hy\_dev & 21359 & 15959 & 1000 \\
id\_dev & 10086 & 9982 & 1000 \\
ig\_dev & 1124 & 1165 & 100 \\
ilo\_dev & 808 & 834 & 100 \\
it\_dev & 12340 & 12520 & 1000 \\
ja\_dev & 29334 & 50743 & 1000 \\
jv\_dev & 1001 & 994 & 100 \\
km\_dev & 553 & 1897 & 100 \\
ko\_dev & 17710 & 17560 & 1000 \\
la\_dev & 10895 & 12183 & 1000 \\
lt\_dev & 14285 & 13411 & 1000 \\
lv\_dev & 15584 & 13843 & 1000 \\
mg\_dev & 1621 & 1758 & 100 \\
mi\_dev & 2988 & 2981 & 100 \\
mk\_dev & 18411 & 17093 & 1000 \\
ms\_dev & 9490 & 9162 & 1000 \\
my\_dev & 4976 & 3125 & 100 \\
ne\_dev & 1949 & 1382 & 100 \\
nl\_dev & 12195 & 12649 & 1000 \\
nn\_dev & 15423 & 16451 & 1000 \\
pdc\_dev & 1570 & 1616 & 100 \\
pl\_dev & 14440 & 14606 & 1000 \\
pt\_dev & 10296 & 10399 & 1000 \\
ro\_dev & 12034 & 12160 & 1000 \\
ru\_dev & 13738 & 13835 & 1000 \\
rw\_dev & 1387 & 1326 & 100 \\
sd\_dev & 3302 & 2834 & 100 \\
sk\_dev & 15142 & 14131 & 1000 \\
sl\_dev & 13155 & 12487 & 1000 \\
so\_dev & 1800 & 1394 & 100 \\
sr\_dev & 15092 & 12946 & 1000 \\
sv\_dev & 13972 & 15943 & 1000 \\
ta\_dev & 22159 & 16942 & 1000 \\
tg\_dev & 1681 & 1918 & 100 \\
th\_dev & 60413 & 72584 & 1000 \\
tl\_dev & 7327 & 7352 & 1000 \\
tr\_dev & 15011 & 13392 & 1000 \\
tt\_dev & 17031 & 19500 & 1000 \\
ug\_dev & 2663 & 2360 & 100 \\
uk\_dev & 17042 & 16073 & 1000 \\
ur\_dev & 14189 & 11129 & 1000 \\
vi\_dev & 8754 & 8755 & 1000 \\
yo\_dev & 1354 & 1463 & 100 \\
zh\_dev & 22833 & 37362 & 1000\\
\bottomrule
\end{xltabular}}

\small{
\begin{xltabular}{\textwidth}{llll}
\toprule
Source & \multicolumn{2}{l}{Source Tokens} & \multicolumn{1}{l}{Num. Examples} \\
 & \multicolumn{1}{l}{mBERT} & \multicolumn{1}{l}{XLM-R} & \multicolumn{1}{l}{} \\
\toprule
\endfirsthead
\toprule
Source & \multicolumn{2}{l}{Source Tokens} & \multicolumn{1}{l}{Num. Examples} \\
 & \multicolumn{1}{l}{mBERT} & \multicolumn{1}{l}{XLM-R} & \multicolumn{1}{l}{} \\
\toprule
\endhead
\toprule
&\multicolumn{2}{c}{\textit{Continued on next page}}  \\
\toprule
\endfoot
\bottomrule
\\ 
\caption{NER Source Token Counts} \label{tab:ner_source_token_counts}
\endlastfoot
af\_train & 18515 & 18890 & 1000 \\
be\_train & 16393 & 14160 & 1000 \\
bg\_train & 15437 & 14366 & 1000 \\
bn\_train & 13055 & 10154 & 1000 \\
br\_train & 14047 & 14708 & 1000 \\
ca\_train & 9613 & 9694 & 1000 \\
ckb\_train & 8503 & 16505 & 1000 \\
cs\_train & 15216 & 14844 & 1000 \\
cy\_train & 15764 & 15279 & 1000 \\
da\_train & 12868 & 13042 & 1000 \\
de\_train & 14609 & 15870 & 1000 \\
el\_train & 22705 & 17864 & 1000 \\
eo\_train & 12347 & 11731 & 1000 \\
es\_train & 10965 & 10681 & 1000 \\
et\_train & 16654 & 15899 & 1000 \\
eu\_train & 16099 & 17887 & 1000 \\
fi\_train & 16416 & 16082 & 1000 \\
fr\_train & 10894 & 11579 & 1000 \\
ga\_train & 15228 & 14307 & 1000 \\
hi\_train & 14020 & 10345 & 1000 \\
hr\_train & 14915 & 14417 & 1000 \\
hu\_train & 17808 & 16664 & 1000 \\
hy\_train & 20793 & 15550 & 1000 \\
id\_train & 9783 & 9839 & 1000 \\
it\_train & 11989 & 12270 & 1000 \\
ja\_train & 31087 & 52334 & 1000 \\
ko\_train & 18293 & 18308 & 1000 \\
la\_train & 10372 & 11720 & 1000 \\
lt\_train & 14992 & 14137 & 1000 \\
lv\_train & 15948 & 14318 & 1000 \\
mk\_train & 18375 & 17090 & 1000 \\
ms\_train & 9696 & 9343 & 1000 \\
nl\_train & 13036 & 13436 & 1000 \\
nn\_train & 15713 & 16631 & 1000 \\
pl\_train & 14120 & 14366 & 1000 \\
pt\_train & 10089 & 9931 & 1000 \\
ro\_train & 12168 & 12339 & 1000 \\
ru\_train & 13698 & 13703 & 1000 \\
sk\_train & 14820 & 13884 & 1000 \\
sl\_train & 12743 & 11799 & 1000 \\
sr\_train & 15033 & 12886 & 1000 \\
sv\_train & 13362 & 15089 & 1000 \\
ta\_train & 22669 & 17414 & 1000 \\
th\_train & 59299 & 70855 & 1000 \\
tl\_train & 7363 & 7399 & 1000 \\
tr\_train & 14478 & 12894 & 1000 \\
tt\_train & 16811 & 19726 & 1000 \\
uk\_train & 17615 & 16453 & 1000 \\
ur\_train & 14422 & 11286 & 1000 \\
vi\_train & 8981 & 8983 & 1000 \\
zh\_train & 21121 & 34428 & 1000 \\
\bottomrule
\end{xltabular}}

\end{document}